\documentclass[11pt]{article}

\usepackage[pdftex]{graphicx}
\usepackage[small,bf,sf,textfont={small,sf}]{caption}
\usepackage{vmargin}
\usepackage{subfig}
\usepackage{amsmath}
\usepackage{amssymb}
\usepackage{fancyheadings}
\usepackage{color}
\usepackage[colorlinks=true,
            linkcolor=blue,
            urlcolor=blue,
            citecolor=blue]{hyperref}
\usepackage[numbers]{natbib}
\usepackage{doi}
\usepackage{booktabs}
\usepackage{float}
\usepackage{lineno}
\usepackage{epstopdf}
\usepackage{bm}
\usepackage{dsfont}
\usepackage{multirow}
\usepackage{pdflscape}

\floatstyle{ruled}
\newfloat{floatbox}{thp}{lob}
\floatname{floatbox}{Algorithm}

\newcommand{\trp}{^{\scriptsize \top}}

\newcommand{\inv}{^{\scriptsize -1}}
\newcommand{\sqr}{^{\scriptsize \textrm{1/2}}}

\newcommand{\x}{\mathbf{x}}

\newcommand{\y}{\mathbf{y}}
\newcommand{\z}{\mathbf{z}}
\newcommand{\m}{\mathbf{m}}
\newcommand{\p}{\mathbf{p}}
\newcommand{\e}{\mathbf{e}}

\newcommand{\dsim}{\mathbf{d}}
\newcommand{\dobs}{\mathbf{d}_{\textrm{obs}}}

\newcommand{\C}{\mathbf{C}}

\newcommand{\Ce}{\mathbf{C}_{\mathbf{e}}}

\newcommand{\R}{\mathbf{R}}

\newcommand{\G}{\mathcal{G}}
\newcommand{\D}{\mathcal{D}}
\newcommand{\Ex}{\mathcal{E}}

\setpapersize{USletter}
\setmarginsrb{1.5in}{1in}{1in}{0.5in}{12pt}{11mm}{12pt}{30pt}

\begin{document}

\date{}

\title{Recent Developments Combining Ensemble Smoother and Deep Generative Networks for Facies History Matching}

\author{Smith W. A. Canchumuni$^1$ \\ Jose D. B. Castro$^2$ \\ J\'ulia Potratz$^3$ \\ Alexandre A. Emerick$^4$ \\  Marco Aurelio C. Pacheco$^5$}

\maketitle

\begin{small}
\begin{center}
$^1$Pontifical Catholic University of Rio de Janeiro, Brazil, \href{mailto://saraucoc@uni.pe}{\texttt{saraucoc@uni.pe}}\\
$^2$Pontifical Catholic University of Rio de Janeiro, Brazil, \href{mailto://bermudez@ele.puc-rio.br}{\texttt{bermudez@ele.puc-rio.br}}\\
$^3$Pontifical Catholic University of Rio de Janeiro, Brazil, \href{mailto://jupotratz@gmail.com}{\texttt{jupotratz@gmail.com}} \\
$^4$Petrobras Research and Development Center, Brazil, \href{mailto://emerick@petrobras.com.br}{\texttt{emerick@petrobras.com.br}} \\
$^5$Pontifical Catholic University of Rio de Janeiro, Brazil, \href{mailto://marco@ele.puc-rio.br}{\texttt{marco@ele.puc-rio.br}}
\end{center}
\end{small}

\section*{Abstract}

Ensemble smoothers are among the most successful and efficient techniques currently available for history matching. However, because these methods rely on Gaussian assumptions, their performance is severely degraded when the prior geology is described in terms of complex facies distributions. Inspired by the impressive results obtained by deep generative networks in areas such as image and video generation, we started an investigation focused on the use of autoencoders networks to construct a continuous parameterization for facies models. In our previous publication, we combined a convolutional variational autoencoder (VAE) with the ensemble smoother with multiple data assimilation (ES-MDA) for history matching production data in models generated with multiple-point geostatistics. Despite the good results reported in our previous publication, a major limitation of the designed parameterization is the fact that it does not allow applying distance-based localization during the ensemble smoother update, which limits its application in large-scale problems.

The present work is a continuation of this research project focusing in two aspects: firstly, we benchmark seven different formulations, including VAE, generative adversarial network (GAN), Wasserstein GAN, variational auto-encoding GAN, principal component analysis (PCA) with cycle GAN, PCA with transfer style network and VAE with style loss. These formulations are tested in a synthetic history matching problem with channelized facies. Secondly, we propose two strategies to allow the use of distance-based localization with the deep learning parameterizations.

\vspace{2mm}

\noindent \textbf{Keywords:} Deep Learning; Data Assimilation; Facies Models; Ensemble Smoother


\section{Introduction}
\label{Sec:Intro}

Ensemble smoother with multiple data assimilation (ES-MDA) \citep{emerick:13b} has been used as a robust history-matching technique due to its ability to condition multiple realizations of reservoir models with a balanced computational cost. However, similarly to other ensemble-based methods, ES-MDA relies on Gaussian assumptions which degrades its performance when the prior geology is described in terms of complex facies distributions. In these cases, the posterior models do not present the expected geological features displayed in the prior ones, loosing the geological plausibility of the ensemble.

The typical procedure to improve the data assimilation performance in case of facies is re-parameterization and the number of works in this direction is quite extensive. Among the methods proposed in the literature we can cite techniques based on principal component analysis (PCA) \citep{sarma:08,vo:14a,chen:16a,emerick:17a}, discrete cosine transform \citep{jafarpour:08a}, sparse dictionary learning with K-SVD \citep{khaninezhad:12a}, level-set functions \citep{chang:10,moreno:11a,lorentzen:12a,ping:14a}, truncated pluri-Gaussian simulation \citep{liu:05a,sebacher:13a}, among many others. Most of these works attempt to re-parameterize geological facies realizations generated with multiple-point geostatistical (MPG) methods \citep{mariethoz:14bk}. Despite the large number of works in this direction, the development of robust parameterization methods for facies history matching is still a problem with no definitive solution.

\subsection{Related Work}
\label{Sec:RelatedWork}

The use of deep learning for parameterization of geological facies has called the attention of some research groups in the last few years. For example, \citet{laloy:17b} and \citet{laloy:18a} applied a variational autoencoder (VAE) \citep{kingma:13a} and a spatial generative adversarial network (GAN) \citep{goodfellow:14b,jetchev:16a} to re-parameterize realizations of facies combined to a Markov chain Monte Carlo method for data assimilation. \citet{chan:17a} trained a Wasserstein GAN \citep{arjovsky:17a} as a generative model for facies and \citep{chan:19b} extended the work by stacking a second network for conditioning the realizations to facies data points (hard data). The problem of conditioning facies realizations to hard data was also addressed by \citet{dupont:18a}, which used the semantic inpainting GAN from \citep{yeh:16a}. \citet{liu:19a} combined ideas from optimization-based PCA \citep{vo:14a} with the transfer style network from \citep{gatys:16a,johnson:16a} as a re-parameterization strategy for facies history matching. More recently, \citet{mosser:19a} presented an implementation named DeepFlow, which uses adjoint-based gradient descent to update the latent representation of facies realizations in a pre-trained GAN for history matching.

Our first attempts to combine ensemble data assimilation with deep learning for facies history matching used standard autoencoders  \citep{canchumuni:17a} and deep belief networks \citep{canchumuni:19a}. In our most recent publication \citep{canchumuni:19b}, we combined a convolutional variational autoencoder (VAE) with ES-MDA for conditioning facies models to static (hard) and dynamic (production) data. The results showed that the combination of ensemble data assimilation and deep learning is a promising research direction, but they also revealed some limitations that need to be addressed before the method can be used operationally. One major limitation of the method described in \citep{canchumuni:19b} is the fact that it does not allow the use of distance-based localization \citep{houtekamer:01} during the ensemble smoother update, which limits its application in large-scale problems. Other problems reported in \citep{canchumuni:19b} include the high computational requirements for training the VAE network and the still limited reconstruction performance for 3D problems.

\subsection{Contributions of this Work}
\label{Sec:Contributions}

In this work, we present our more recent developments in the direction of developing a deep leaning-based parameterization for facies history matching. The contribution of this work is twofold:

\begin{itemize}
  \item We benchmark seven different deep learning formulations for parameterization, including VAE, GAN, Wasserstein GAN, Variational Auto-encoding GAN, Cycle-GAN, Transfer Style Networks and VAE with style loss.
  \item We extended our original formulation based on VAE to allow localization and propose a more effective formulation based on PCA and Cycle-GAN that also allows localization during the data assimilation.
\end{itemize}

The remaining of the paper is organized as follows. In the next section, we briefly present the background of the methods discussed in the paper. In particular, we introduce generative models and discuss the formulation of each network evaluated in the paper. After that, we present the workflow adopted to combine the generative networks with ES-MDA for history matching followed by a discussion on the architecture of each network. The different networks are tested in a simple history-matching problem based on a 2D channelized facies model generated with the MPG algorithm of \citet{strebelle:02}. In the second part of the paper, we introduce two strategies to allow the use of distance-based localization during the ES-MDA updates. In this case, we consider only two network formulations, one based on VAE and another based on Cycle-GANs \citep{zhu:17a}. The localization strategies are tested in another 2D problem with a large number of wells. The last section of the paper summarizes the conclusions. Details about the architecture of each neural networks used in this work are provided in the Appendix.

\section{Background}
\label{Sec:Background}

\subsection{Generative Models}
\label{Sec:GenerativeModels}

\begin{figure}[!h]
    \centering
    \includegraphics[width=0.5\textwidth]{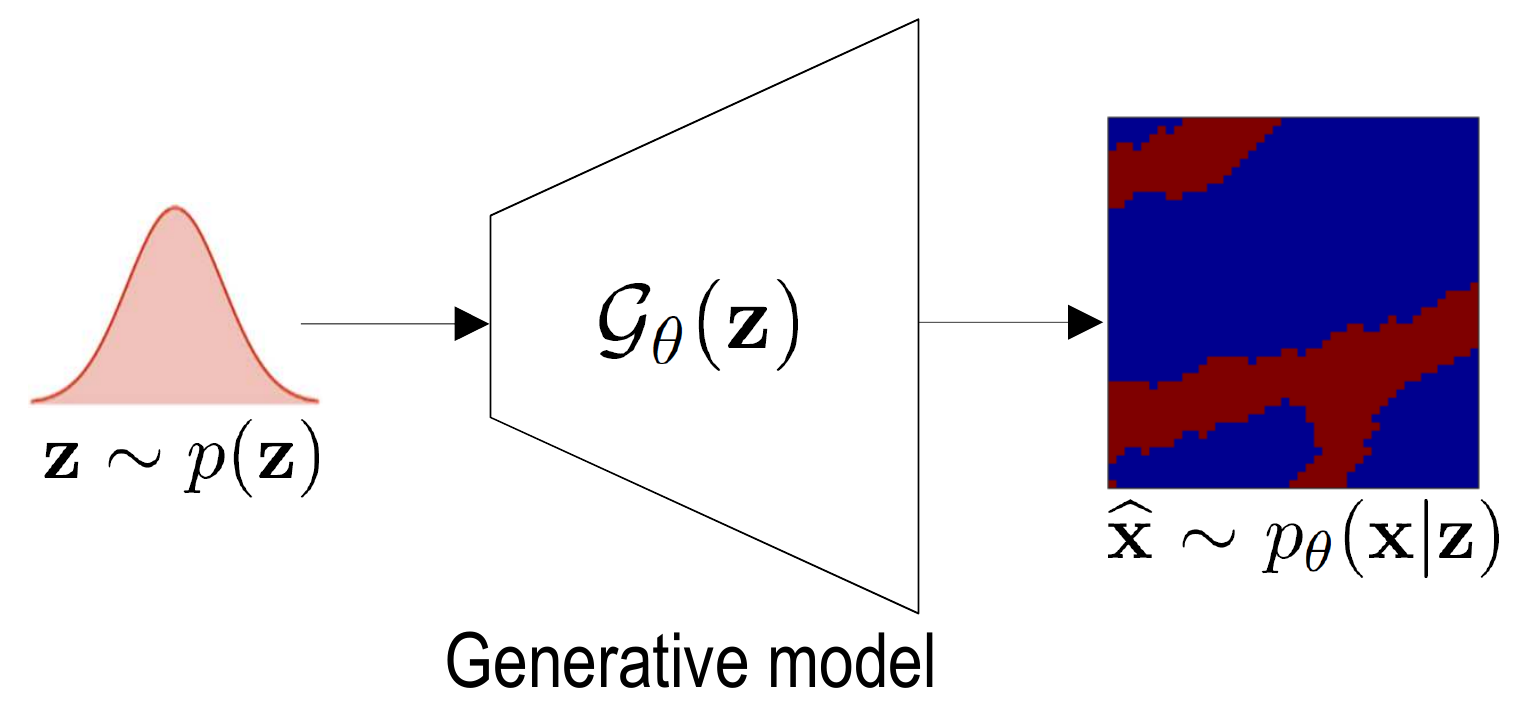}
    \caption{Schematic of a generative model}
    \label{Fig:GenerativeModel}
\end{figure}

In machine learning, generative models are parametrical models focused on learning how to generate samples from complex probability distributions. A typical application of a generative model aims to generate new samples $\x$ from a probability density function (PDF) $p(\x)$. This PDF is often known only from a given set of samples (training data). Then, we construct a deterministic function $\G_\theta(\z)$ which receives a random argument $\z \sim p(\z)$, following a know PDF. $\G_\theta(\z)$ is modeled as a neural network, trained with a set of data points $\x_i$ such that if we provide $\z \sim p(\z)$, it generates $\widehat{\x} \sim p_\theta(\x | \z)$ which resembles samples from $p(\x)$. This process is illustrated in Fig.~\ref{Fig:GenerativeModel}.

In the application of interest of this paper, $\x$ represents a realization of facies in a reservoir model. Training samples, $\x_i$, are available from MPG algorithms. Our goal is to use deep learning to construct generative models that are able to generate new facies realizations $\widehat{\x}$, which are indistinguishable from realizations generated with geostatistics. After training, the random vectors $\z$ are used as a continuous parameterization for history matching with ES-MDA. In the following sections, we briefly review the deep learning techniques investigated in this work.

\subsubsection{Comments About Notation}
\label{Sec:notation}

In this paper, we discuss several generative models proposed in the literature, which makes important to clarify the overall notation first. We reserve $\x \in \mathds{R}^{N_x}$ and $\z \in \mathds{R}^{N_z}$ to denote the vectors of facies and latent representation, respectively. Both vectors belong to real-valued spaces of dimensions $N_x$ and $N_z$, respectively. We reserve $p(\cdot)$ to represent a PDF and upper case letters with calligraphic font to denote nonlinear functions. In particular, $\G$ denotes a generative model and $\D$ denotes a discriminative model. A Greek letter subscript in a function means that this function has been modeled by a neural network. For example, $p_\theta(\x)$ means that we use a neural network with learning parameters $\theta$ to model the actual distribution $p(\x)$. The same is valid for other nonlinear functions. For example, we denote a trained generative model as  $\G_\theta$. Finally, we use $\mathcal{L}$ to denote loss functions used to train the networks.

\subsubsection{Variational Autoencoders}
\label{Sec:VAE}

\begin{figure}[!h]
    \centering
    \includegraphics[width=0.7\textwidth]{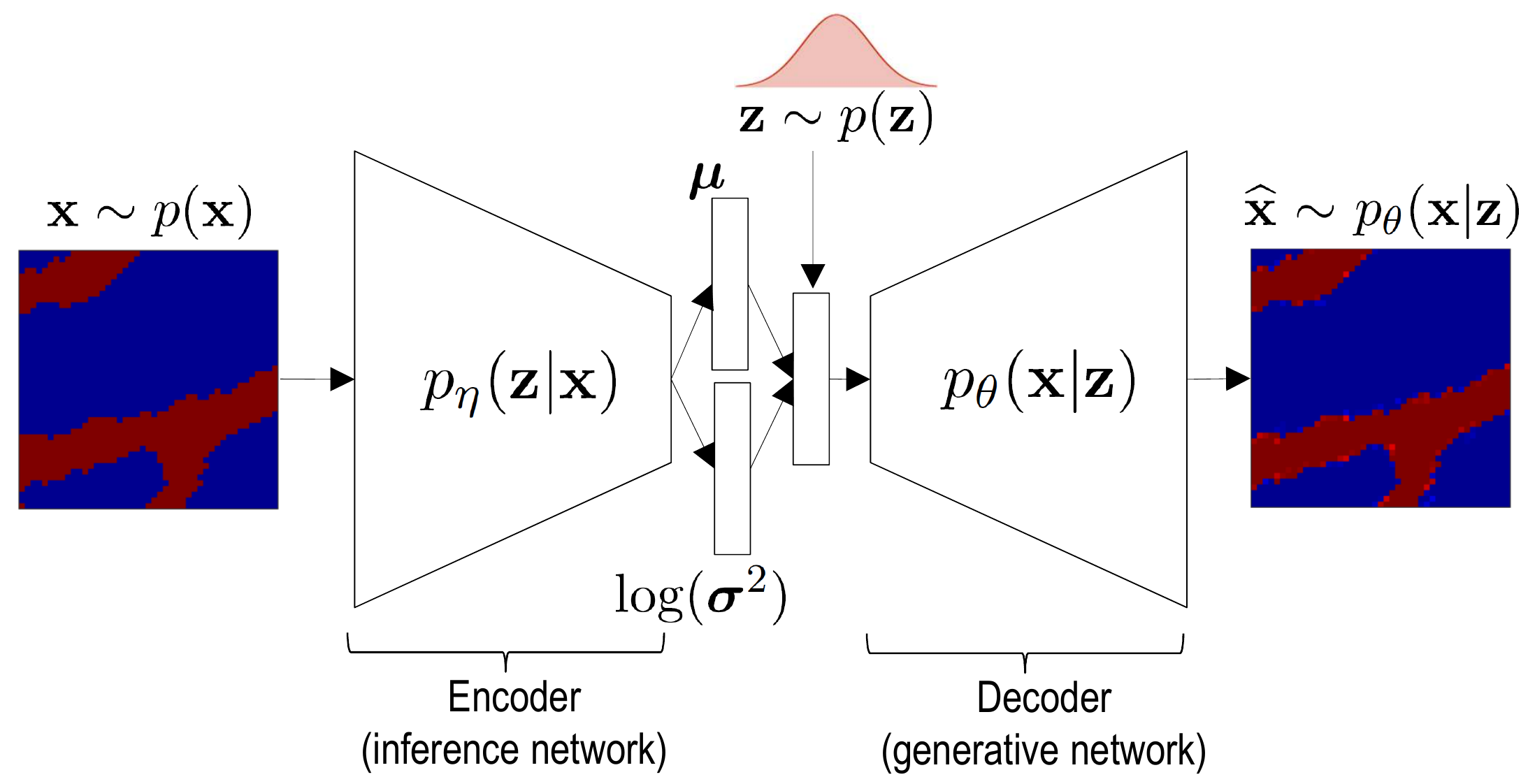}
    \caption{Schematic architecture of a variational autoencoder network}
    \label{Fig:VAE}
\end{figure}

A variational autoencoder (VAE) \citep{kingma:13a} is a neural network used in applications such as data compression, noise removal, feature learning, and synthesization of new samples. Similarly to standard autoencoders, a VAE is composed of two networks (Fig.~\ref{Fig:VAE}) an ``encoder'' that learns how to map the input data $\x$ into a latent representation $\z$ and a ``decoder'' that makes the inverse mapping reconstruction $\x$ given the latent representation $\z$. However, unlike standard autoencoders, the central layer of a VAE has an additional layer responsible for sampling both the latent vector $\z$ and an extra term for the loss function, forcing the latent vector to follow a designed prior distribution, $p(\z)$, usually assumed a standard Gaussian, $\mathcal{N}(\mathbf{0},\mathbf{I})$.

The loss function minimized during the training of a VAE has the following form

\begin{equation}\label{Eq:VAE}
  \mathcal{L}_{\textrm{VAE}} (\theta, \eta) =  \mathcal{L}_{\textrm{RE}}(\theta) + \textrm{KL} \left( p_\eta(\z|\x) \| p(\z) \right),
\end{equation}
where $\mathcal{L}_{\textrm{RE}}(\theta)$ is the reconstruction error. The minimization of this term makes $\widehat{\x}$ to resamble $\x$. In this work, we use either the binary cross-entropy or the mean squared error depending if we use a sigmoid or hyperbolic tangent function (tanh), respectively. The binary cross-entropy push the reconstruction to generate values close to zero or one and it is computed as

\begin{equation}\label{Eq:VAE-RE}
\mathcal{L}_{\textrm{RE,CE}} (\theta) = - \frac{1}{N_x}\sum_{i=1}^{N_x}\left[x_i \log(\widehat{x}_i) + (1 - x_i) \log(1-\widehat{x}_i)\right].
\end{equation}

The mean squared error is used to obtain reconstruction values within the range $[-1,1]$ and it is given by
\begin{equation}\label{Eq:VAE-RE-MSE}
\mathcal{L}_{\textrm{RE,MSE}} (\theta) =  \frac{1}{N_x}\sum_{i=1}^{N_x}\left(\widehat{x}_i-x_i\right)^2.
\end{equation}

$\textrm{KL}\left( p_\eta(\z|\x) \| p(\z) \right)$ in Eq.~\ref{Eq:VAE} is the Kullback-Leibler divergence from $p_\eta(\z|\x)$ to $p(\z)$. Minimization of this term pushes the encoding distribution $p_\eta(\z|\x)$ towards a standard Gaussian $\mathcal{N}(\mathbf{0},\mathbf{I})$. $\textrm{KL}\left( p_\eta(\z|\x) \| p(\z) \right)$ is computed as

and
\begin{equation}\label{Eq:VAE-KL}
  \textrm{KL} \left( p_\eta(\z|\x) \| p(\z) \right) = \frac{1}{2} \sum_{i=1}^{N_z} \left( \mu_i^2 + \sigma_i^2 - \log \left(\sigma_i^2\right) - 1 \right),
\end{equation}
$\mu_i$ and $\sigma_i^2$ are components of the vectors of the mean and variance of the distribution $p_\eta(\z|\x)$. More details about VAE can be found in \citep{kingma:13a,doersch:16a}.

\subsubsection{Generative Adversarial Networks}
\label{Sec:GAN}

\begin{figure}[!h]
    \centering
    \includegraphics[width=0.7\textwidth]{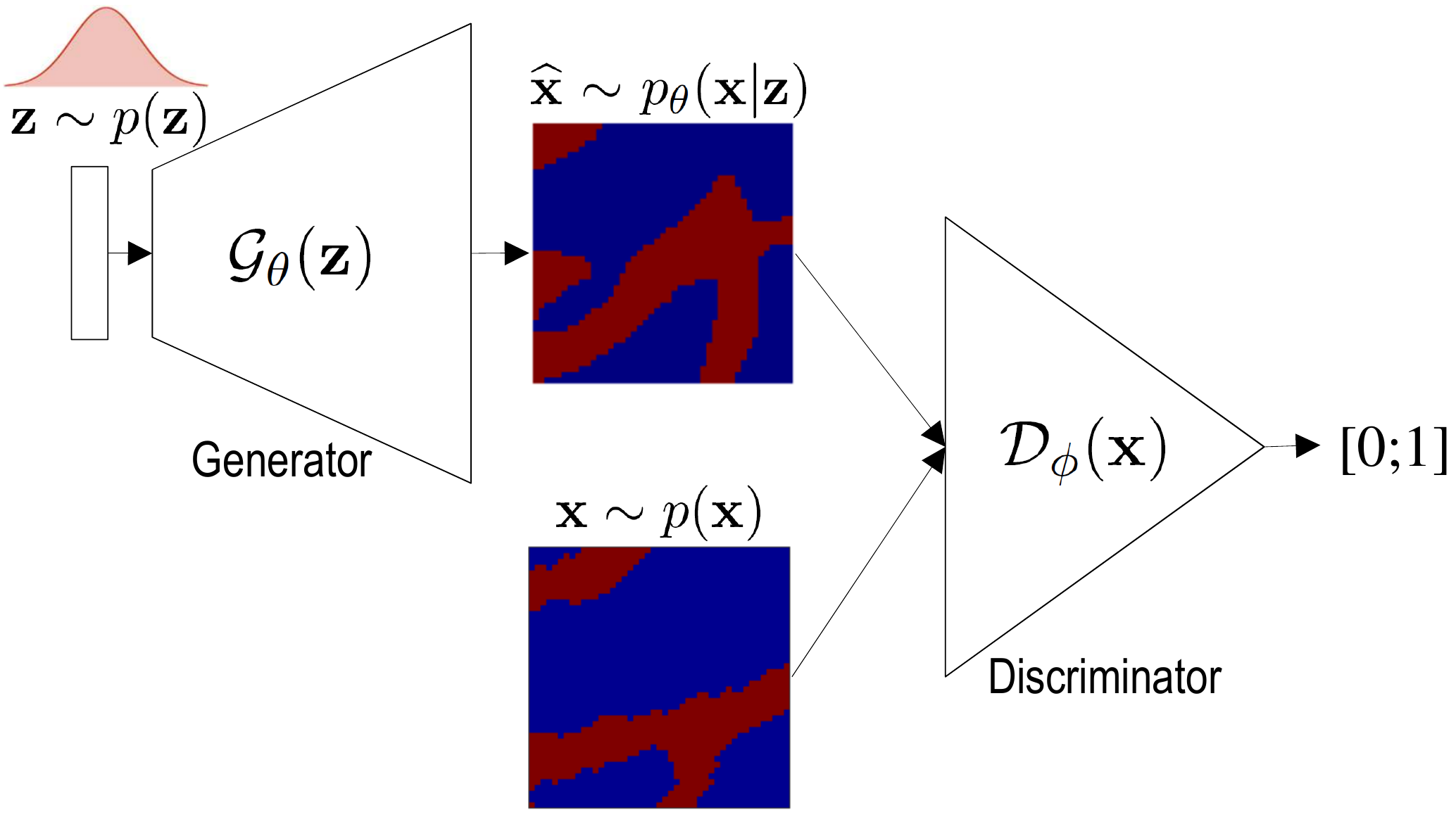}
    \caption{Schematic architecture of a generative adversarial network}
    \label{Fig:GAN}
\end{figure}

Generative adversarial networks (GAN) \citep{goodfellow:14b} are generative models composed of two neural networks (Fig.~\ref{Fig:GAN}), a ``generator,'' denoted by $\G_\theta(\z)$, and a ``discriminator,'' denoted by $\D_\phi(\x)$. In this scheme, $\G_\theta(\z)$ seeks to synthesize realistic samples $\widehat{\x} \sim p_\theta(\x|\z)$ that $\D_\phi(\x)$ cannot distinguish whether this sample is from the distribution $p_\theta(\x|\z)$ or from the actual distribution $p(\x)$. While $\G_\theta(\z)$ returns a vector, $\D_\phi(\x)$ returns a scalar that represents the probability that $\x$ came from $p(\x)$. Formally, $\G_\theta(\z)$ and $\D_\phi(\x)$ are trained following a \textit{min-max} problem

\begin{equation} \label{Eq:Lgan}
   \mathrm{arg} \min_{\theta} \max_{\phi} \mathcal{L}_{\textrm{GAN}}(\theta, \phi)
\end{equation}
where
\begin{equation} \label{Eq:gansobjective}
  \mathcal{L}_{\textrm{GAN}}(\theta, \phi) = \mathbb{E}_{\x \sim p(\x)}\left[\log \D_\phi(\x) \right] + \mathbb{E}_{\z \sim p(\z)} \left[ \log\left(1 - \D_\phi(\G_\theta(\z))\right)\right].
\end{equation}

\citet{goodfellow:14b} showed that if $\G_\theta(\z)$ and $\D_\phi(\x)$ have enough capacity, the equilibrium solution of the \textit{min-max} problem leads to $\G_\theta(\z)$ to sample the actual distribution $p(\x)$, in which case $\D_\phi(\x)$ always result in $1/2$, i.e., the discriminator cannot distinguish between training data and generated samples. In practice, however, standard implementation of GAN networks may present instability during training leading to a situation where $\G_\theta(\z)$  fails to capture the diversity of actual samples, and generates samples only from a specific region of the distribution. This situation is sometimes referred to a mode collapse problem \citep{goodfellow:17a}. This fact led to the proposal of several strategies to improve the training of GANs; see, e.g., \citep{hong:19a} for a recent overview.

\subsubsection{Wasserstein GAN}
\label{Sec:WGAN}

Wasserstein GANs (WGAN) were proposed by \citep{arjovsky:17a} aiming to improve the stability of the training process using a loss function that correlates with the quality of generated images. Essentially, a WGAN seeks to minimize an approximation of the Wasserstein distance between the true distribution $p(\x)$ and the distribution $p_\theta(\x)$ learned by the generator. The mathematical formulation that supports WGANs is quite dense, but the final method requires very few modifications in a standard GAN. The main change is the loss function which is given by

\begin{equation}\label{Eq:LossWGAN}
  \mathcal{L}_{\textrm{WGAN}}(\theta, \omega) = \mathbb{E}_{\x\sim \p(\x)}\left[ \mathcal{F}_\omega(\x) \right] - \mathbb{E}_{\z\sim \p(\z)}\left[ \mathcal{F}_\omega\left(\mathcal{G}_\theta(\x)\right) \right],
\end{equation}
where $\mathcal{F}_\omega(\x)$ is sometimes referred to as ``critic.'' Unlike traditional GANs, which use a discriminator to classify samples in real or fake, WGANs use a critic that computes a score related to the realness (or fakeness) of a given sample. \citet{arjovsky:17a} showed examples indicating that the lower the loss of the critic, the higher the expected quality of the generated images, which may not be true with the standard GAN loss function. Therefore, instead of seeking for an equilibrium between a generator and a discriminator, WGANs seeks for convergence. \citet{arjovsky:17a} presents a pseudo-code of the training process of a WGAN.

\subsubsection{Auto-Encoding Generative Adversarial Networks: $\alpha$-GAN}
\label{Sec:AlphaGan}

\begin{figure}[!h]
    \centering
    \includegraphics[width=0.8\textwidth]{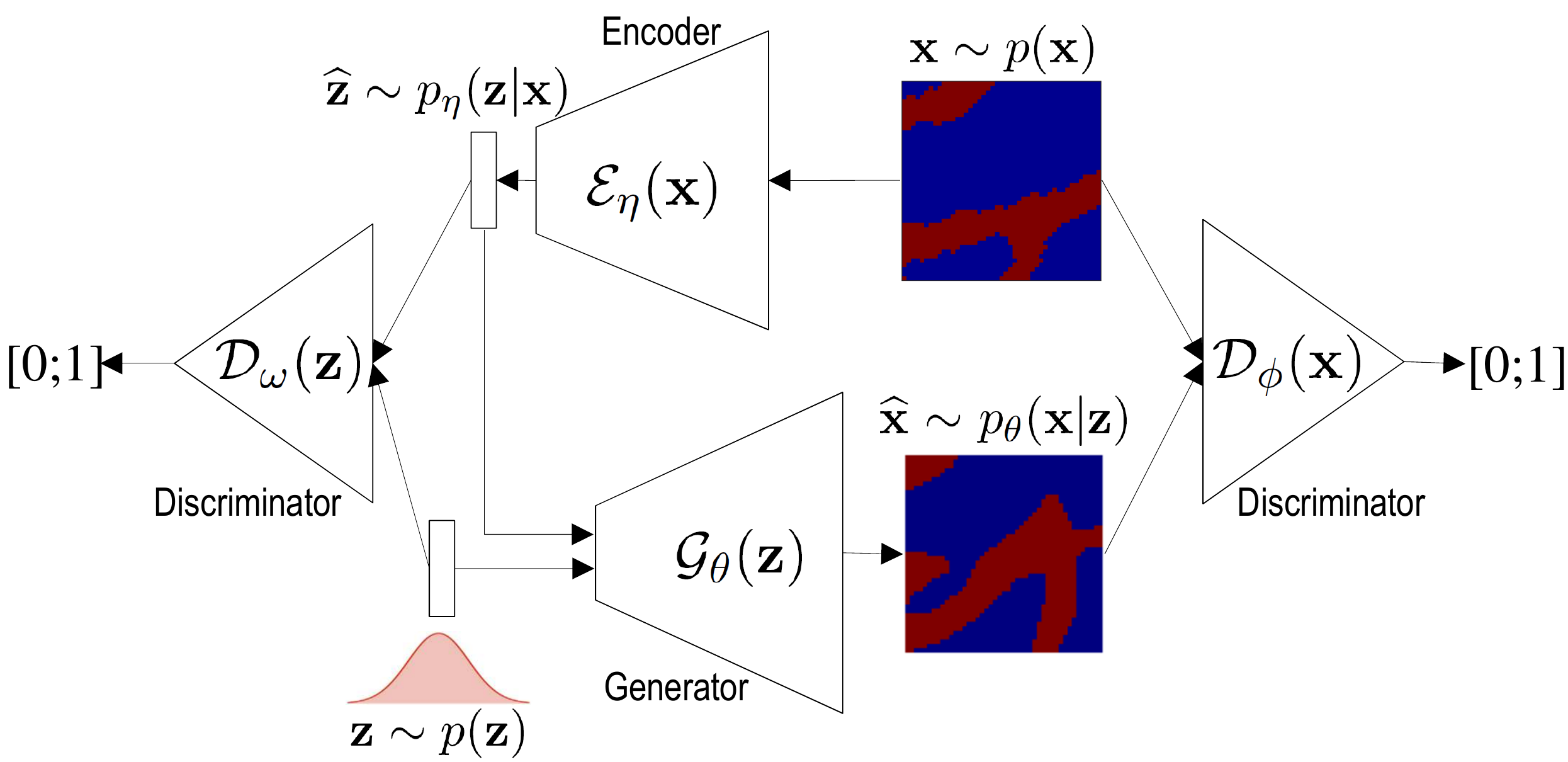}
    \caption{Schematic architecture of an $\alpha$-GAN}
    \label{Fig:AlphaGAN}
\end{figure}

$\alpha$-GAN \citep{rosca:17a} combines a GAN, which discriminates between real and fake samples and a reconstruction loss given by VAE to avoid the training instabilities and prevent the collapse problem commonly found in traditional GANs. An $\alpha$-GAN consists of four networks (Fig.~\ref{Fig:AlphaGAN}): a generative network $\G_\theta(\z)$, one discriminator $\D_\phi(\x)$ used to classify between reconstructions from a decoder and real training samples; a second discriminator $\mathcal{D}_\omega(\z)$ used to discriminate between latent samples $\z$ produced by an encoder $\mathcal{E}_\eta(\x)$ used to represent the distribution $p_\eta(\z|\x)$ and samples from a standard Gaussian.

The training process consists of a nested minimization of four loss functions by updating the network parameters $\eta$, $\theta$, $\phi$ and $\omega$:

\begin{itemize}
    \item Encoder $\Ex_\eta(\x)$:
    \begin{equation} \label{Eq:alphaencoder}
    \mathcal{L}_{\Ex_\eta}(\theta, \eta, \omega) =  \mathbb{E}_{\x\sim p(\x)} \left[ c \| \x - \widehat{\x} \|_{1} + \mathcal{R}_{\D_\omega}(\widehat{\z}) \right]
    \end{equation}

    \item Generator $\G_\theta(\z)$:
    \begin{equation} \label{Eq:alphagenerator}
    \mathcal{L}_{\G_\theta}(\theta, \eta, \phi) =  \mathbb{E}_{\x\sim p(\x)} \left[ c \| \x - \widehat{\x} \|_{1} + \mathcal{R}_{\D_{\phi}}(\widehat{\x}) \right] + \mathbb{E}_{\z \sim p(\z)}[\mathcal{R}_{\D_{\phi}}(\G_\theta(\z))]
    \end{equation}

    \item Discriminator $\D_\phi(\x)$:
    \begin{equation} \label{Eq:alphadiscriminator}
    \mathcal{L}_{\D_\phi}(\theta, \eta, \phi) =  \mathbb{E}_{\x\sim p(\x)} \left[ -\log  \D_{\phi}(\x) - \log \left(1 - \D_{\phi}(\widehat{\x})\right) \right] + \mathbb{E}_{\z \sim p(\z)}\left[ -\log  \left( 1 - \D_{\phi}(\G_\theta(\z)) \right) \right]
    \end{equation}

    \item Discriminator $\D_\omega(\z)$:
    \begin{equation} \label{Eq:alphadiscriminator_code}
    \mathcal{L}_{\D_\omega}(\theta, \eta, \omega) =  \mathbb{E}_{\x\sim p(\x)}\left[-\log \left( 1- \D_\omega(\widehat{\z}) \right) \right] +  \mathbb{E}_{\z \sim p(\z)}\left[ -\log \D_\omega(\z) \right]
    \end{equation}
\end{itemize}
In the above equations, $c$ is a scale parameter which controls the relative weight of the $L_1$ norm of the reconstruction error. $\widehat{\z} = \Ex_\eta(\x)$ and $\widehat{\x} = \G_\theta(\widehat{\z})$ and

\begin{equation}
  \mathcal{R}_{\D_\omega}(\z) = -\log \D_\omega(\z) + \log(1-\D_\omega(\z)),
\end{equation}

\begin{equation}
  \mathcal{R}_{\D_{\phi}} (\x) = -\log \D_{\phi}(\x) + \log\left(1-\D_{\phi}(\x)\right).
\end{equation}
The Appendix~B of \citep{rosca:17a} presents a pseudo-code of the training procedure.

\subsubsection{Cycle-GAN}
\label{Sec:CycleGan}

\begin{figure}[h!]
\centering
\includegraphics[width=0.8\textwidth]{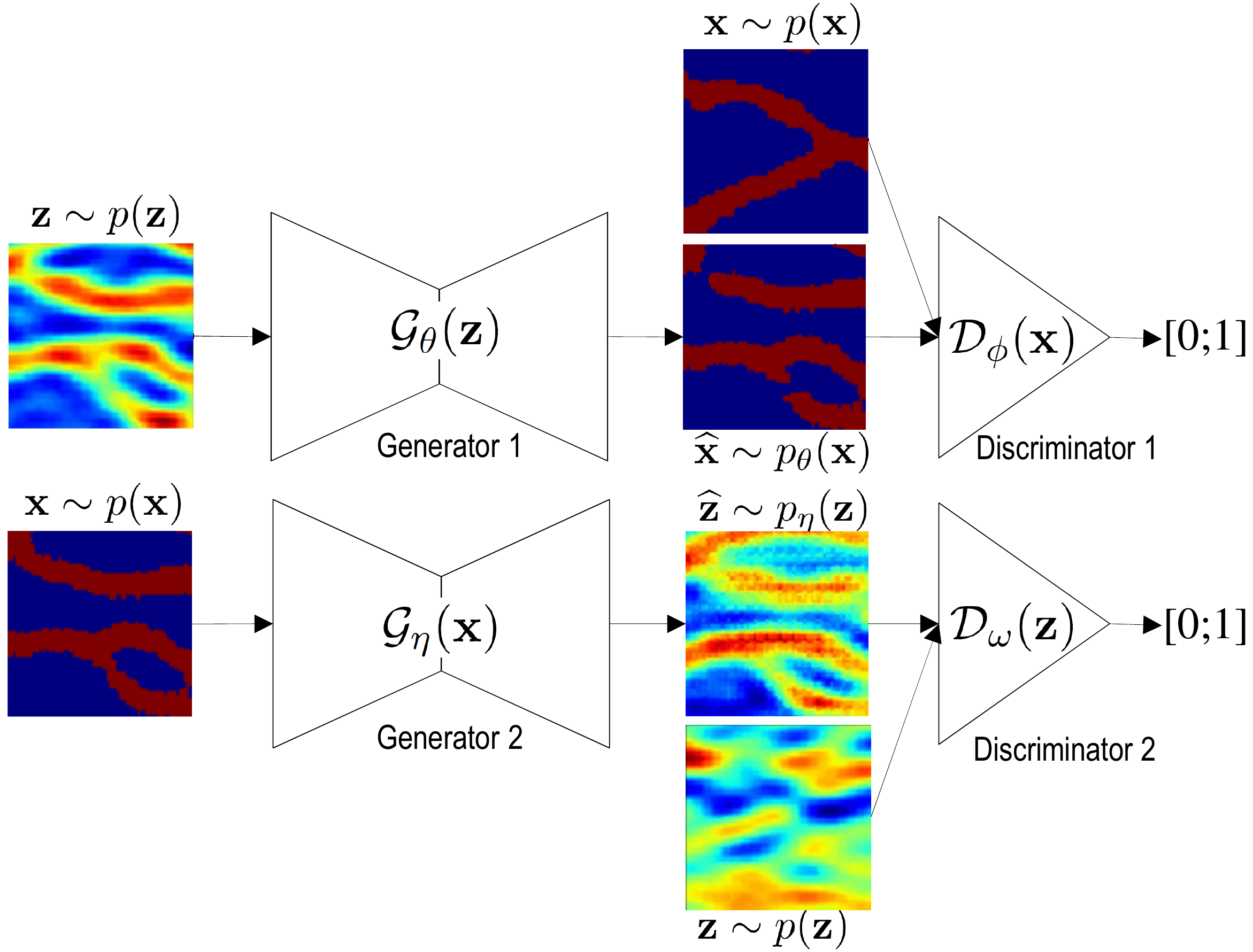}
\caption{Schematic architecture of a Cycle-GAN}
\label{Fig:cyclegans}
\end{figure}

Cycle-GANs \citep{zhu:17a} were proposed in the context of image-to-image translation, which allows to generate new images by combining the content of one image with the style of another. The classical example is to transform a photo to a painting style. Mathematically, the goal is to learn a nonlinear mapping function $\G_\theta(\z)$ which generates images $\widehat{\x}$ that are indistinguishable from images sampled from the actual target distribution $\x \sim p(\x)$. In practice, this mapping is not unique and may be highly under-constrained. Cycle-GANs address this problem by simultaneously learning an inverse mapping $\G_\eta (\x)$ and introducing a cycle consistency loss such that $\G_\eta ( \G_\theta (\z) ) \approx \z$ and $\G_\theta ( \G_\eta (\x) ) \approx \x$. Fig.~\ref{Fig:cyclegans} illustrates the structure of a Cycle-GAN, where $\z$ represents a continuous version of a facies $\x$. A Cycle-GAN consists of four networks, two discriminators $\D_\phi(\x)$ and $\D_\omega(\z)$ and two generators $\G_\theta (\z)$ and  $\G_\eta (\x)$. $\D_\phi(\x)$ aims to discern between $\x$ and $\widehat{\x} = \G_\theta (\z)$, while $\D_\omega(\z)$ aims to discern between $\z$ and $\widehat{\z} = \G_\eta (\x)$. The generators $\G_\theta (\z)$ and $\G_\eta (\x)$ are trained to fool their corresponding discriminators. The loss function of a Cycle-GAN has the form

\begin{equation}\label{Eq:cyclegan}
  \mathcal{L}(\theta, \eta, \phi, \omega)  = \mathcal{L}_{\textrm{GAN},1}(\theta, \phi) + \mathcal{L}_{\textrm{GAN},2}(\eta, \omega) + c \mathcal{L}_{\textrm{cyc}}(\theta, \eta)
\end{equation}
where,

\begin{equation} \label{Eq:cyclegan1}
    \mathcal{L}_{\textrm{GAN},1}(\theta, \phi) = \mathbb{E}_{\x \sim p(\x)} [\log \D_\phi (\x)] + \mathbb{E}_{\z \sim p(\z)}\left[ \log \left(1 - \D_\phi\left(\G_\theta(\z)\right) \right) \right],
\end{equation}

\begin{equation} \label{Eq:cyclegan2}
    \mathcal{L}_{\textrm{GAN},2}(\eta, \omega) = \mathbb{E}_{\z \sim p(\z)} [\log \D_\omega (\z)] + \mathbb{E}_{\x \sim p(\x)}\left[ \log \left(1 - \D_\omega\left(\G_\eta(\x)\right) \right) \right]
\end{equation}
and
\begin{equation} \label{Eq:cyclegan3}
    \mathcal{L}_{\textrm{cyc}}(\theta, \eta) = \mathbb{E}_{\z \sim p(\z)} \left[ \left\| \G_\eta\left( \G_\theta(\z)\right) - \z \right\|_1 \right] + \mathbb{E}_{\x \sim p(\x)} \left[ \left\| \G_\theta\left( \G_\eta(\x)\right) - \x \right\|_1 \right].
\end{equation}

In Eq.~\ref{Eq:cyclegan}, $c$ is a constant used to control the relative weight of the cycle loss in the total loss function.

\subsubsection{Transfer Style Networks}
\label{Sec:TransferStyleNetworks}

Transfer style network \citep{gatys:16a} is another technique to perform image-to-image translation, which is used to generate an image $\x$ with the appearance of a given image $\z$ and the visual style of another image $\y$ (style target). The process is carried out using a convolutional network pre-trained for image classification, usually a normalized version of the 16 convolutional layers of the VGG networks \citep{simonyan:15a}. The rationality of the method is that if we feed the image $\z$ through the network, the activations of each convolutional layer extract different features of the image, and deep layers tend to extract high level features which are increasingly sensitive to the actual content of the image, but relatively invariant to its appearance. Let $\mathcal{C}_\ell(\z)$ denote the content of the image $\z$ extracted in the $\ell$th convolutional layer of the network. $\mathcal{C}_\ell(\z)$ can be stored in a $N_f \times N_c$ matrix, where $N_f$ is the number of feature maps and $N_c$ is the size of the feature maps (height times the width) at the $\ell$th layer. The content loss between a new image $\x$ and the $\z$ is defined using the Frobenius norm

\begin{equation}
  \mathcal{L}_{\textrm{cont}}(\x, \z) = \left\| \mathcal{C}_\ell(\x) - \mathcal{C}_\ell(\z) \right\|_{\textrm{F}}^2.
\end{equation}

The ``style'' of an image $\y$ which is related to its texture information can also extracted computing correlations between the activations in different layers of the network. Therefore the style of $\y$ at the $\ell$th layer, that is referred to as Gram matrices and denoted by $\mathcal{S}_\ell(\y)$, is a $N_f \times N_f$ matrix given by

\begin{equation}
  \mathcal{S}_\ell(\y) = \mathcal{C}_\ell(\y)\mathcal{C}_\ell(\y)\trp,
\end{equation}
and the corresponding style loss is given by

\begin{equation}
  \mathcal{L}_{\textrm{style}}(\x, \y) = \sum_{\ell} \left\| \mathcal{S}_\ell(\x) - \mathcal{S}_\ell(\y) \right\|_{\textrm{F}}^2.
\end{equation}

The total loss is given by the weighted sum of the content and style losses
\begin{equation}\label{Eq:TransferLoss}
  \mathcal{L}_{\textrm{total}}(\x, \y, \z) =  c_1 \mathcal{L}_{\textrm{cont}}(\x, \z) + c_2 \mathcal{L}_{\textrm{style}}(\x, \y),
\end{equation}
where $c_1$ and $c_2$ are constants that define the relative weight of the content and style losses.

\begin{figure}[h!]
\centering
\includegraphics[width=0.7\textwidth]{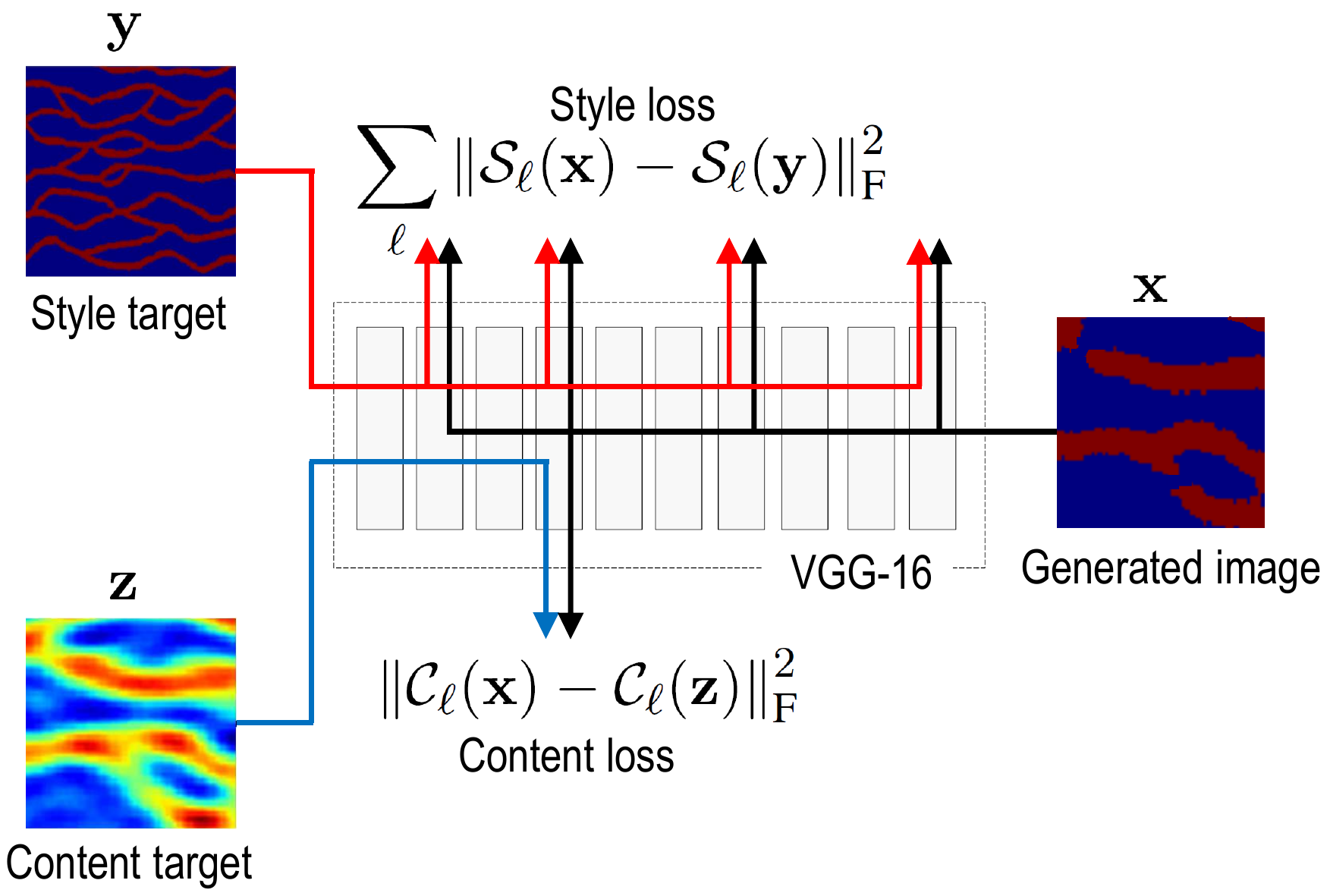}
\caption{Schematic architecture of a transfer style network}
\label{Fig:TransferStyle}
\end{figure}

The reconstruction of the image $\x$ combining the content of $\z$ and the style of $\y$ is done pixel-by-pixel by minimizing $\mathcal{L}_{\textrm{total}}(\x, \y, \z)$ using gradient descent. Note that unlike the previous methods, the parameters of the network are fixed, the minimization is with respect to $\x$. Fig.~\ref{Fig:TransferStyle} illustrates the schematic architecture of the transfer style network used in the work based on the first 10 convolutional layers of the VGG-16 network, without considering pooling layers.

\subsection{Ensemble Smoother with Multiple Data Assimilation}
\label{Sec:ESMDA}

ES-MDA \citep{emerick:13b} is an iterative version of the ensemble smoother motivated by the equivalence between single and multiple data assimilation for the linear-Gaussian case. This method performs multiples smaller corrections in the ensemble in each iteration by inflating the data-error covariance matrix. In its standard form, the number of data assimilations, $N_a$, and the inflated coefficients, $\alpha_k$ for $k=1,...,N_a$, must be selected such the condition $\sum_{k=1}^{N_a} \alpha_k\inv = 1$ is satisfied. The ES-MDA analysis for history matching is usually presented in terms of updating a vector of uncertain model parameters, typically gridblock petrophysical properties such as porosity and permeability. Here, we are interested in update facies which are re-parameterized in terms of a latent vector $\z$ using generative networks. Therefore, we present the ES-MDA equations in terms of the vector $\z$. Note, however, that it is still possible to simultaneously update the facies type and the petrophysical properties for each facies by augmenting the vector of model parameters updated by ES-MDA as discussed in \citep{emerick:17a}. The ES-MDA analysis equation for a vector $\z \in \mathds{R}^{N_{z}}$ can be written as

\begin{equation}\label{Eq:ES-MDA}
  \z_{j}^{k+1}=\z_{j}^{k} + \C_{\z\dsim}^{k}\left(\C_{\dsim\dsim}^{k}+\alpha_{k}\Ce\right)\inv \left(\dobs+\e_{j}^{k}-\dsim_j^k \right),
\end{equation}
for $j=1,\ldots,N_e$ with $N_e$ denoting the ensemble size. In this equation, $\dobs \in \mathds{R}^{N_{d}}$ is the vector of observed data; $\e_{j}^{k}\in\mathds{R}^{N_{d}}$ is the vector of random perturbations which is obtained by sampling $\mathcal{N}(\mathbf{0},\alpha_{k}\Ce)$, with $\Ce\in\mathds{R}^{N_{d}\times N_{d}}$ denoting the data-error covariance matrix. $\dsim_j^k \in \mathds{R}^{N_{d}}$ is the vector of predicted data. $N_d$ is the number of data points. The matrices $\C_{\z\dsim}^{k}\in \mathds{R}^{N_{z}\times N_{d}}$ and $\C_{\dsim\dsim}^{k}\in\mathds{R}^{N_{d}\times N_{d}}$ are estimated using the current ensemble.

\section{Methodology}
\label{Sec:Methodology}

\begin{figure}[!h]
    \centering
    \includegraphics[width=0.85\textwidth]{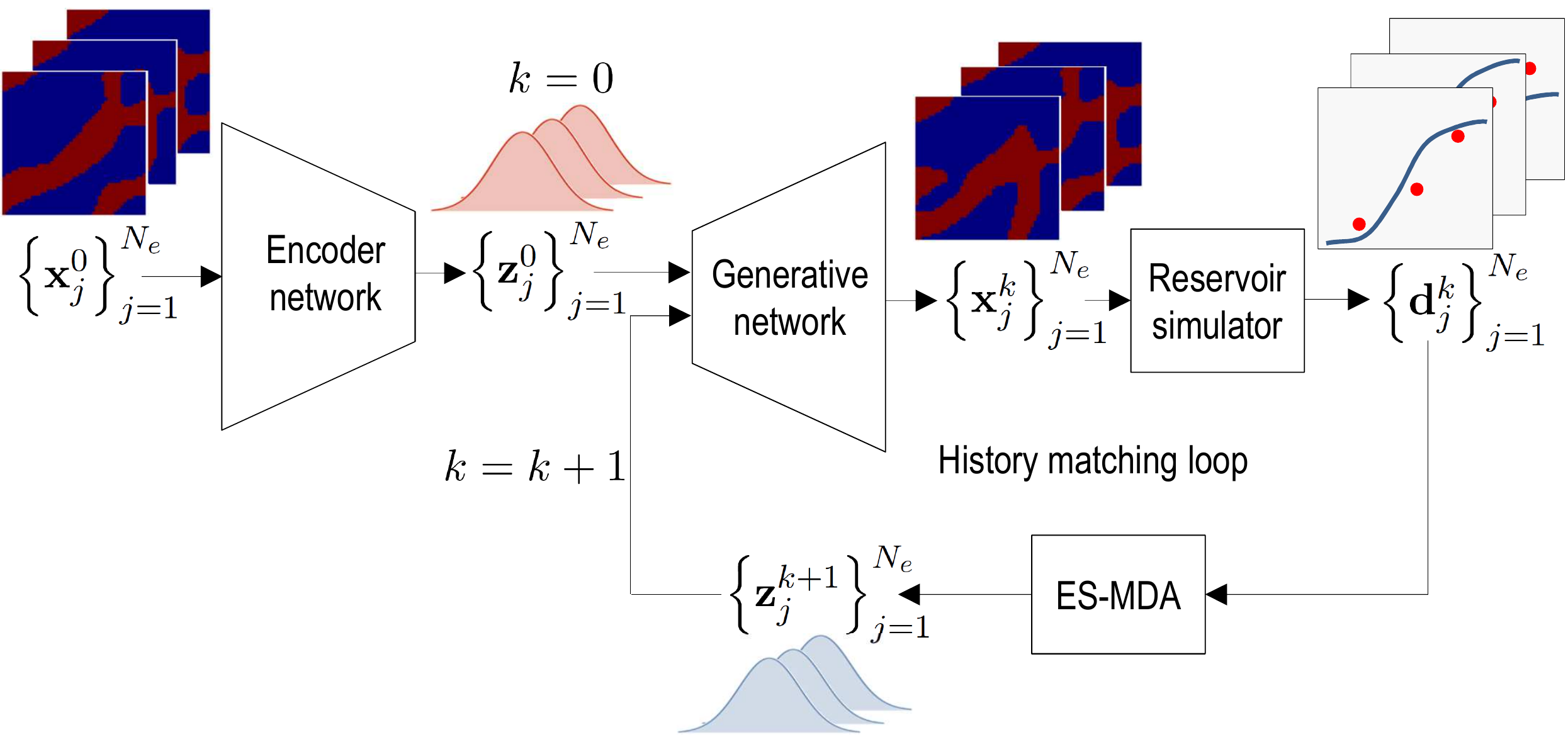}
    \caption{History matching workflow using ES-MDA with deep learning parametrization}
    \label{Fig:DL-ESMDA}
\end{figure}

Fig.~\ref{Fig:DL-ESMDA} shows the overall workflow to combine a trained generative network and ES-MDA for facies history matching. The history matching process starts with a set of prior realizations, $\x_j^0$, of facies and their corresponding latent representation $\z_j^0$ obtained with a trained encoder network. The history matching loop consist of $N_a$ ES-MDA iterations, where the ensemble of $\z_j^k$ is fed to the generative network to generate facies realizations which are used in reservoir simulations to compute the predicted production data $\dsim_j^k$. The ES-MDA analysis is used to update the ensemble of $\z_j^k$, initiating another history matching iteration.

\subsection{Network Architectures}
\label{Sec:Architectures}

In this section, we describe the network architectures tested in this work. All networks are based on convolutional layers \citep{lecun:89} and implemented using \verb"Keras" \citep{keras:15} with \verb"TensorFlow" \citep{tensorflow:15} as the backend engine. The appendix section summarizes the main elements of the networks used in this work. The corresponding codes are available at \href{https://github.com/smith31t/GeoFacies_DL/}{\texttt{github.com/smith31t/GeoFacies\_DL}}.

\subsubsection{VAE}
\label{Sec:ArchVAE}

The first network investigated was based on the VAE as described in \citep{canchumuni:19b}. Table~\ref{Tab:VAE} in the appendix section summarizes the main elements of the network used in this work, which contains three convolutional layers in the encoder, three transposed convolutional layers and one convolutional layer in the decoder. Each input facies realization is transformed into two images, one for each facies type with the value one at the corresponding facies and zero elsewhere. The output of the decoder is a facies image with the facies type decided by the highest activation value computed with a sigmoid function, which is interpreted as the facies with higher probability. During training, we use the Adam optimizer \citep{kingma:14a} with initial learning rate of 0.0001. The stop criteria used the ``EarlyStopping'' function from \verb"Keras" monitoring the loss function and stopping the training when no improvement is observed after ten epochs (patience parameter of the EarlyStopping function).

\subsubsection{GAN}
\label{Sec:ArchGAN}

In order to make fair comparisons between the generative models it is important to be able to start the history matching of all cases with the same ensemble of prior realizations. For some networks, such as VAE this is straightforward because we can fed the prior ensemble to the trained encoder and obtain the initial ensemble of latent vectors to start the history matching as described in the workflow of Fig.~\ref{Fig:DL-ESMDA}. However, other generative models such as GAN does not have an encoder network. In a standard GAN the training starts by sampling direct $\z$ from a designed distribution. For this reason, we introduced an encoder network, which is trained after the adversarial training. The training of this encoder is done using the generative network of the trained GAN as decoder. During the training of the encoder the parameters of the generator are not optimized. Table~\ref{Tab:GAN} in the appendix section summarizes the main elements of the encoder and the GAN network used in this work. The encoder network has four convolutional layers and a batch-normalization layer to normalize the entries of the latent vector to have zero mean and unity variance. The generator network has three transposed convolutional layers and one convolutional layer to construct the final image. The discriminator also uses four convolutional layers followed by a fully-connected with sigmoid activation, which used to classify the images during training. Note that after training, only the encoder and the generative network are used in the history matching workflow indicated in Fig.~\ref{Fig:DL-ESMDA}.

Similarly to the VAE, the training data were pre-processed in two color channels, one for each facies type. However, the values were transformed to one at the current facies and minus one elsewhere. The output of the generator selects the facies type based on the tanh activation function. The termination of the training process of GANs is often done by visual inspection of the generated images. Here, we introduced an alternative procedure to monitor the reconstruction error of the generator. In this procedure, we select a set of validation samples which are not part of the training set. Then, after each training epoch, we use the backpropagation algorithm to find the latent vectors $\z$ corresponding to the validation samples. During this step the parameters of the generator are kept constant as the optimization is carried out over $\z$ to minimize the mean square error. Besides the reconstruction error, we also evaluate the quality of generated samples using the discriminator. Finally, we select the generator model with the higher reconstruction accuracy and with the lower discriminator loss. We only accept a generator with a minimum accuracy of 90\%. We use the Adam optimizer with constant learning rates of 0.0002 for the generator and discriminator networks and 0.001 for the encoder.

\subsubsection{WGAN}
\label{Sec:ArchWGAN}

The architecture of the WGAN is essentially the same used in the GAN (Table~\ref{Tab:GAN}), including the use of an encoder to sample $\z$. The only difference is that last fully-connected layer of discriminator network uses a linear activation function instead of the sigmoid. The objective is make the network to work as a critic instead of a discriminator during the minimization of the WGAN loss function. Unlike the GAN, the stop criteria for the WGAN was based on the convergence during the minimization of the loss function. We also monitor a reconstruction accuracy with a validation set following the same procedure used for the GAN to ensure a minimum value of 90\%. We use the RMSprop \citep{hinton:14a} optimizer as suggested in the original paper with learning rate of 0.0005 for the generator and critic network, and 0.001 in the encoder.

\subsubsection{$\alpha$-GAN}
\label{Sec:ArchAlphaGAN}

The $\alpha$-GAN network already has an encoder network which is trained in the same process of the generator and two discriminators. Table~\ref{Tab:alpha-GAN} summarizes the components of these networks, which are very similar to the networks used in the standard GAN, including the pre-processing of facies images and stop criteria based on the reconstruction accuracy for a validation set. We use the Adam optimizer with 0.0001 of learning rate for all networks with a reconstruction weights of 100.

\subsubsection{PCA-Cycle-GAN}
\label{Sec:ArchPCA-Cycle-GAN}

Similarly to transfer style networks, Cycle-GAN can be used to generate images combining the content of one image with the style of another. Here, we follow the ideas from \citep{liu:19a} and use the PCA coefficients of facies realizations as content images and actual facies realizations as style images. In this case, the first step of the training process is to perform a PCA with a set of facies realizations. Here, we use the symmetric square root scheme described in \citep{emerick:17a} to generate realizations of $\z$ with the same dimension of realizations of facies $\x$. PCA is also used to generate the initial ensemble of $\z$ for history matching. The Cycle-GAN is composed of two generators that contains residuals blocks (Table~\ref{Tab:Residual-Block}) and two discriminators, whose components are summarized in Table~\ref{Tab:PCA-Cycle-GAN} in the appendix. The generators have an structure that resembles autoencoders because they have input and output images with the same dimensions. The stop criteria was also based on visual inspection and reconstruction accuracy for a validation set. We use the Adam optimizer with 0.0002 of learning rate for all networks. Cycle-GAN usually are trained with small bath size, here used 4 and with a weight of the cycle loss of 10.

\subsubsection{PCA-Style}
\label{Sec:ArchAlphaPCA-Style}

\begin{figure}[h!]
\centering
\includegraphics[width=0.75\textwidth]{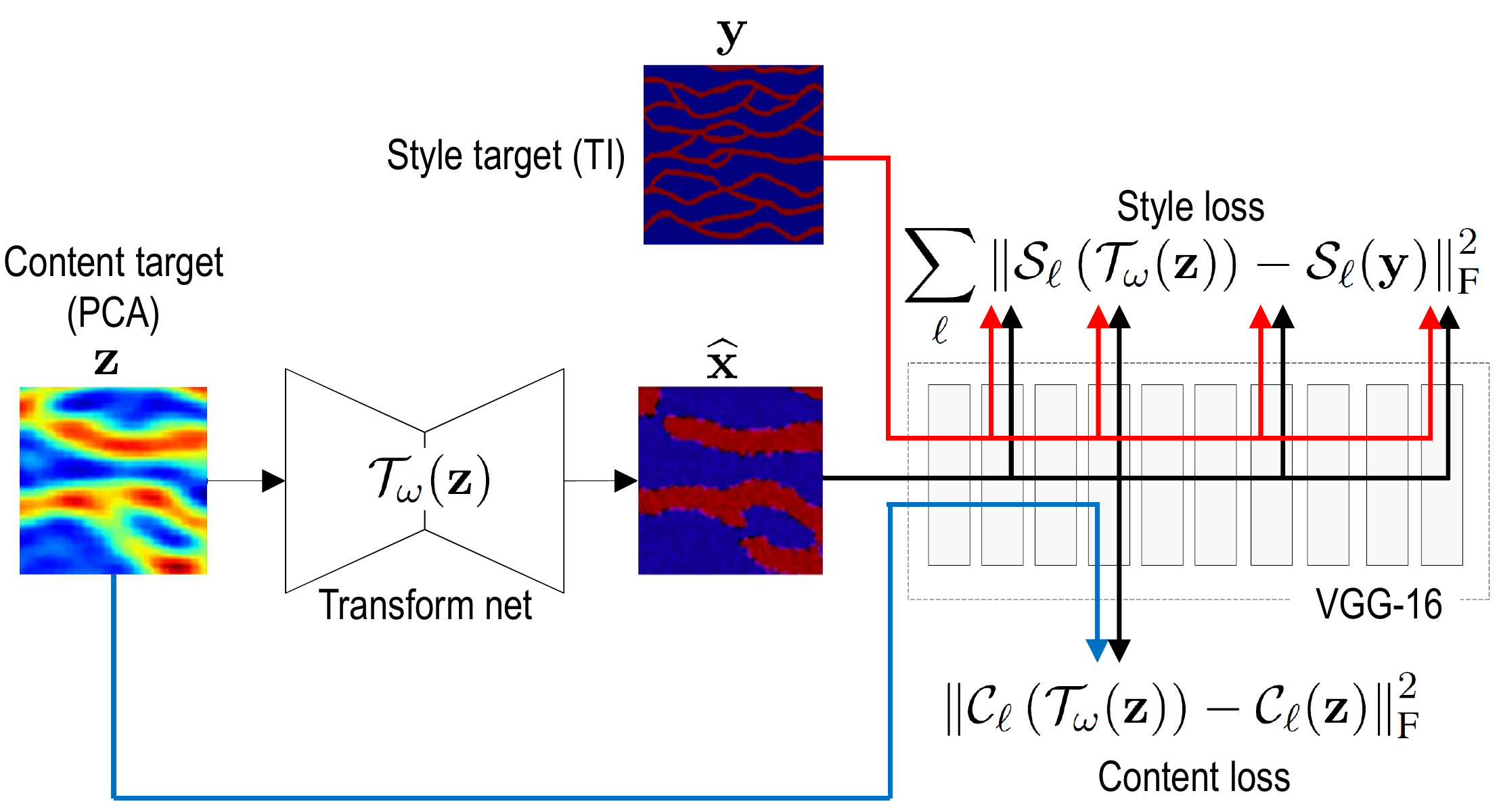}
\caption{Schematic architecture for training the transform network used in the PCA-Style}
\label{Fig:PCA-Style}
\end{figure}

Here, we use the same ideas presented in \citep{liu:19a} which introduced a transform network $\mathcal{T}_\omega(\z)$ trained with a set of random PCA realizations. The objective is to avoid solving the minimization of $\mathcal{L}_{\textrm{total}}(\x,\y,\z)$ (Eq.~\ref{Eq:TransferLoss}) of the transfer style network for reconstructing the facies $\x$ during history matching. Table~\ref{Tab:Style} (appendix) shows the architecture of the transform network used in this work. The training of $\mathcal{T}_\omega(\z)$ uses a pre-trained VGG-16 network to compute the content and style losses, similarly to transfer style networks. However, instead of finding for a optimal reconstruction of a particular realization, the network is trained to generate facies with small content and style losses. Fig.~\ref{Fig:PCA-Style} illustrates the process. After trained, $\mathcal{T}_\omega(\z)$ is used as a generative network in the history-matching process depicted in Fig.~\ref{Fig:DL-ESMDA}. The loss function minimized during training has the form

\begin{equation}\label{Eq:TransferLossPCA}
  \mathcal{L}_{\textrm{PCA-Style}}(\omega) =  c_1 \left\| \mathcal{C}_\ell\left(\mathcal{T}_\omega(\z) \right) - \mathcal{C}_\ell(\z) \right\|_{\textrm{F}}^2 + c_2 \sum_{\ell} \left\| \mathcal{S}_\ell\left(\mathcal{T}_\omega(\z) \right) - \mathcal{S}_\ell(\y) \right\|_{\textrm{F}}^2.
\end{equation}

Unlike the previous networks, the input images in this case correspond to PCA realizations such that there is no pre-processing dividing the image in color channels. In the output images we have values in the range [-1, 1]. In order to obtain binary facies, we apply a truncation threshold of 0, which represent the  $50\%$ of the domain for each facies. In the training process, we used the 100 epochs as stop criteria.

\subsubsection{VAE-Style}
\label{Sec:ArchAlphaVAE-Style}

Inspired by the PCA-Style implementation of \citet{liu:19a}, we also investigate the use of a VAE-Style network. The resulting model is a hybrid generative model that uses the probabilistic formulation of VAE and the style losses computed in the pre-trained VGG-16 network. The motivation is to improve the quality of generated VAE images using transfer style ideas. The loss function minimized during training combines the reconstruction error and the KL divergence of the VAE loss and the style loss of the transfer style network. Additionally, we include a variation loss, $\mathcal{L}_{\textrm{var}}(\theta)$ defined as the sum of the absolute difference for neighboring pixel-values in the input images. This loss is designed to measure the amount of noise in the facies realization. The final loss function minimized during training is

\begin{equation}\label{Eq:VAE-Style}
  \mathcal{L}_{\textrm{VAE-Style}} (\theta, \eta) =  \mathcal{L}_{\textrm{RE}_{MSE}}(\theta) + \textrm{KL} \left( p_\eta(\z|\x) \| p(\z) \right) + c_1 \sum_{\ell} \left\| \mathcal{S}_\ell\left(\widehat{\x} \right) - \mathcal{S}_\ell(\y) \right\|_{\textrm{F}}^2 + c_2 \mathcal{L}_{\textrm{var}}(\theta),
\end{equation}
where $c_1$ and $c_2$ are constant defining the relative weights of the style and variation losses. After training, the decoder of the VAE is used as generative network in the history matching process (Fig.~\ref{Fig:DL-ESMDA}).

\begin{figure}[h!]
\centering
\includegraphics[width=0.75\textwidth]{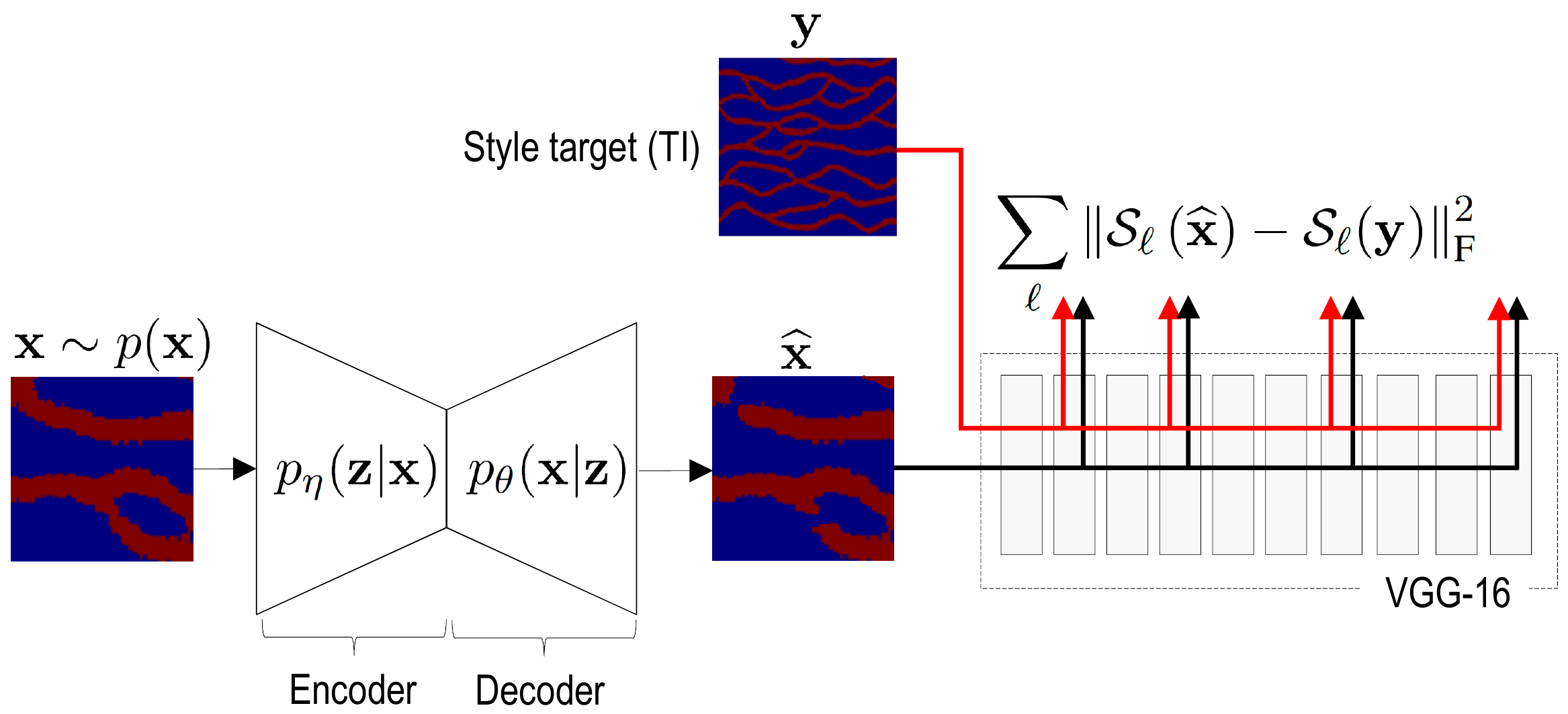}
\caption{Schematic architecture for training VAE-Style network}
\label{Fig:VAE-Style}
\end{figure}

The VAE architecture used here is similar to the one presented in Table~\ref{Tab:VAE} with two differences: (1) the input dimension is ($60\times60\times 1$) instead of (60~$\times$~60~$\times$~2), and (2) the last activation function of the decoder network is tanh instead of the sigmoid function. These modification were required to adapt the images to the VGG-16 network such that the backpropagation algorithm can be applied, which is not possible with the binary outputs of our VAE. In order to obtain binary facies, we use a truncation threshold of 0.

\section{Test Case 1: Comparison of Network Formulations}
\label{Sec:Case1}

\begin{figure}[!h]
    \centering
    \includegraphics[width=0.22\textwidth]{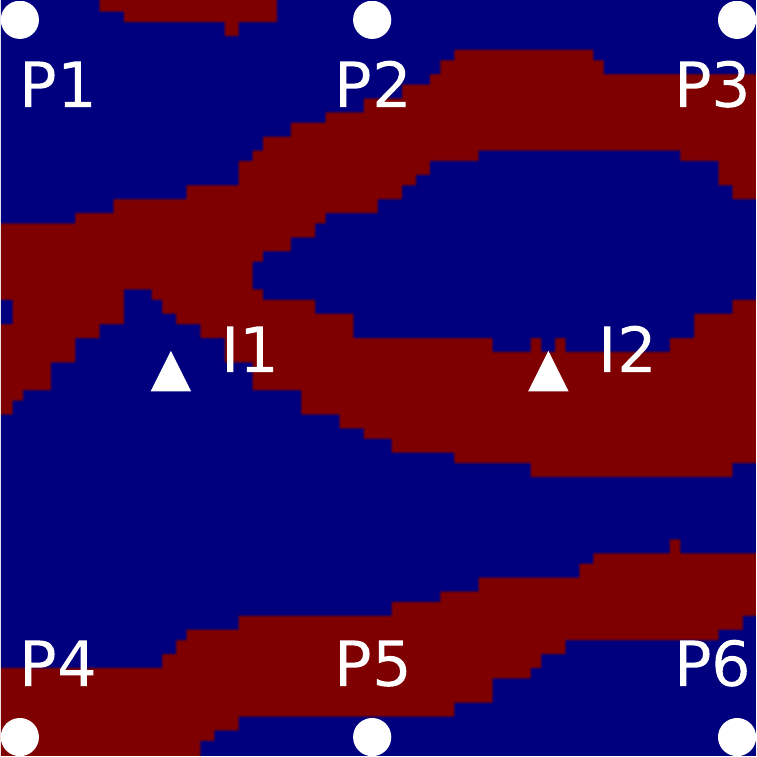}
    \caption{Reference facies model. Red colors correspond to high-permeability channels and blue the background sand. Case 1.}
    \label{Fig:Case1-true}
\end{figure}

The objective of the first test case is to compare the performance of the different networks as parameterization techniques for facies history matching. The reservoir model corresponds to a small 2D case with $60 \times 60$ gridblocks of constant size of 100~m~$\times$~100~m with constant thickness of 25~m. A reference model (Fig.~\ref{Fig:Case1-true}) with two facies (channel and background) was generated using the MPG algorithm \verb"snesim" with the well-known channel training image presented in~\citep{caers:04}. We adopted a constant permeability value for each facies; 1000~mD for the channels and 100~mD for the background sand. All prior realizations used for training the networks and for history matching were generated using the \verb"snesim" algorithm with the same parameters of the reference model. All initial realizations were not conditioned to facies data (hard data) at well locations to make the problem of estimating the correct position of the channels more challenging. We placed six oil producing and two water injection wells symmetrically distributed as shown in  Fig.~\ref{Fig:Case1-true}.

\subsection{Training Process}
\label{Sec:TranningCase1}

All neural networks were trained with the same dataset corresponding to 20000 facies realizations divided in 70\% for training and 30\% for validation. The PCA required for the networks PCA-Style and PCA-Cycle-GAN used 3000 facies realizations to construct the covariance matrix. The size of the $\z$ vector used in the networks is $N_z =500$, except for the PCA-Cycle-GAN and PCA-Style, which use $\z$ with the same dimension of the model, because in these cases the vector $\z$ represents the ``content image.'' The training was executed in a cluster with four GPUs NVIDIA Tesla V100 of Volta architecture with 16~GB of RAM and 5120 Cuda cores each one. Table~\ref{Tab:Case1-TrainTime} shows the time required for training each network using just one GPU from the cluster. Except for the WGAN, which spent four and half hours, and the PCA-Cycle-GAN with 79 minutes, all other networks required between 30 minutes and one hour to complete training. It is important to note that the current test problem is unrealistically small compared to reservoir models used in practice. In fact, the computational cost of the training process is currently one major limitation of the process.

\begin{table}[ht!]
\caption{Elapsed time for training the networks. Case 1}
\label{Tab:Case1-TrainTime}
\begin{small}
\begin{center}
\begin{tabular}{lc}
\toprule
Network & Time (minutes) \\
\midrule
VAE & 29 \\
GAN & 23 \\
WGAN & 272 \\
$\alpha$-GAN & 59 \\
PCA-Cycle-GAN & 79 \\
PCA-Style & 28 \\
VAE-Style & 54 \\
\bottomrule
\end{tabular}
\end{center}
\end{small}
\end{table}

\begin{figure}
  \centering
  \subfloat[Training]{\includegraphics[width=0.110\linewidth]{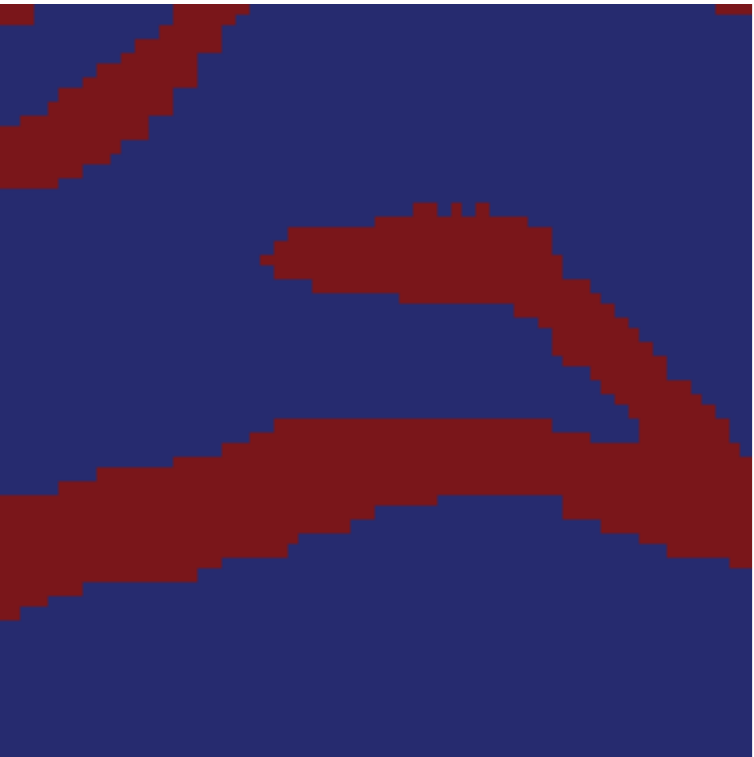}
  \hspace{1mm}\includegraphics[width=0.110\linewidth]{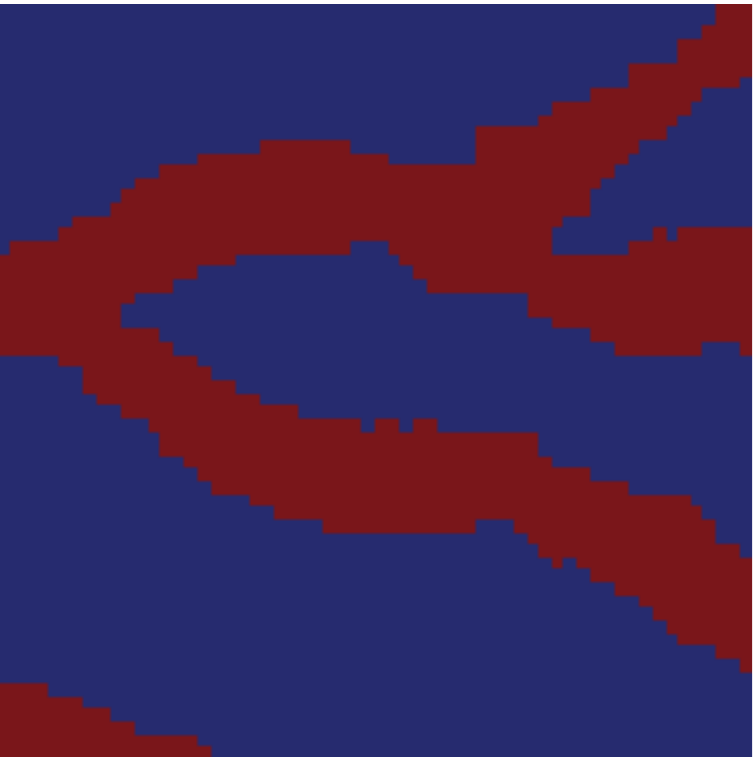}
  \hspace{1mm}\includegraphics[width=0.110\linewidth]{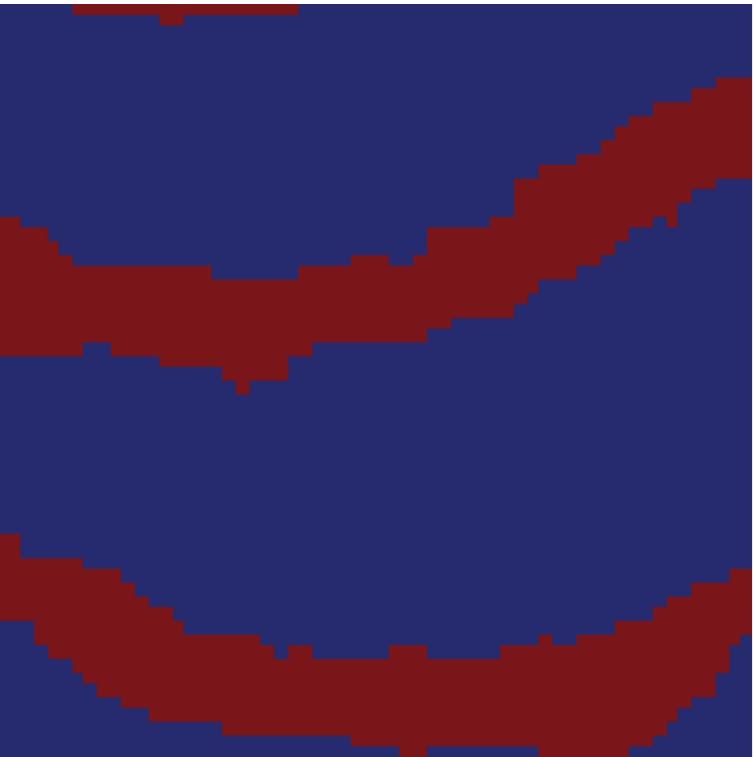}
  \hspace{1mm}\includegraphics[width=0.110\linewidth]{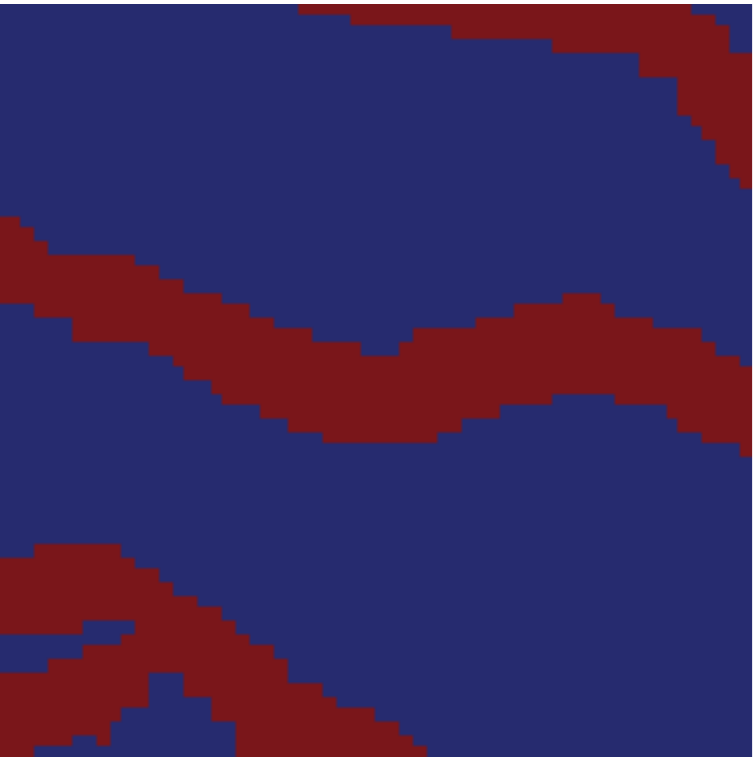} \hspace{1mm}\includegraphics[width=0.110\linewidth]{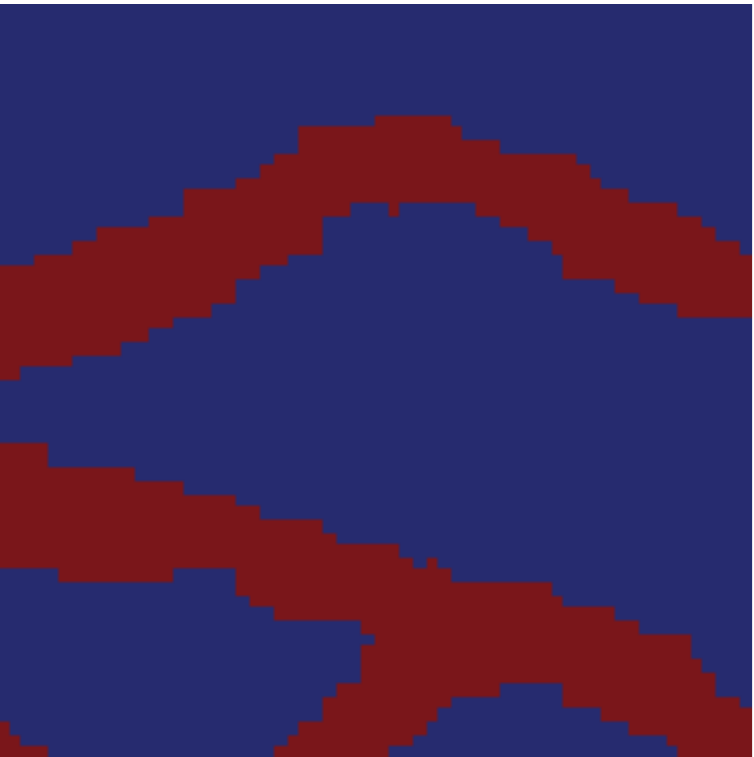}}\\
  \subfloat[VAE]{\includegraphics[width=0.110\linewidth]{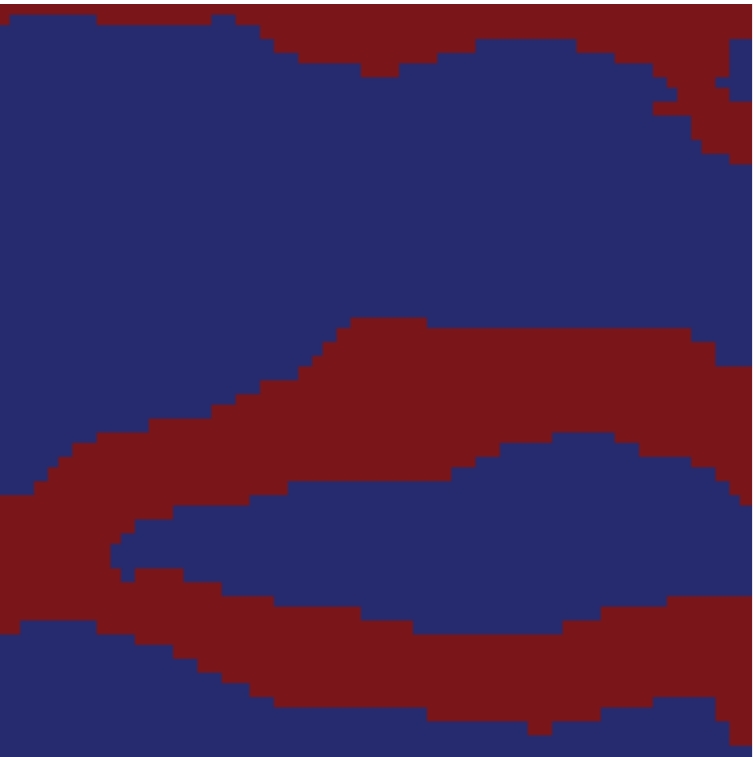}
  \hspace{1mm}\includegraphics[width=0.110\linewidth]{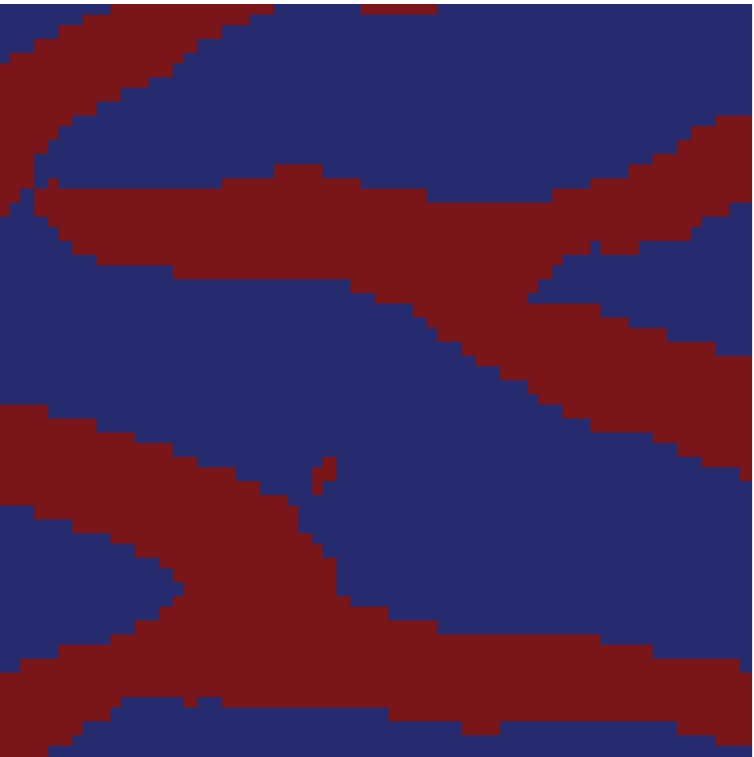}
  \hspace{1mm}\includegraphics[width=0.110\linewidth]{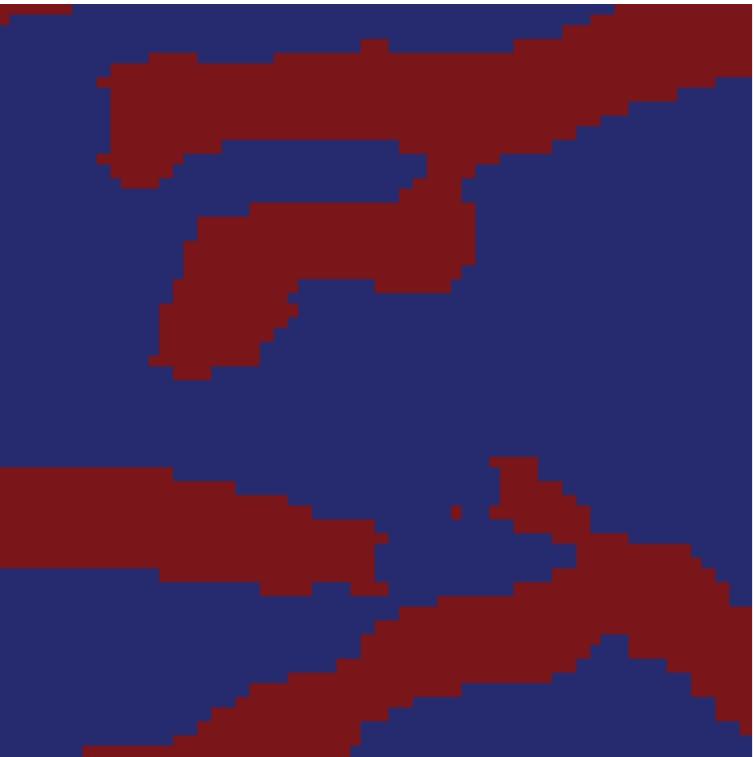}
  \hspace{1mm}\includegraphics[width=0.110\linewidth]{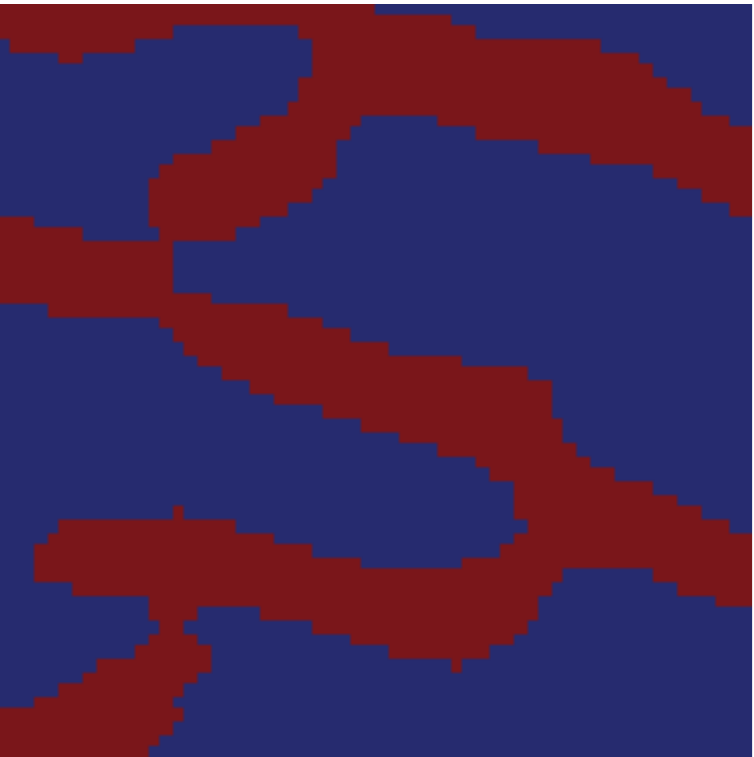}
  \hspace{1mm}\includegraphics[width=0.110\linewidth]{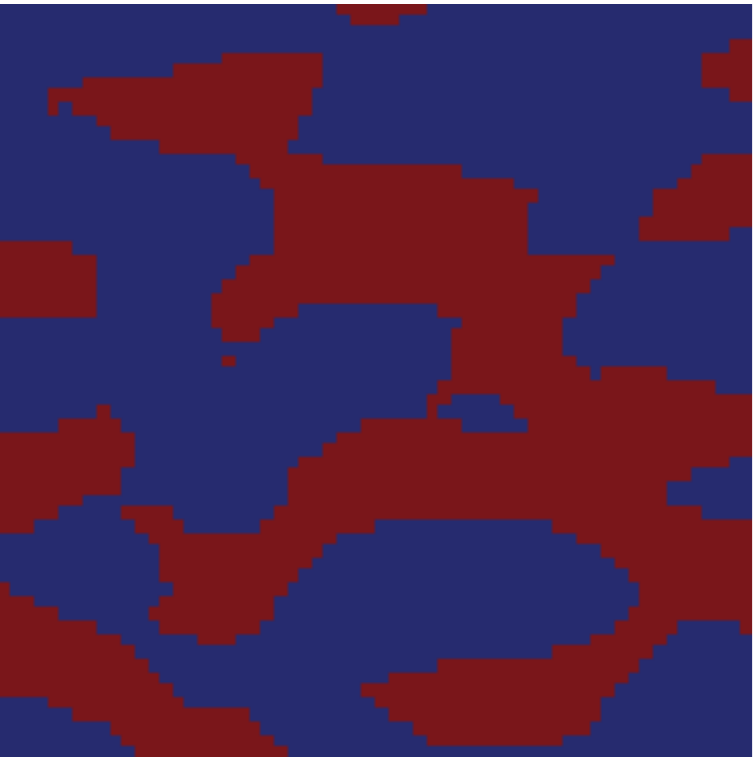}}\\
  \subfloat[GAN]{\includegraphics[width=0.110\linewidth]{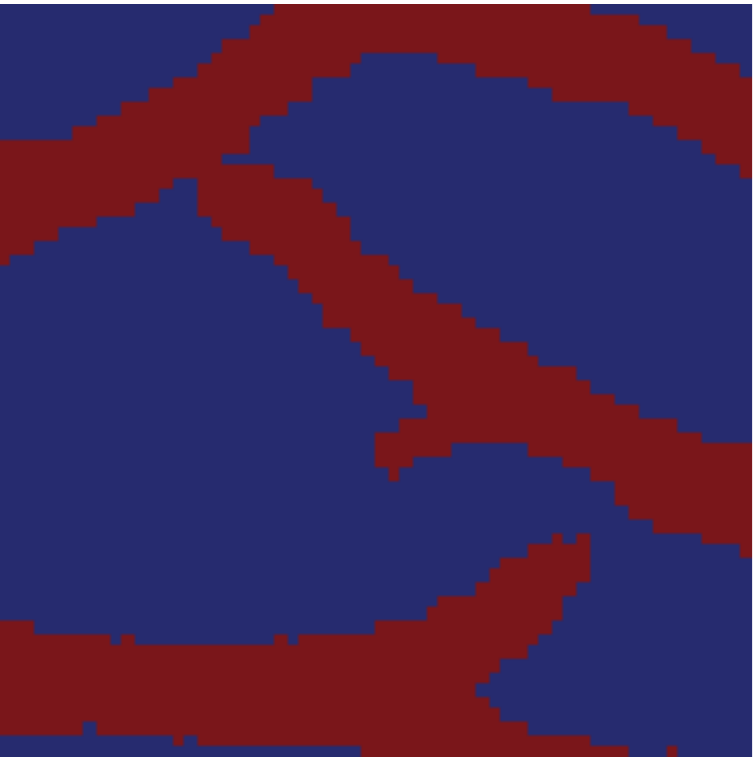}
  \hspace{1mm}\includegraphics[width=0.110\linewidth]{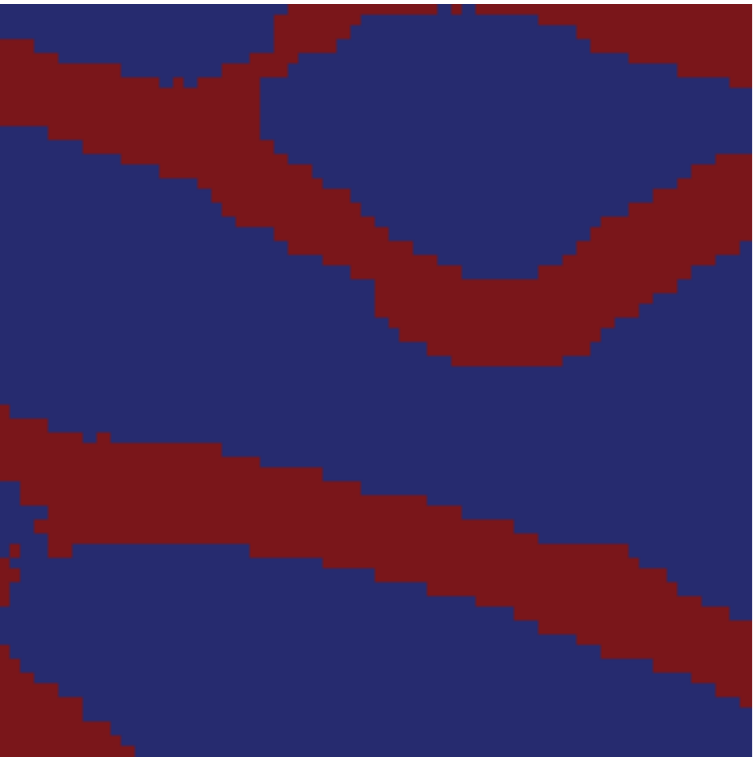}
  \hspace{1mm}\includegraphics[width=0.110\linewidth]{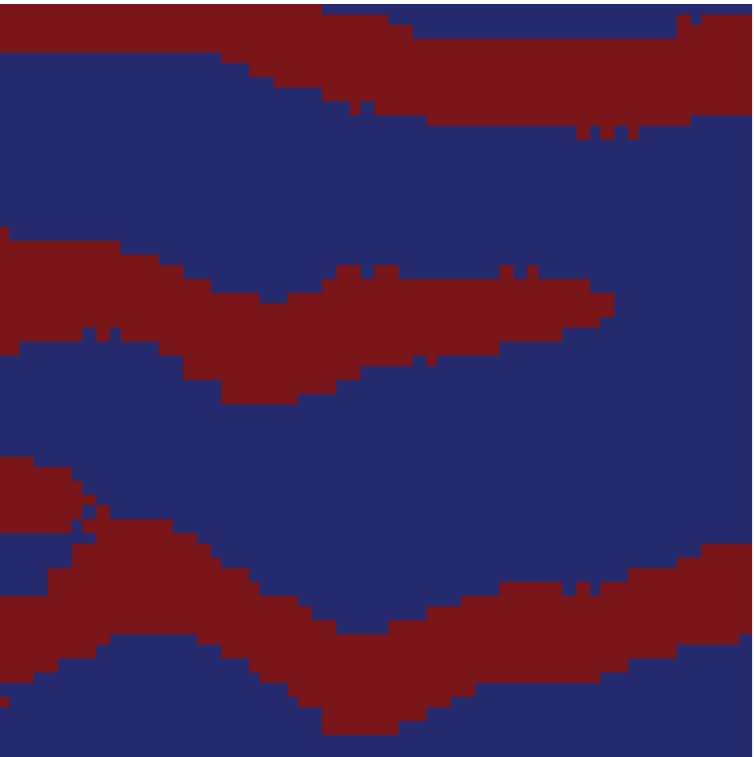}
  \hspace{1mm}\includegraphics[width=0.110\linewidth]{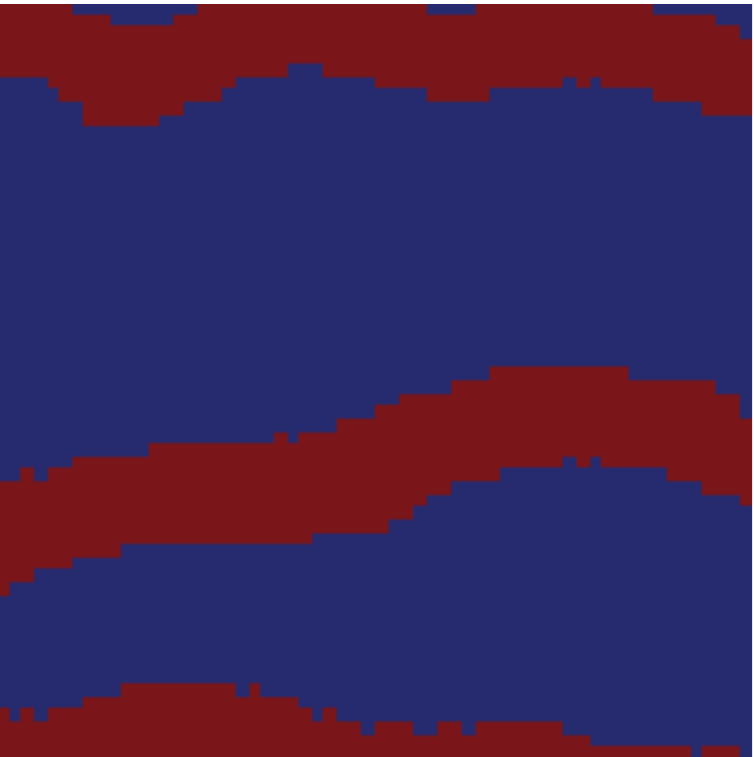}
  \hspace{1mm}\includegraphics[width=0.110\linewidth]{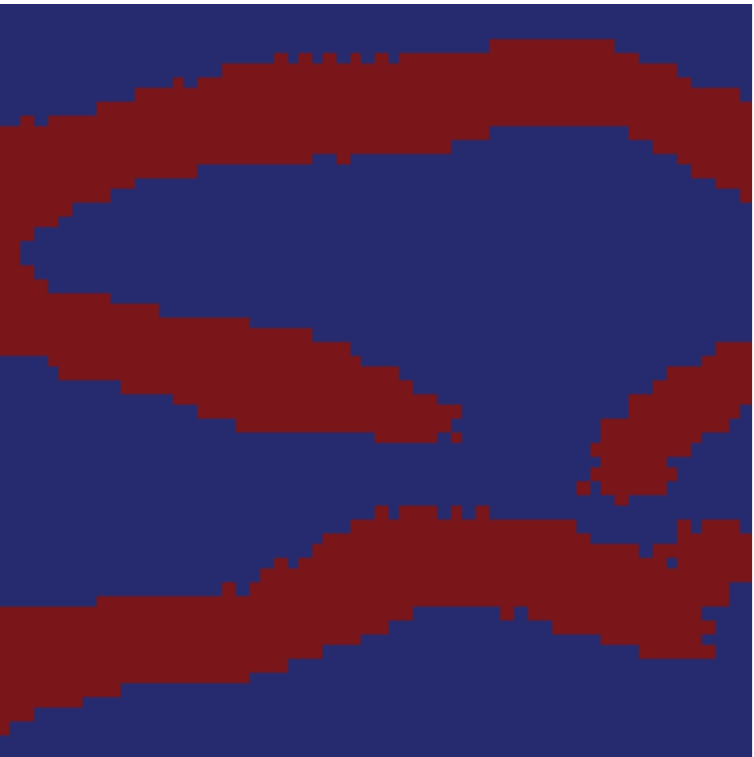}}\\
  \subfloat[WGAN]{\includegraphics[width=0.110\linewidth]{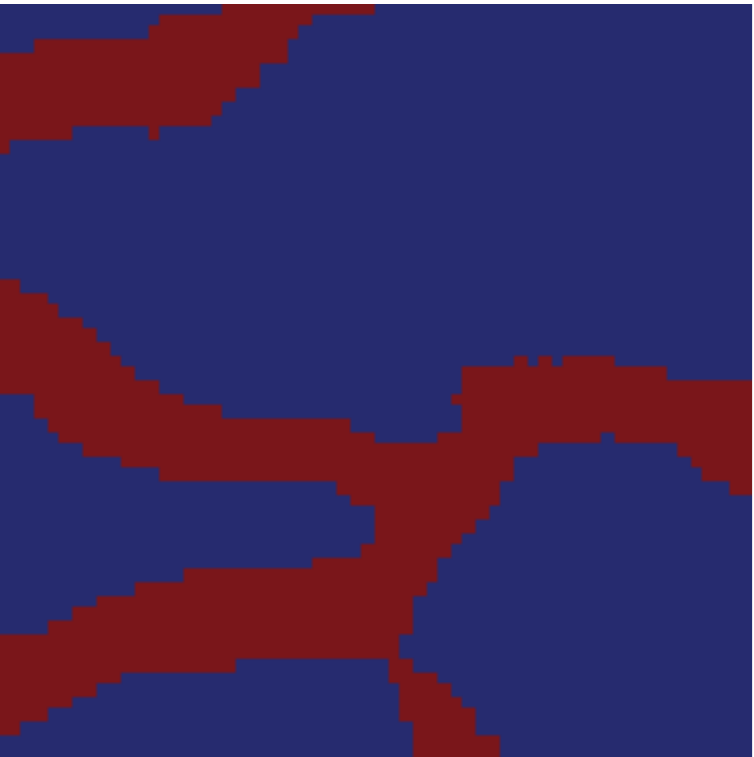}
  \hspace{1mm}\includegraphics[width=0.110\linewidth]{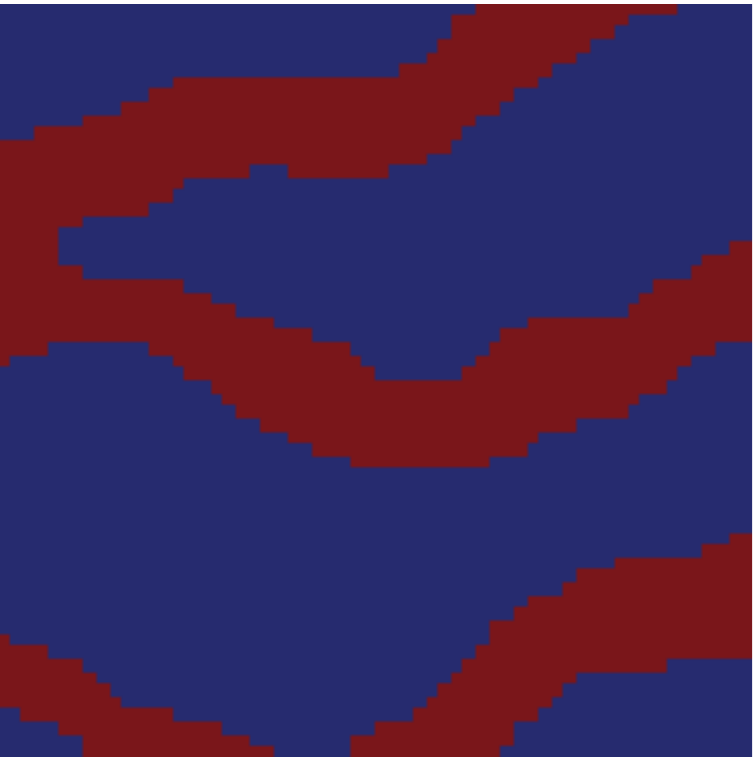}
  \hspace{1mm}\includegraphics[width=0.110\linewidth]{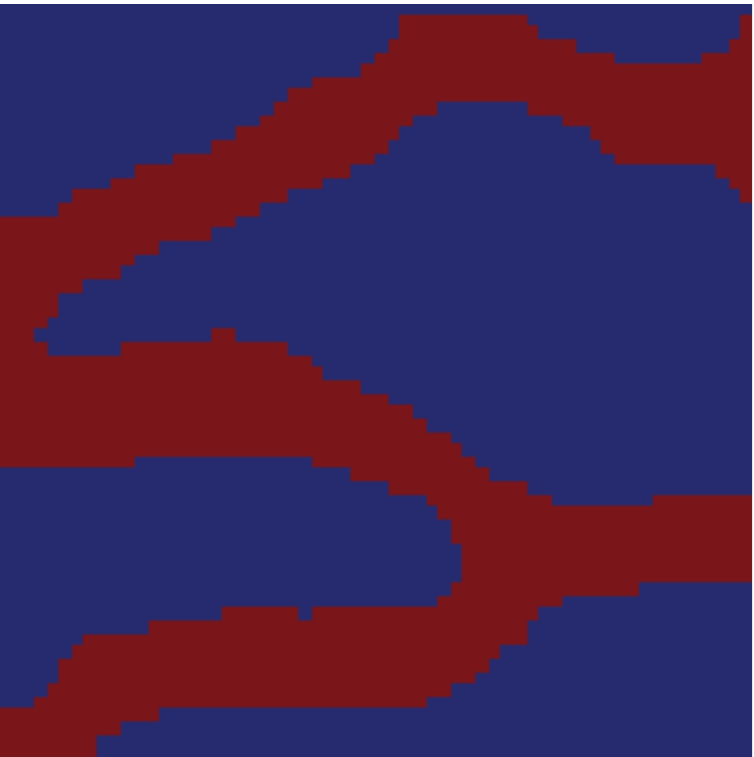}
  \hspace{1mm}\includegraphics[width=0.110\linewidth]{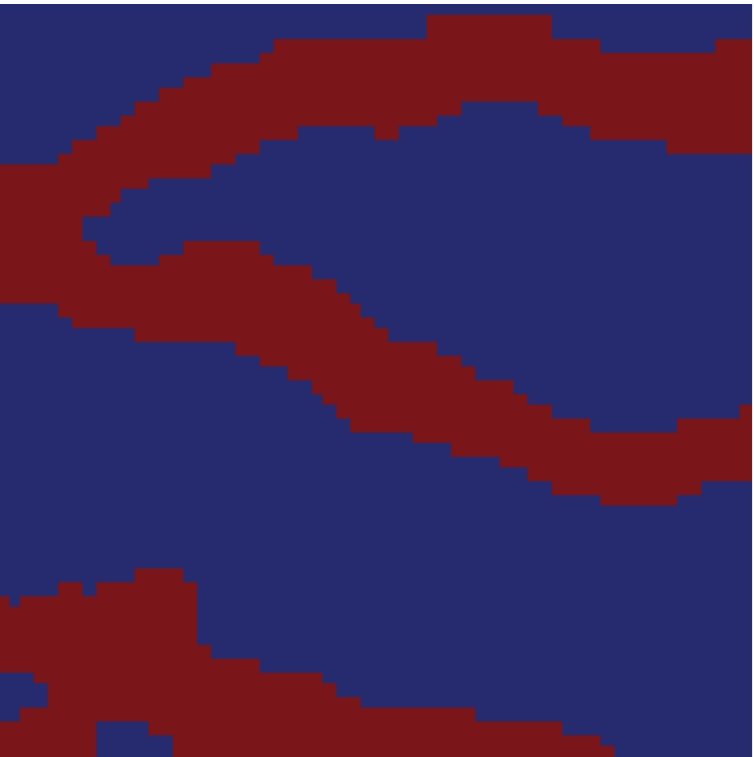}
  \hspace{1mm}\includegraphics[width=0.110\linewidth]{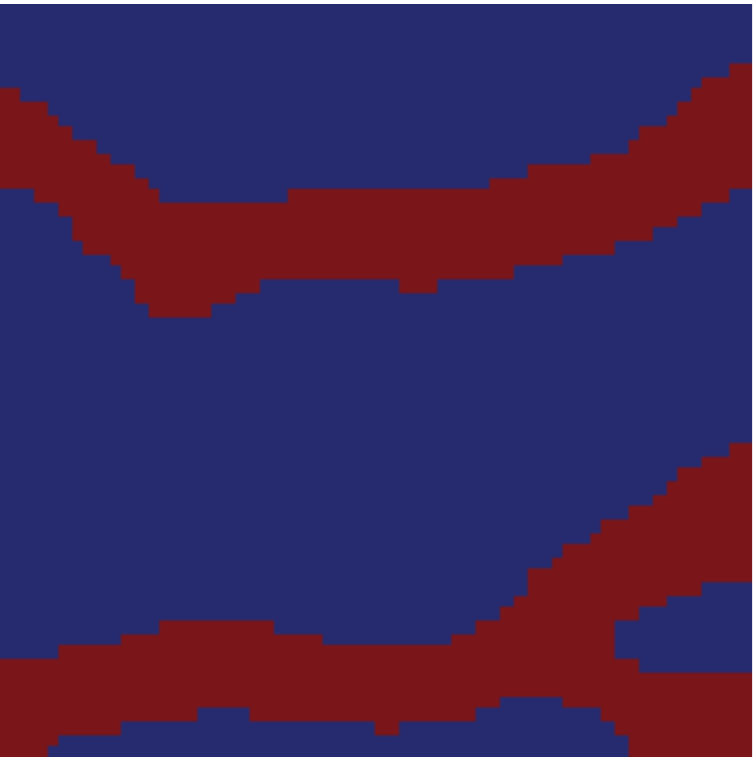}}\\
  \subfloat[$\alpha$-GAN]{\includegraphics[width=0.110\linewidth]{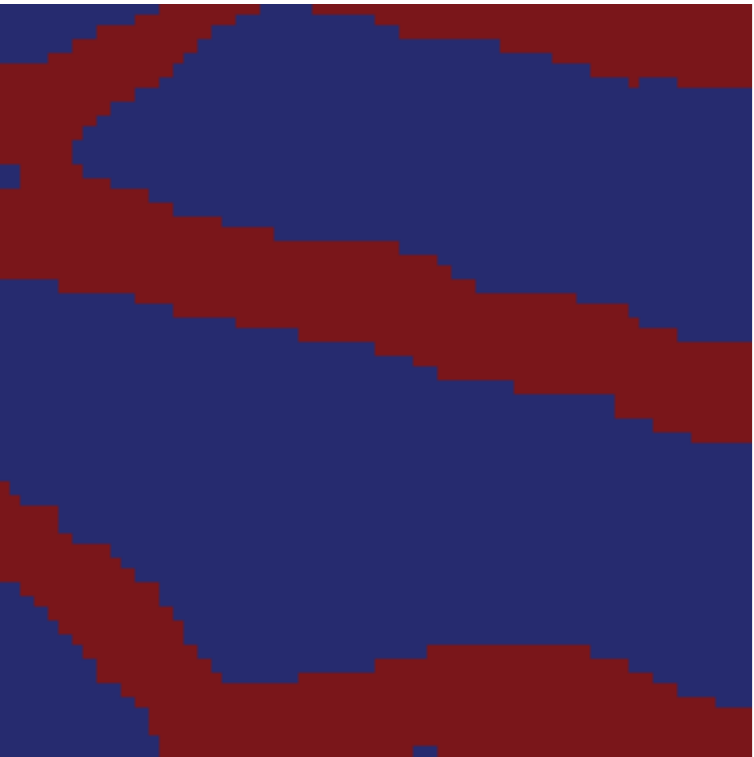}
  \hspace{1mm}\includegraphics[width=0.110\linewidth]{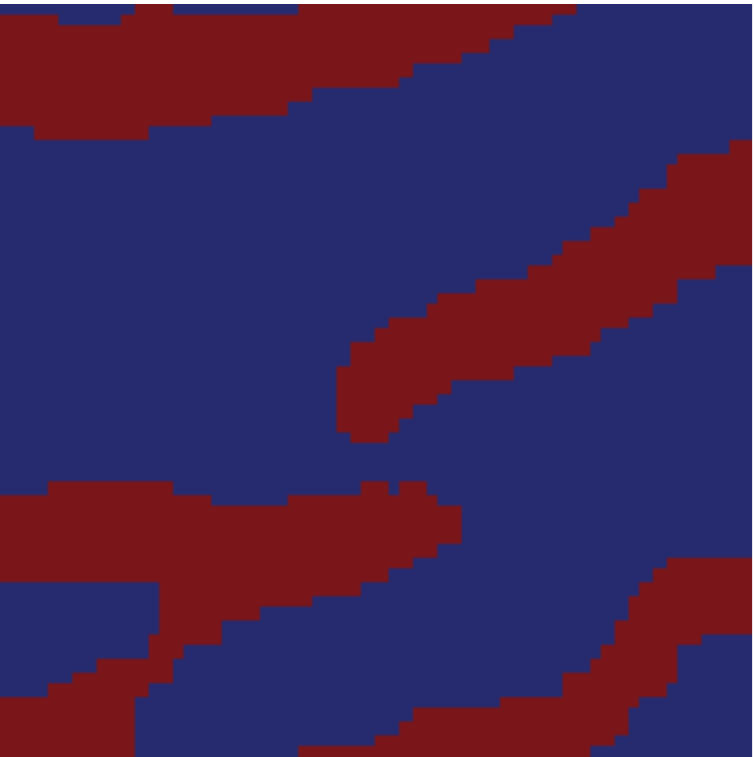}
  \hspace{1mm}\includegraphics[width=0.110\linewidth]{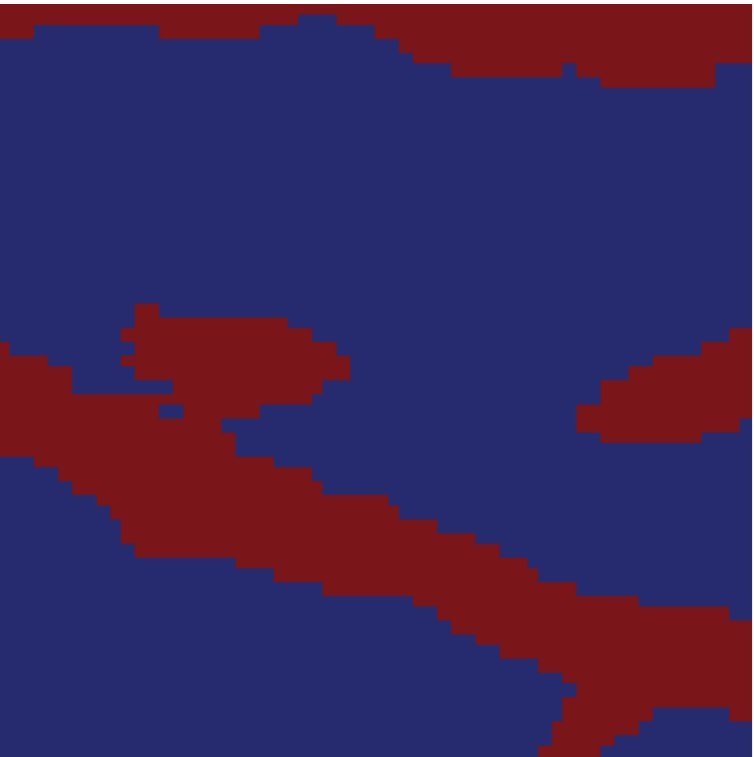}
  \hspace{1mm}\includegraphics[width=0.110\linewidth]{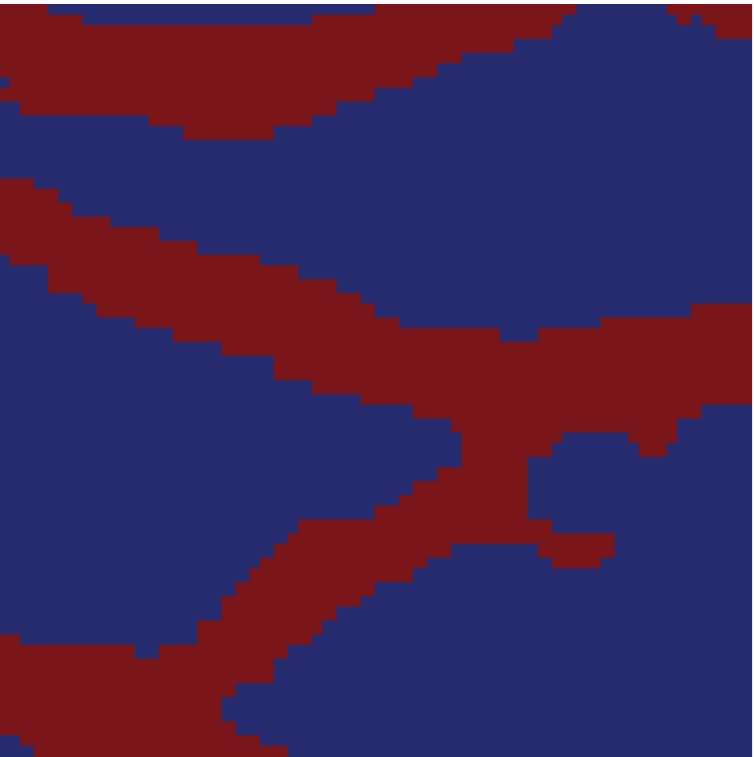}
  \hspace{1mm}\includegraphics[width=0.110\linewidth]{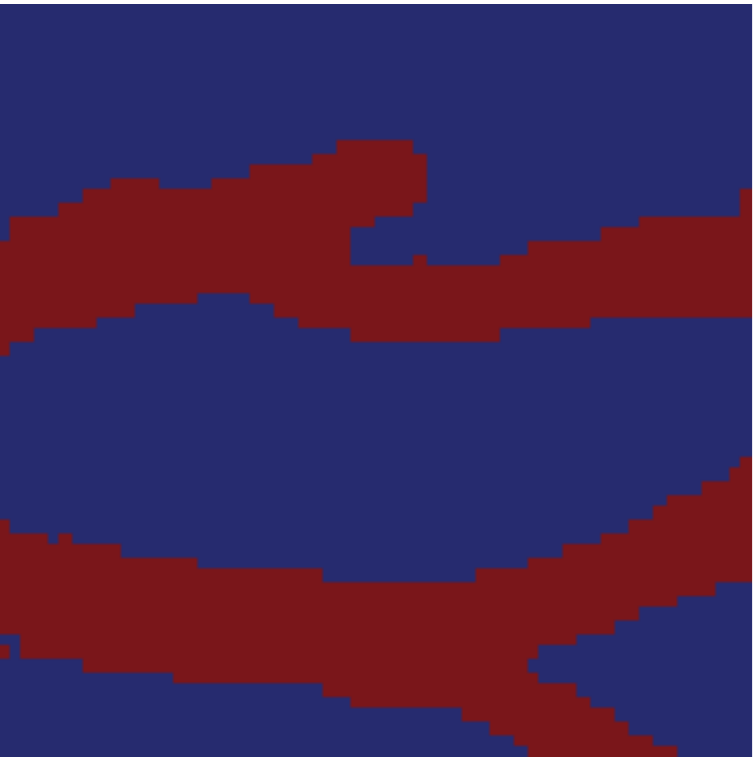}}\\
  \subfloat[PCA-Cycle-GAN]{\includegraphics[width=0.110\linewidth]{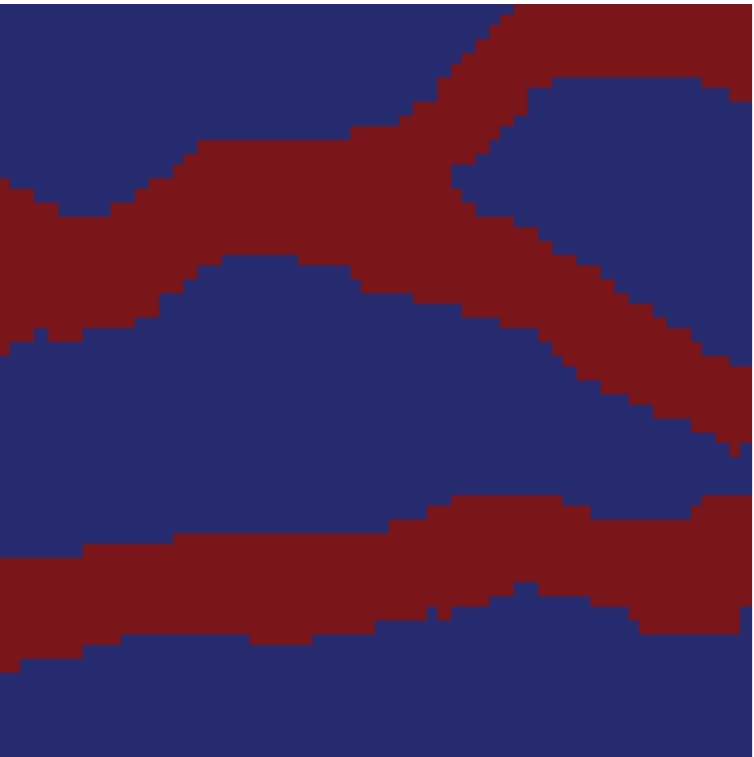}
  \hspace{1mm}\includegraphics[width=0.110\linewidth]{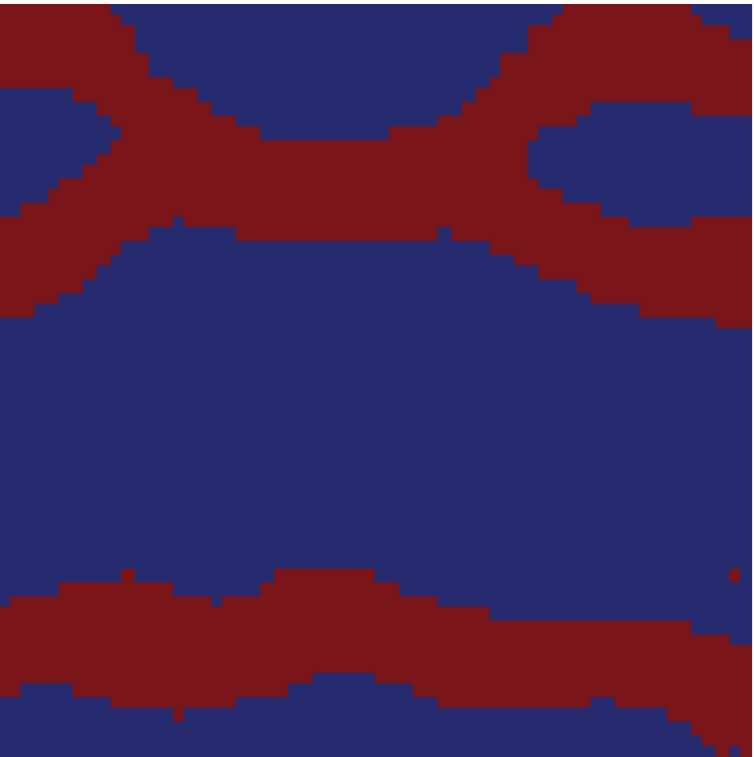}
  \hspace{1mm}\includegraphics[width=0.110\linewidth]{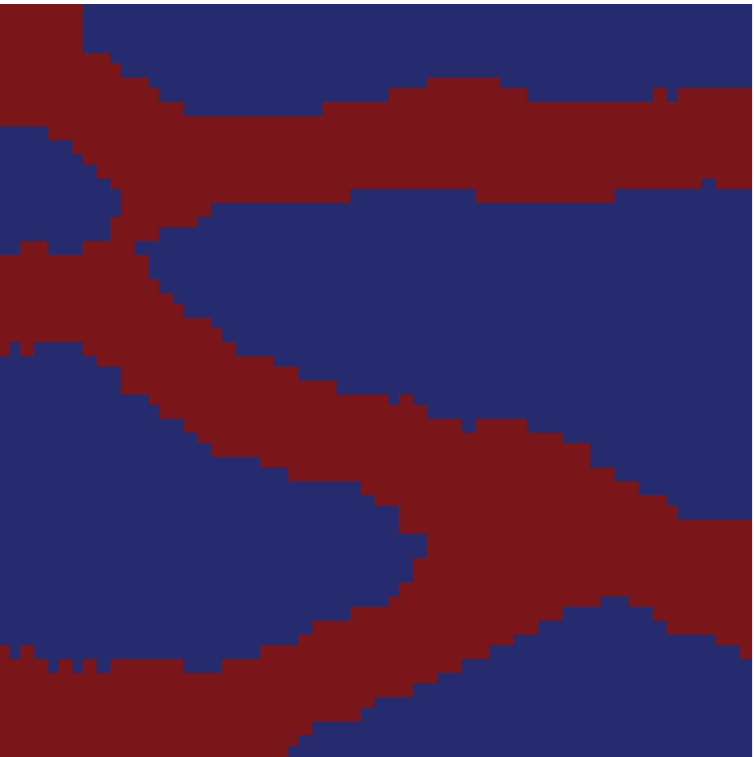}
  \hspace{1mm}\includegraphics[width=0.110\linewidth]{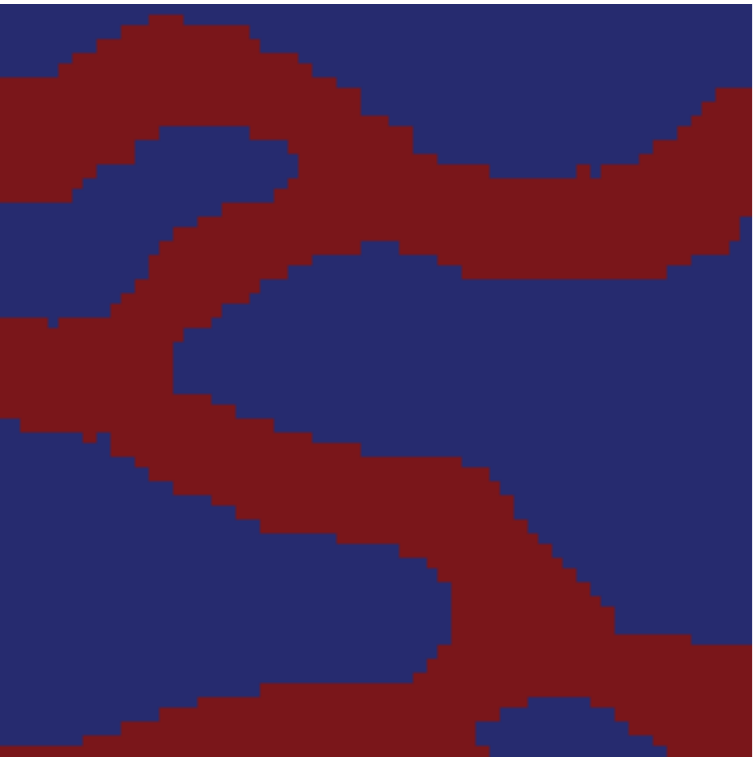}
  \hspace{1mm}\includegraphics[width=0.110\linewidth]{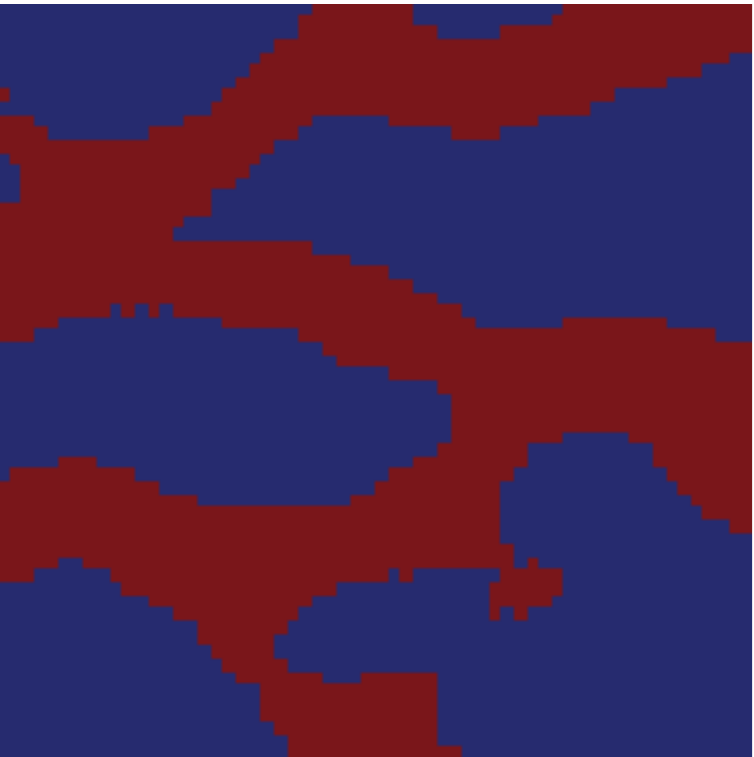}}\\
  \subfloat[PCA-Style]{\includegraphics[width=0.110\linewidth]{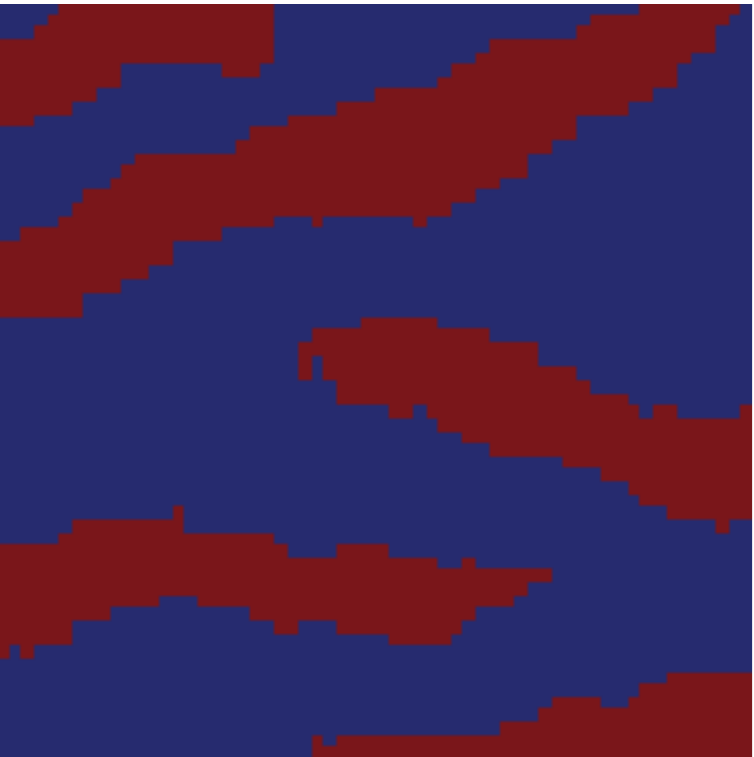}
  \hspace{1mm}\includegraphics[width=0.110\linewidth]{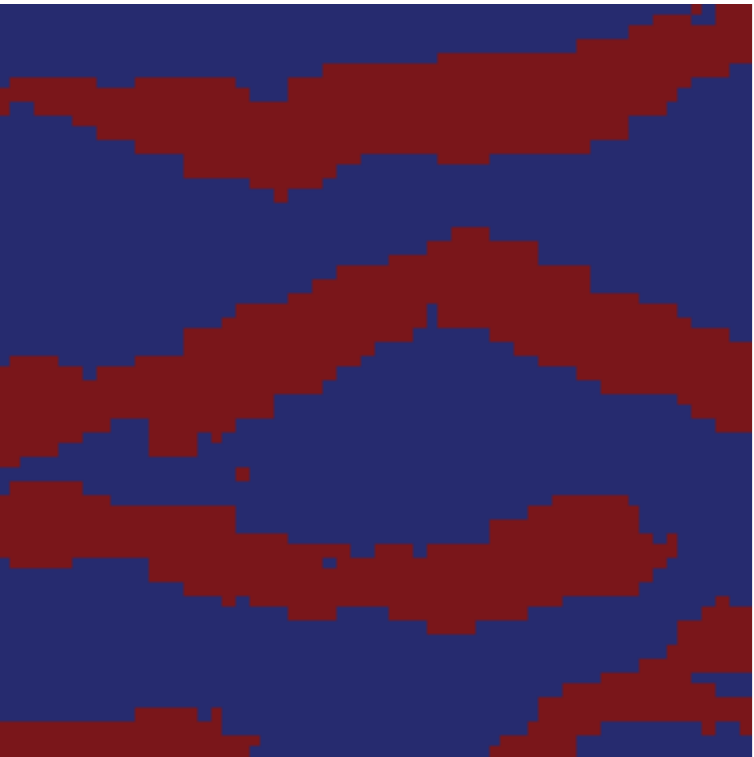}
  \hspace{1mm}\includegraphics[width=0.110\linewidth]{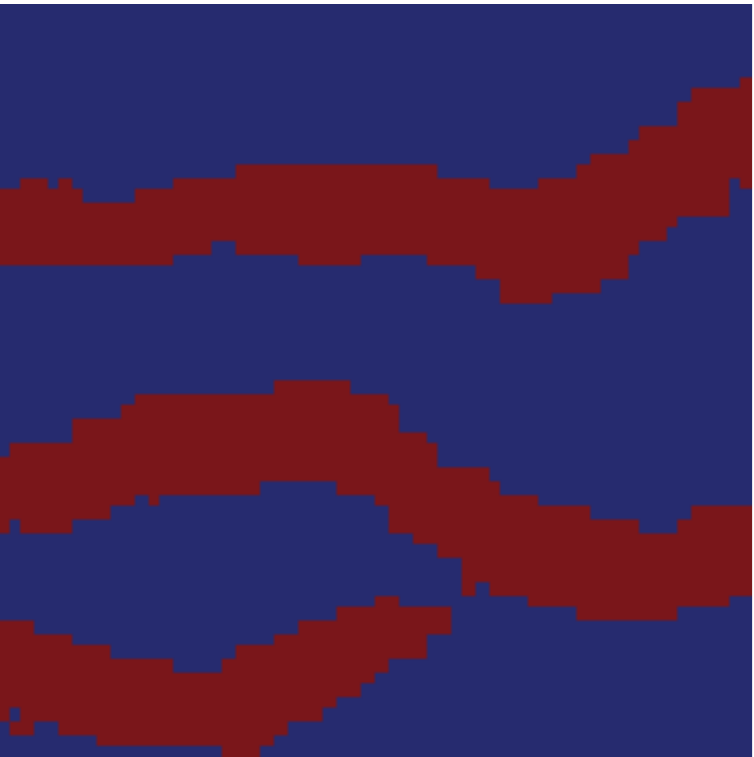}
  \hspace{1mm}\includegraphics[width=0.110\linewidth]{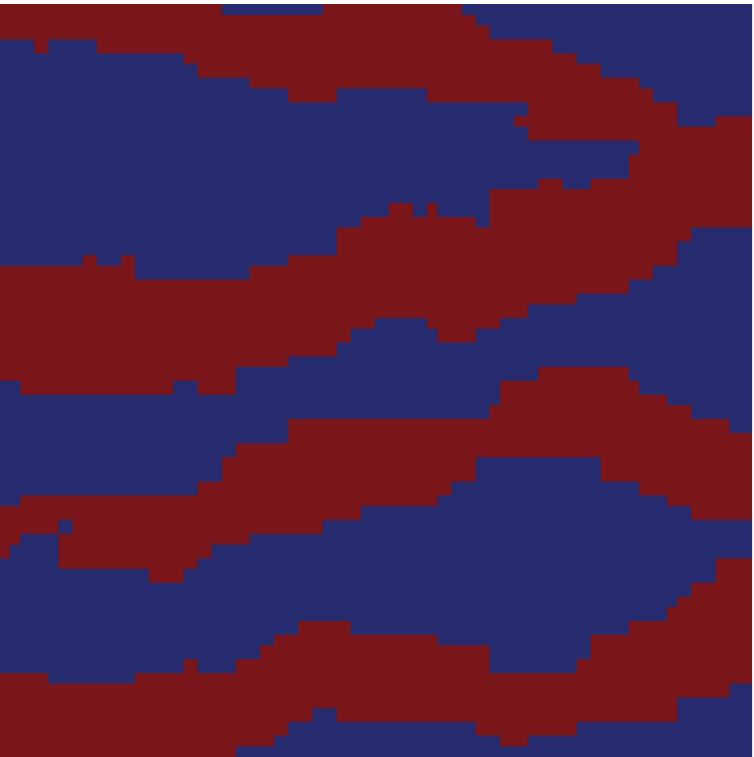}
  \hspace{1mm}\includegraphics[width=0.110\linewidth]{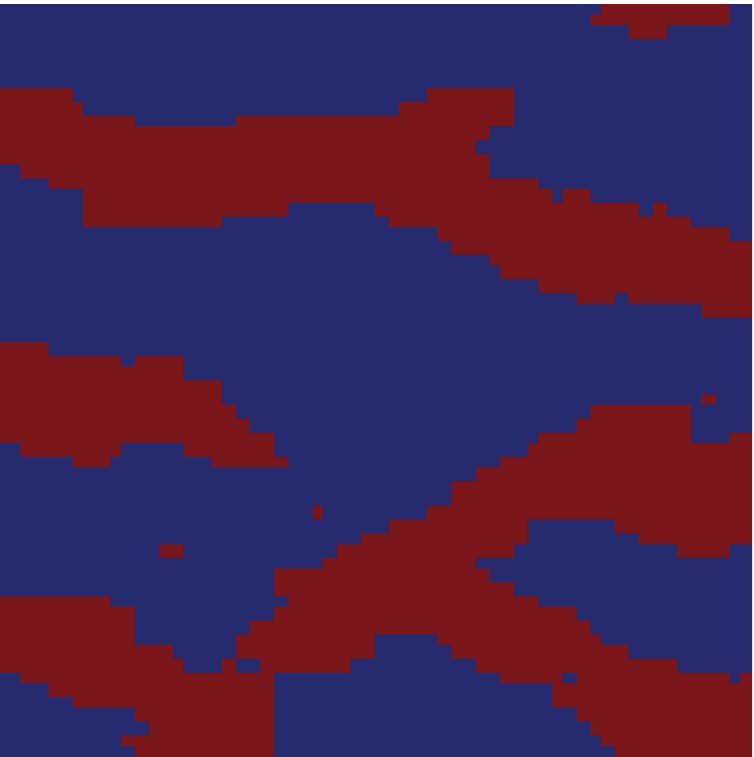}}\\
  \subfloat[VAE-Style]{\includegraphics[width=0.110\linewidth]{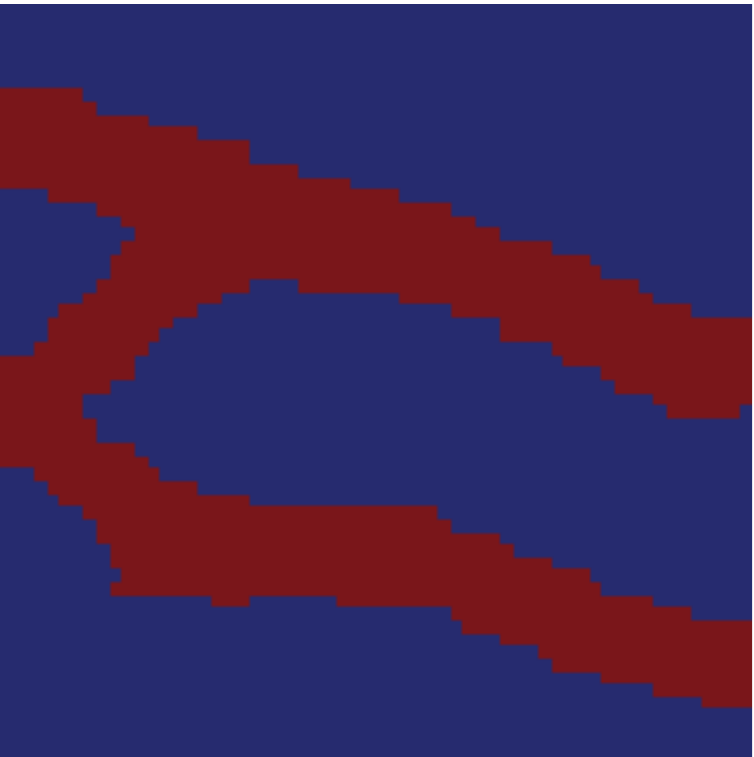}
  \hspace{1mm}\includegraphics[width=0.110\linewidth]{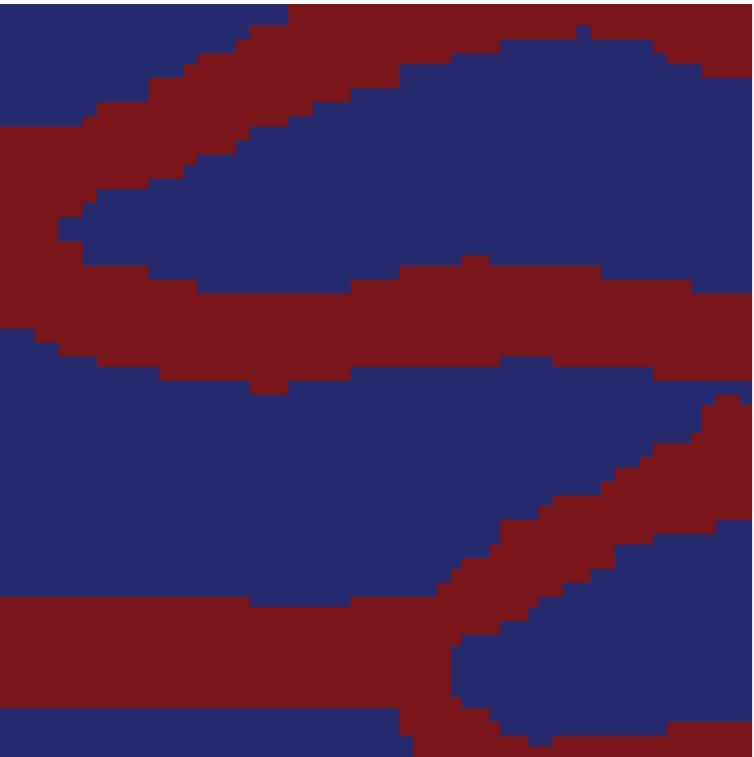}
  \hspace{1mm}\includegraphics[width=0.110\linewidth]{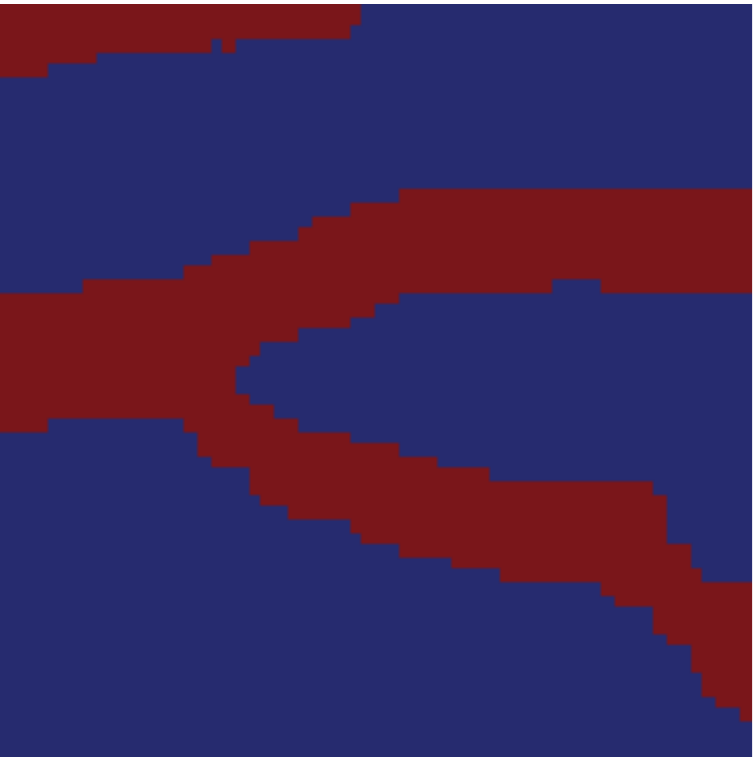}
  \hspace{1mm}\includegraphics[width=0.110\linewidth]{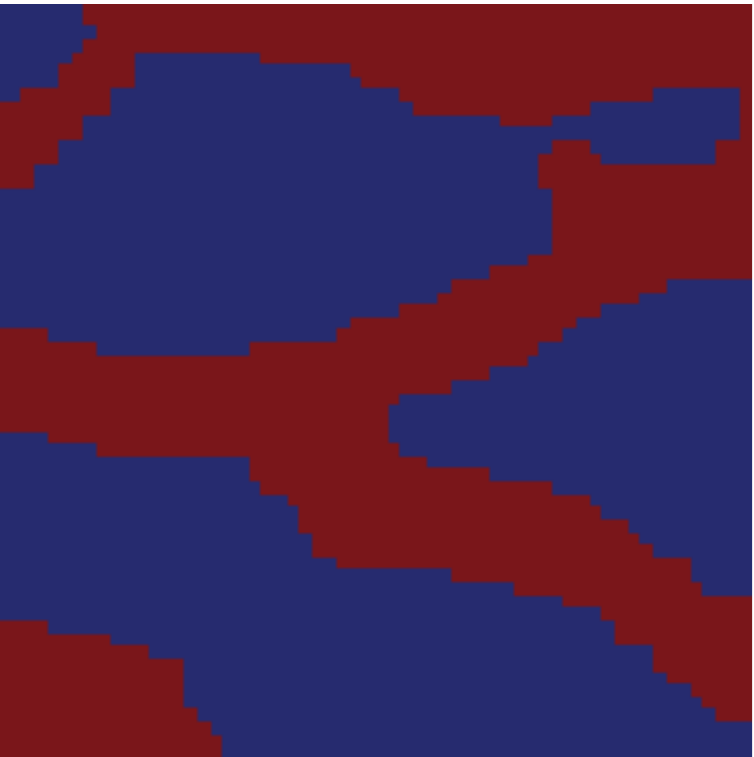}
  \hspace{1mm}\includegraphics[width=0.110\linewidth]{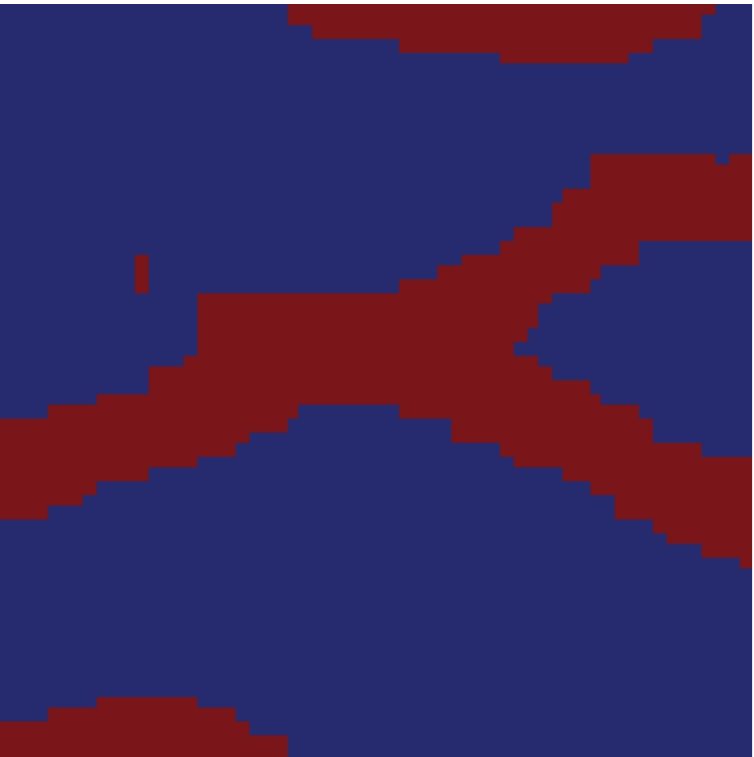}}
\caption{Random realizations of facies from the training set and generated with the different networks. Case 1.}
\label{Fig:Case1-RandomSamples}
\end{figure}

Fig.~\ref{Fig:Case1-RandomSamples} shows five random realizations of facies generated by each trained network. For comparisons, we also show five realizations of the training set. This figure shows that all networks were able to generate realistic facies realizations, with similar channel structures observed in the realizations of the training set.

\begin{figure}
  \centering
  \subfloat[VAE]{\includegraphics[width=1.05\linewidth]{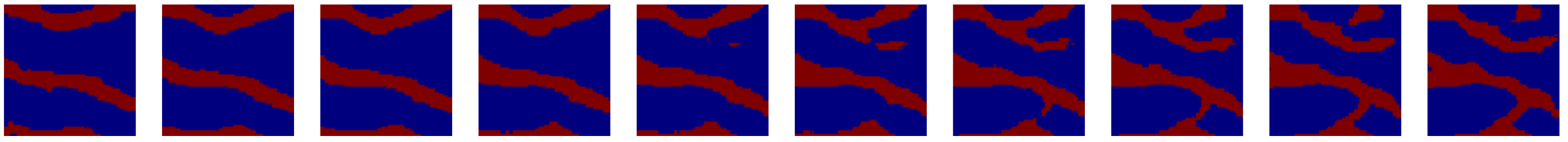}}\\
  \subfloat[GAN]{\includegraphics[width=1.05\linewidth]{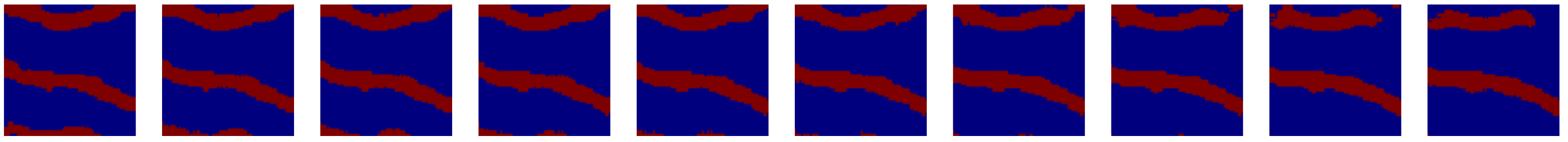}}\\
  \subfloat[WGAN]{\includegraphics[width=1.05\linewidth]{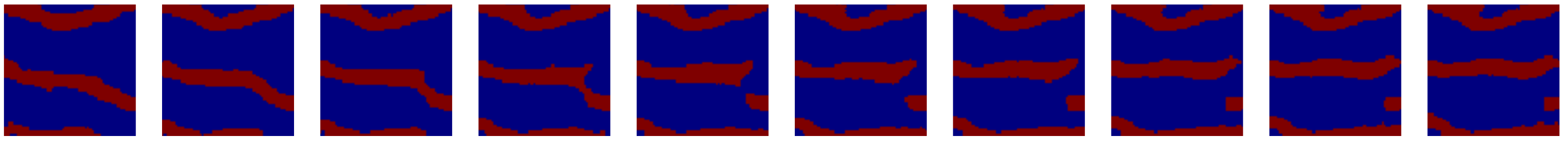}}\\
  \subfloat[$\alpha$-GAN]{\includegraphics[width=1.05\linewidth]{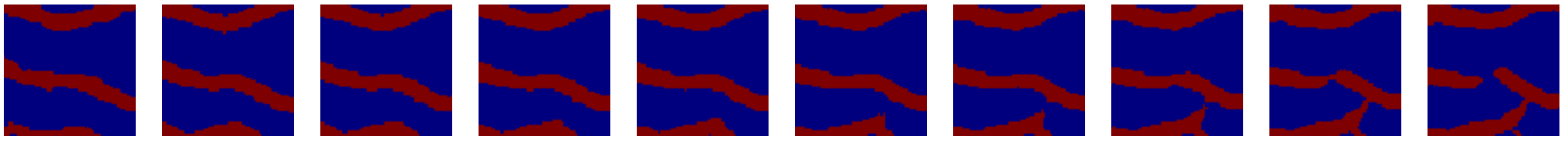}}\\
  \subfloat[PCA-Cycle-GAN]{\includegraphics[width=1.05\linewidth]{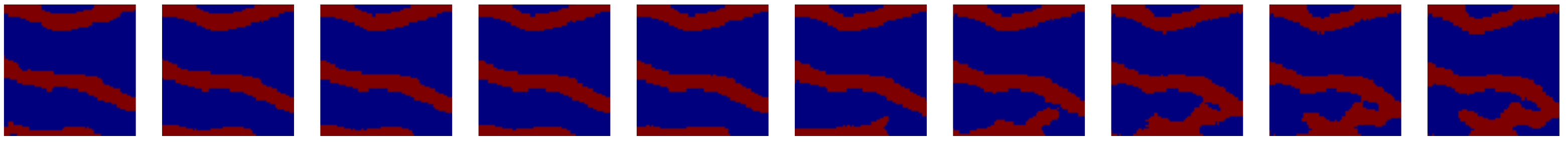}}\\
  \subfloat[PCA-Style]{\includegraphics[width=1.05\linewidth]{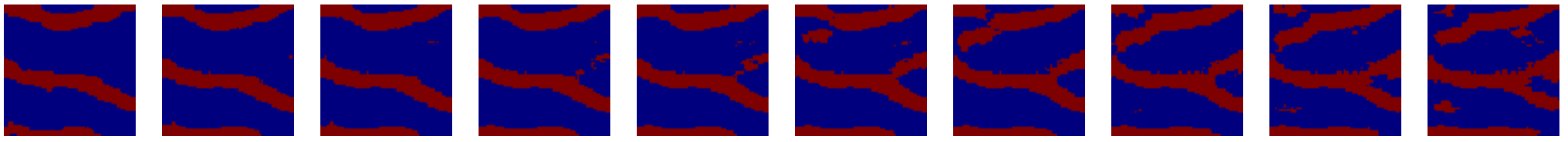}}\\
  \subfloat[VAE-Style]{\includegraphics[width=1.05\linewidth]{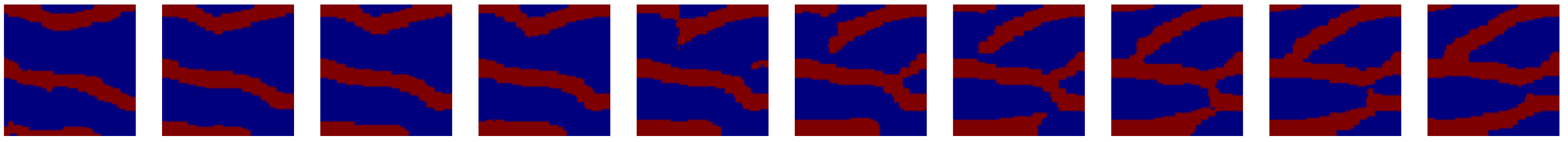}}
\caption{Effect of small perturbation in the latent vector $\z_0$ for each generative network. From left to right it is shown the reconstructed facies using $\gamma = \left[0.0, 0.2, 0.3, 0.4, 0.5, 0.6, 0.7, 0.8, 0.9, 1.0 \right]$ in Eq.~\ref{Eq:Gamma}}
\label{Fig:Case1-gamma0}
\end{figure}

The objective of using the generative models in this work is to construct parameterizations that can be updated using ES-MDA for assimilating observed production data. Mathematically, the generative models learn mappings from latent to facies space. One desirable property is that the learned mappings are relatively continuous in the sense that small changes in the representation in the latent space corresponds to small changes in the resulting facies realization. This phenomena is often referred to manifold learning in the machine learning literature \citep{goodfellow:16bk}. Here, we conducted a small experiment to check the effect of small changes in the latent representation $\z$ in each generative network. We started with 1000 facies realizations which are not part of the training set. For each realization, we found the corresponding latent representation, denoted by $\z_0$. Then, we use PCA to generate a square root of the covariance matrix $\C_\z$. Now, we performed perturbations of increasing size using the following expression

\begin{equation}\label{Eq:Gamma}
  \z_\gamma = \z_0 + \gamma \C_\z\sqr \widehat{\z},
\end{equation}
where $\widehat{\z}$ is a random vector sampled from $\mathcal{N}(\mathbf{0}, \mathbf{I})$. For a fixed $\widehat{\z}$, we increase the size of the perturbation by increasing the value of the coefficient $\gamma$. After the perturbation, the vector $\z_\gamma$ is fed to the generative network to generate the corresponding facies. We repeated this process for each network and the results are summarized in Fig.~\ref{Fig:Case1-gamma0}. This figure shows the same initial facies realization after different perturbation sizes for the different generative networks. For most of the networks, the behaviour is similar. Small values of $\gamma$ results in no noticeable changes in the channels of the initial realizations. For $\gamma > 0.5$, we notice some changes, such as the appearance of new branches of channels, but even for $\gamma = 1$ there are still some resemblance with the initial channel distribution. However, a different behaviour was observed for the GAN network. In this case, the reconstructed facies present very small changes for all values of $\gamma$. The same behaviour was observed for other initial realizations (not shown here). This result may indicate a mode collapse during GAN training, in which case the network is not able to generate completely new facies realizations, instead the model returns only small variations around the closest facies model (closest mode).

\subsection{Assimilation of Facies Data}
\label{Sec:AssimilationFaciesData}

\begin{figure}
  \centering
  \subfloat[Prior]{\includegraphics[width=0.155\linewidth]{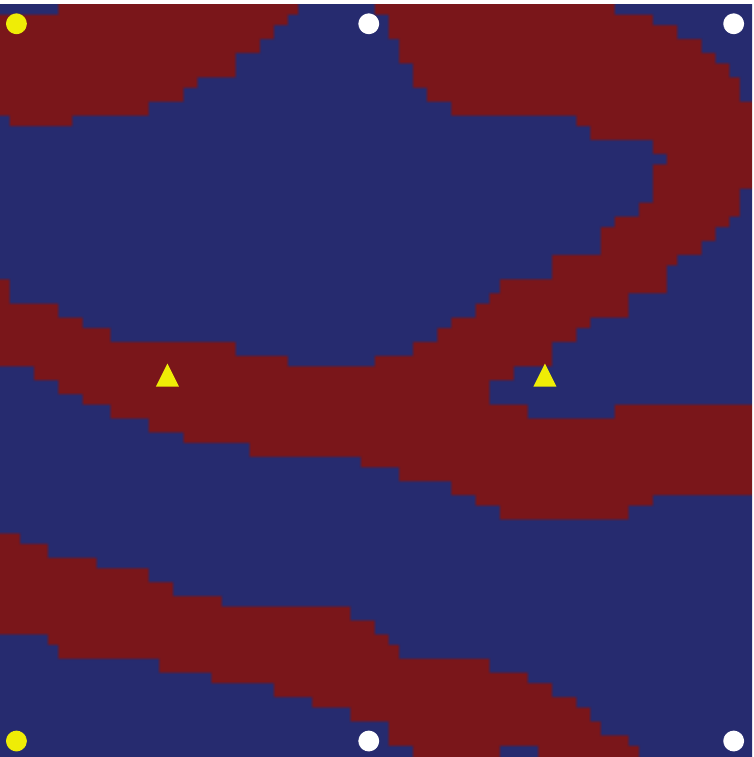}
  \hspace{0.1mm}\includegraphics[width=0.155\linewidth]{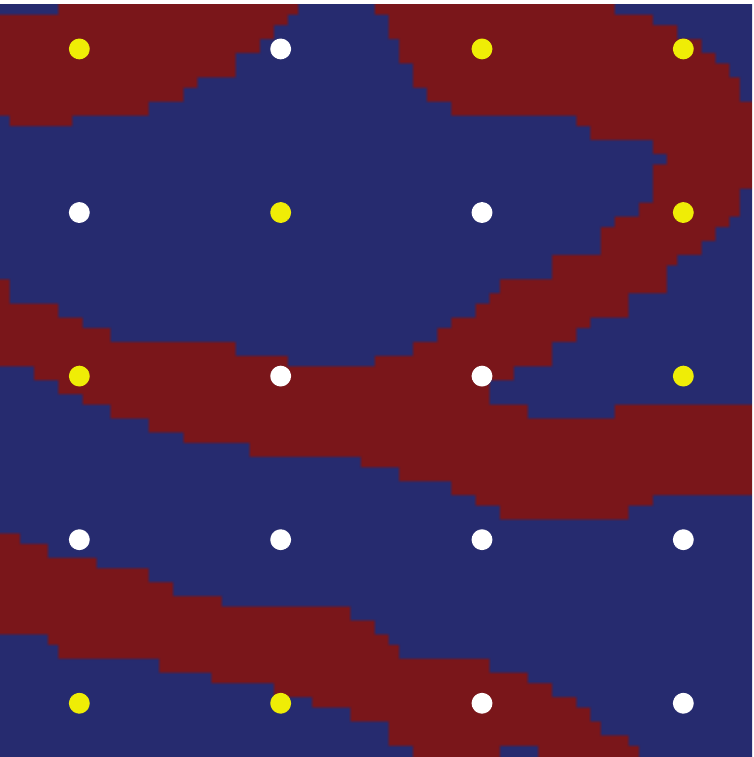}
  \hspace{0.1mm}\includegraphics[width=0.155\linewidth]{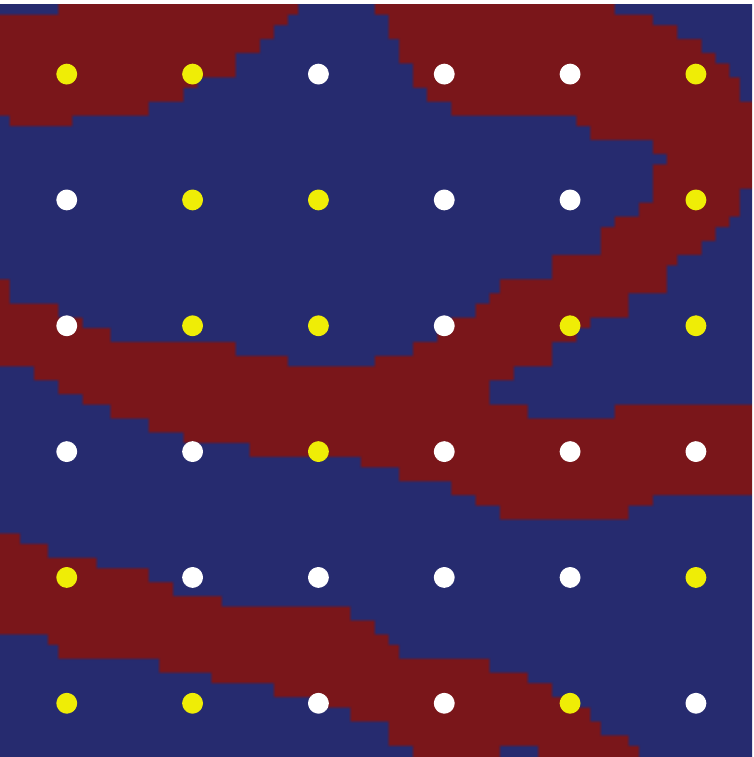}}
  \hspace{3mm}\subfloat[VAE]{\includegraphics[width=0.155\linewidth]{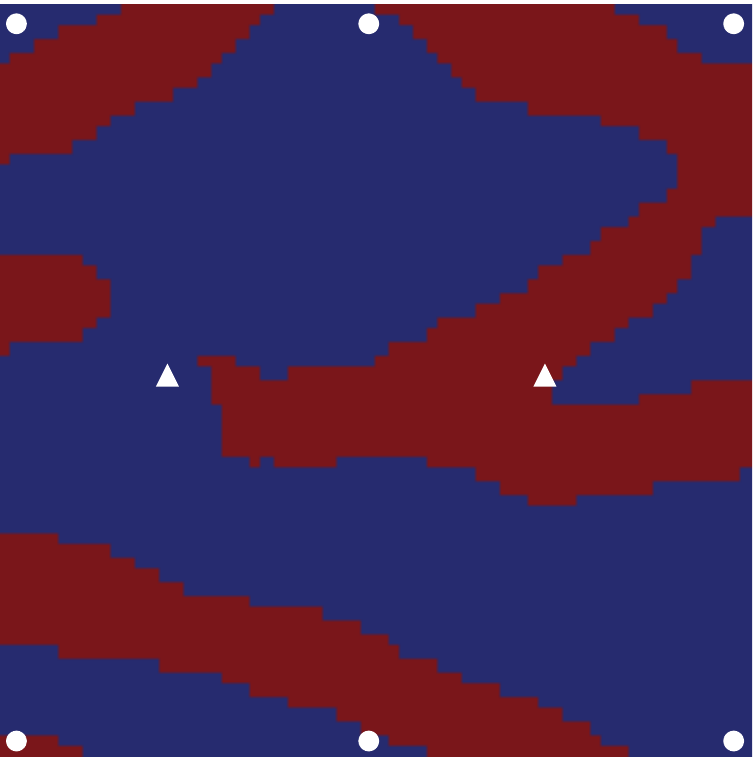}
  \hspace{0.1mm}\includegraphics[width=0.155\linewidth]{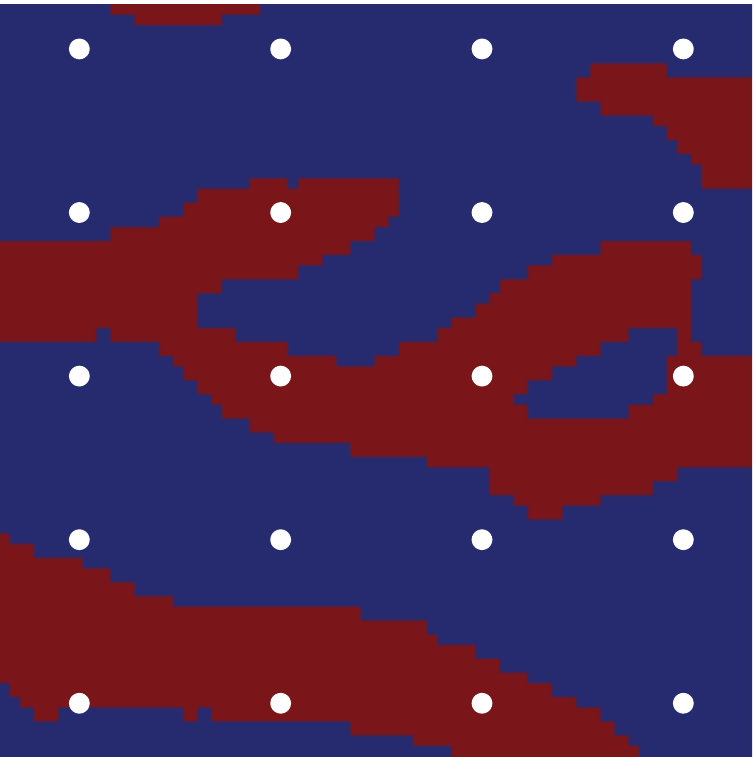}
  \hspace{0.1mm}\includegraphics[width=0.155\linewidth]{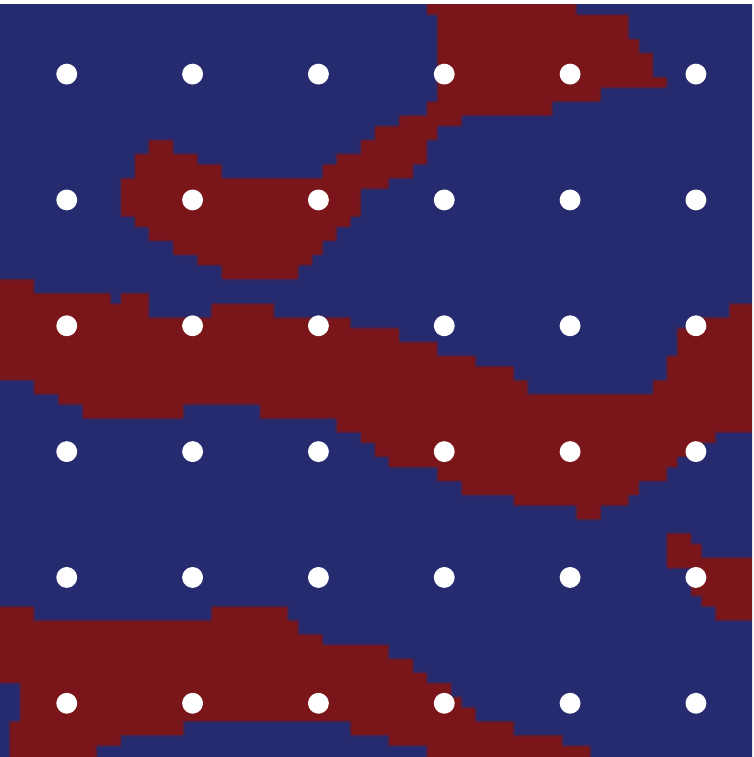}}\\
  \subfloat[GAN]{\includegraphics[width=0.155\linewidth]{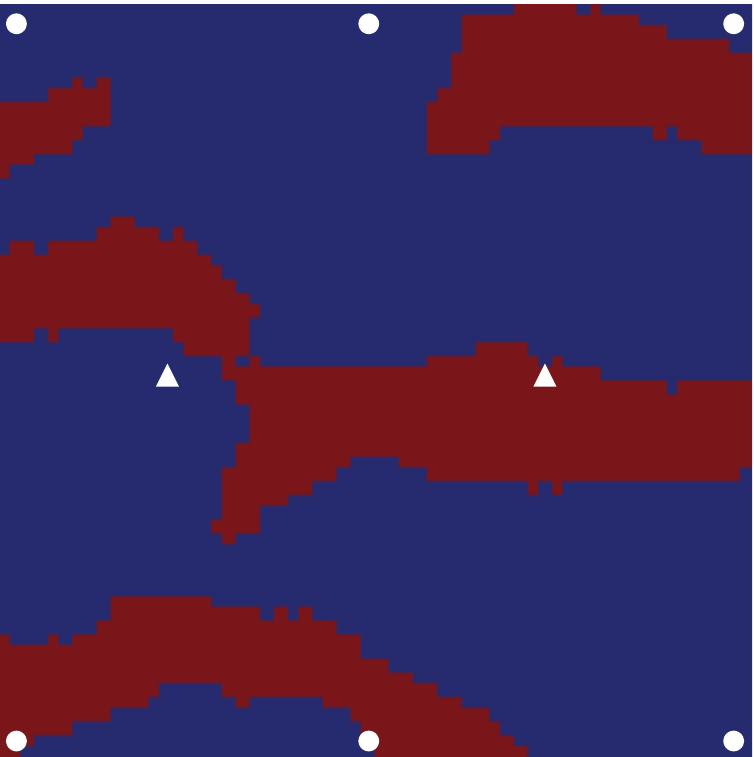}
  \hspace{0.1mm}\includegraphics[width=0.155\linewidth]{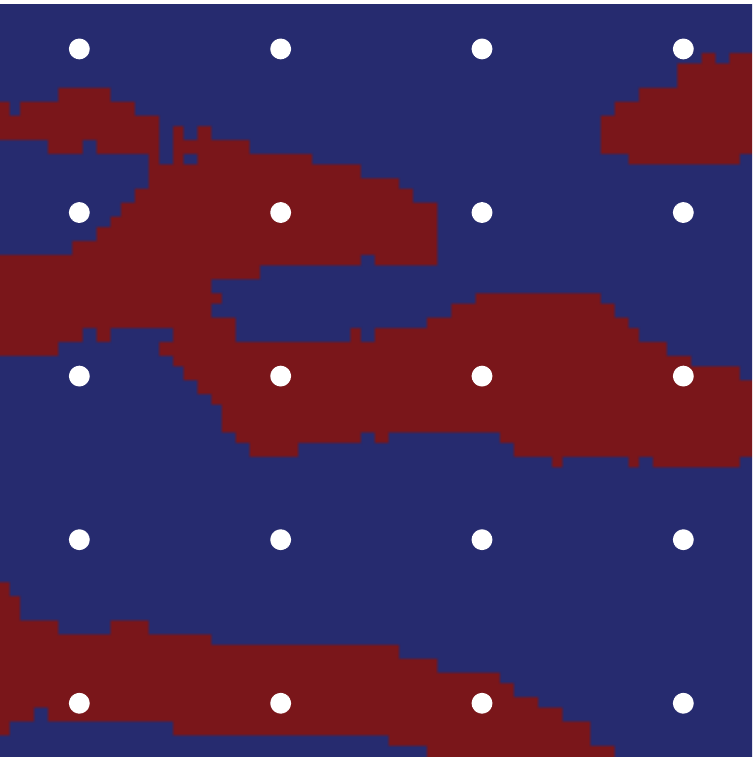}
  \hspace{0.1mm}\includegraphics[width=0.155\linewidth]{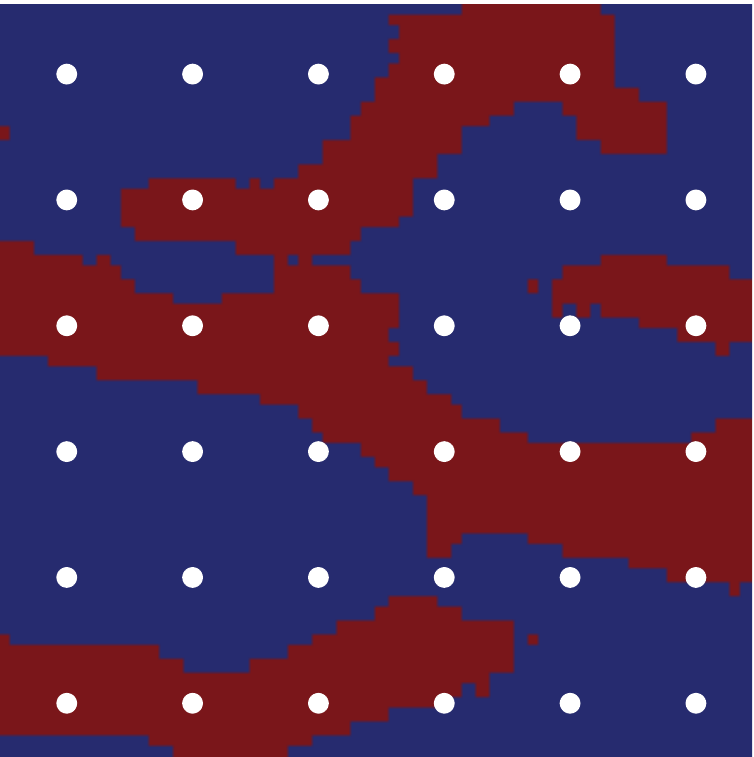}}
  \hspace{3mm}\subfloat[WGAN]{\includegraphics[width=0.155\linewidth]{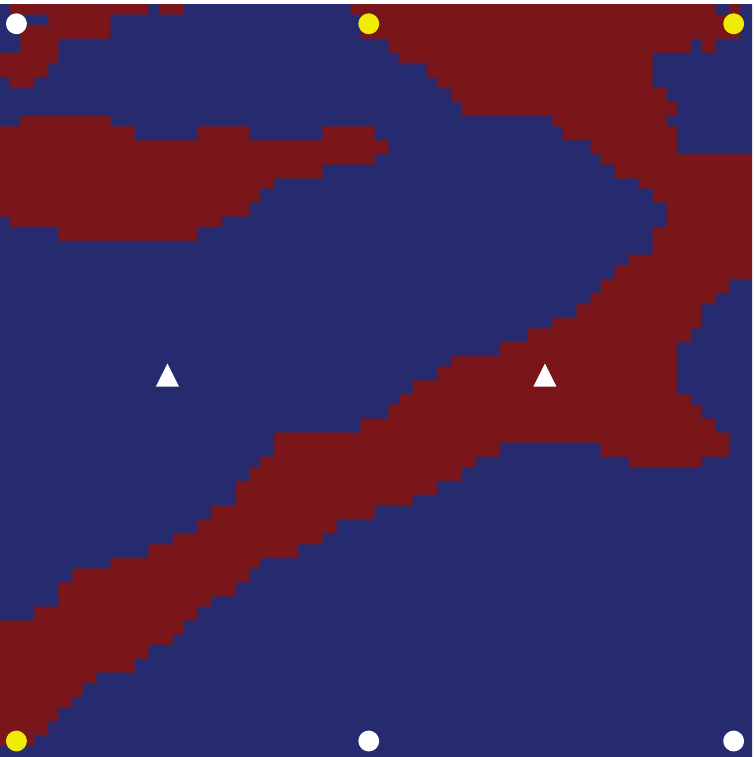}
  \hspace{0.1mm}\includegraphics[width=0.155\linewidth]{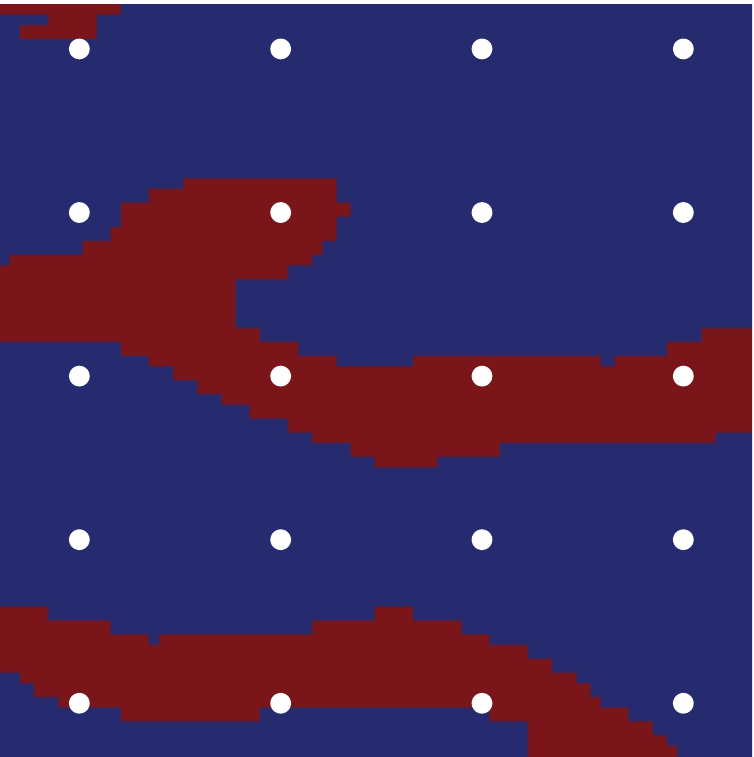}
  \hspace{0.1mm}\includegraphics[width=0.155\linewidth]{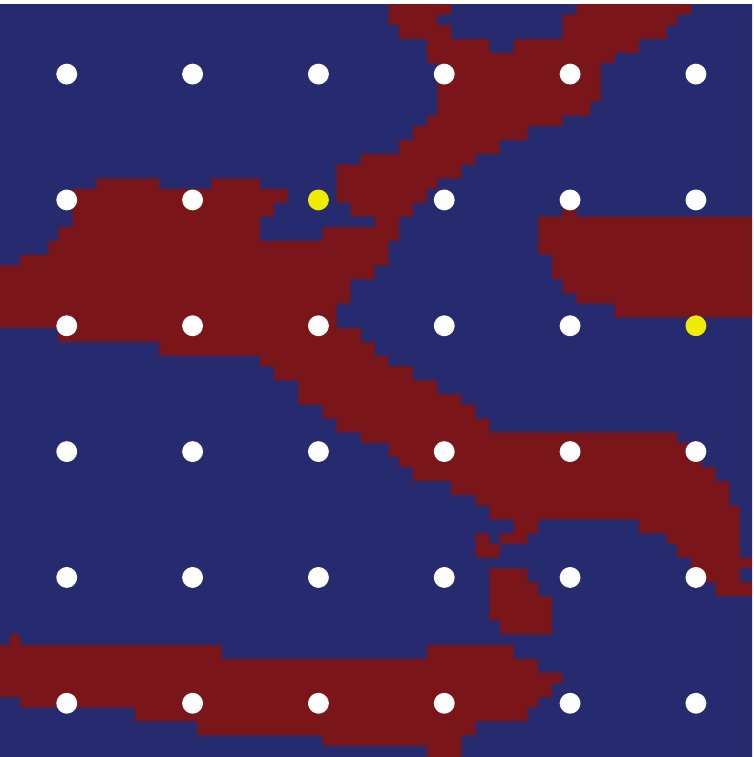}}\\
  \subfloat[$\alpha$-GAN]{\includegraphics[width=0.155\linewidth]{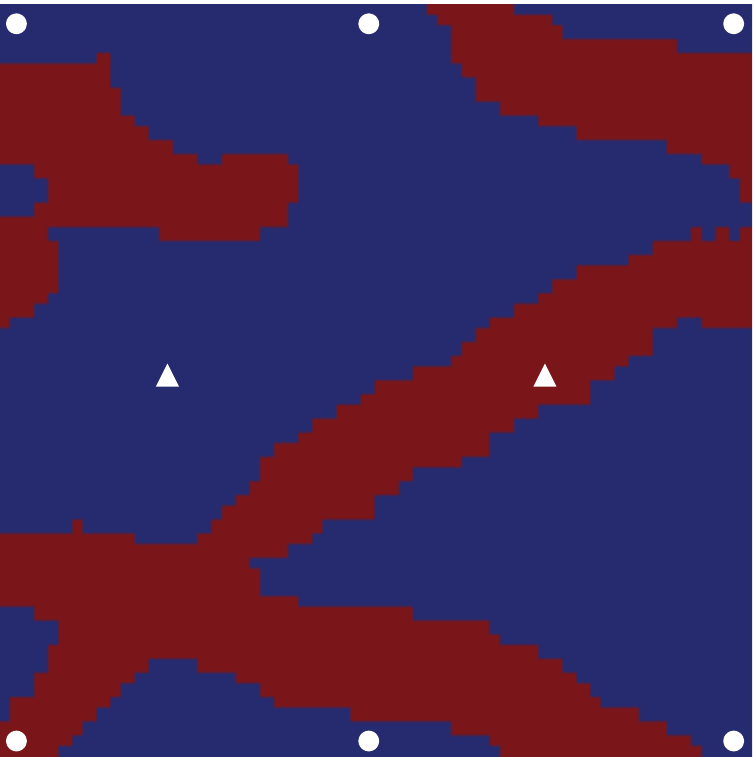}
  \hspace{0.1mm}\includegraphics[width=0.155\linewidth]{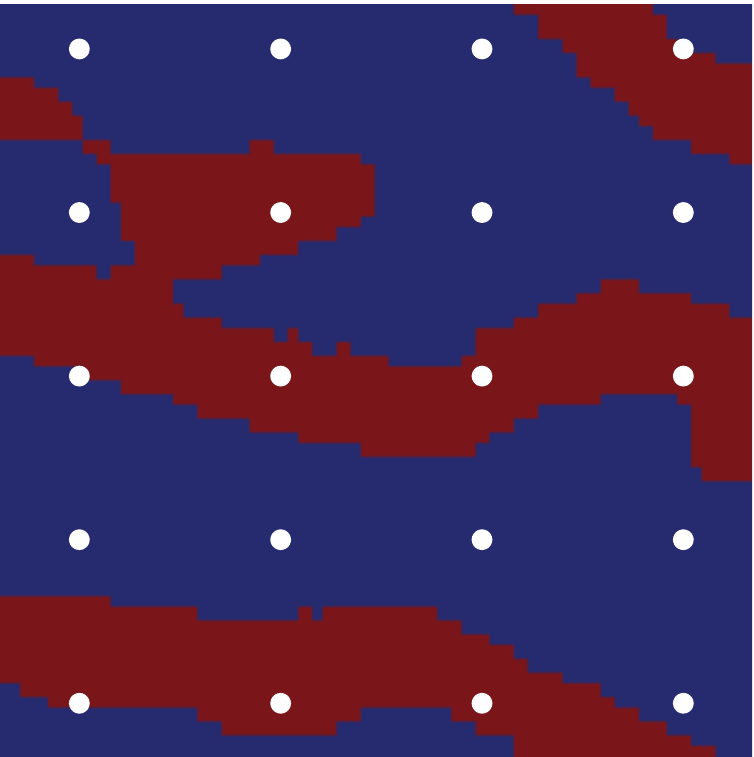}
  \hspace{0.1mm}\includegraphics[width=0.155\linewidth]{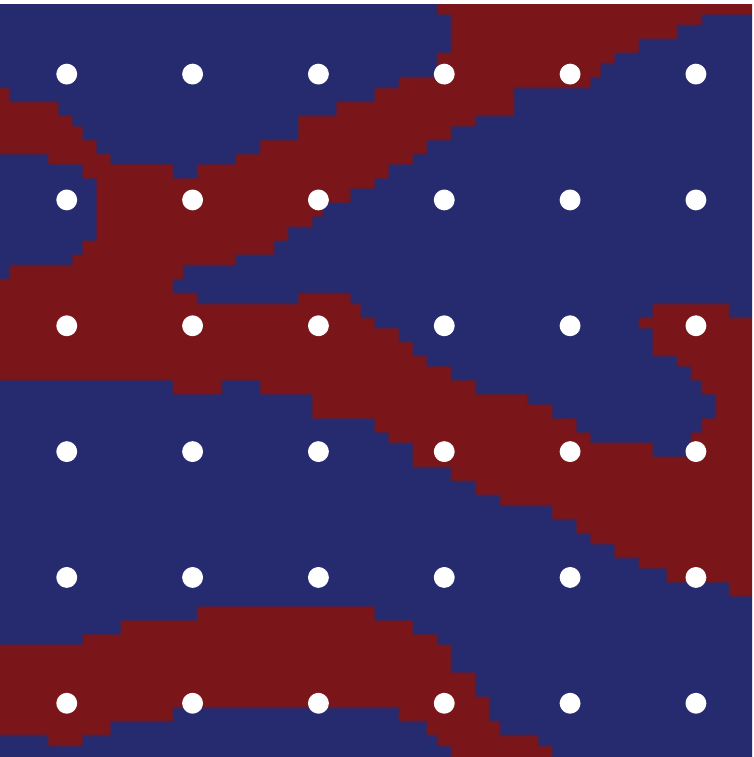}}
  \hspace{3mm}\subfloat[PCA-Cycle-GAN]{\includegraphics[width=0.155\linewidth]{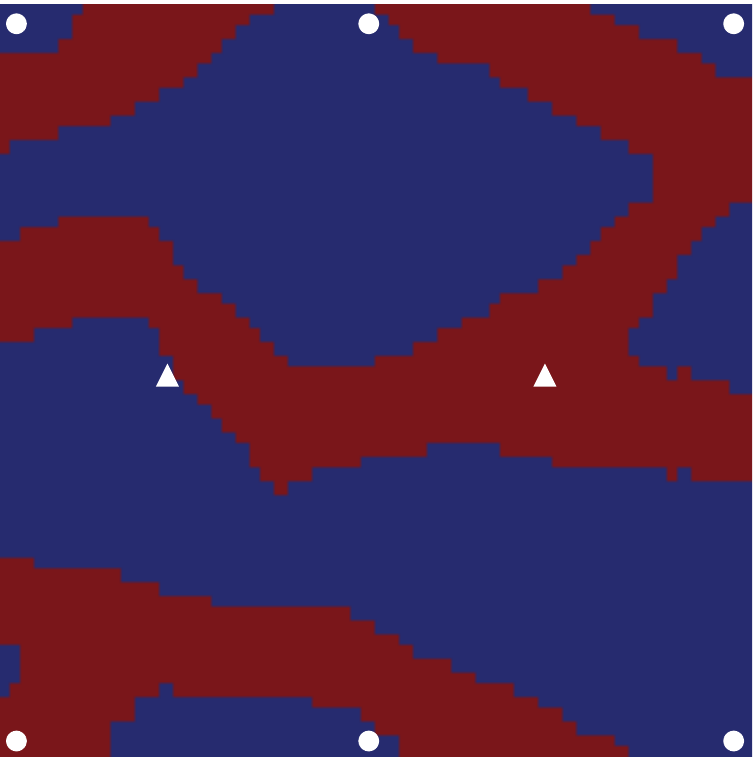}
  \hspace{0.1mm}\includegraphics[width=0.155\linewidth]{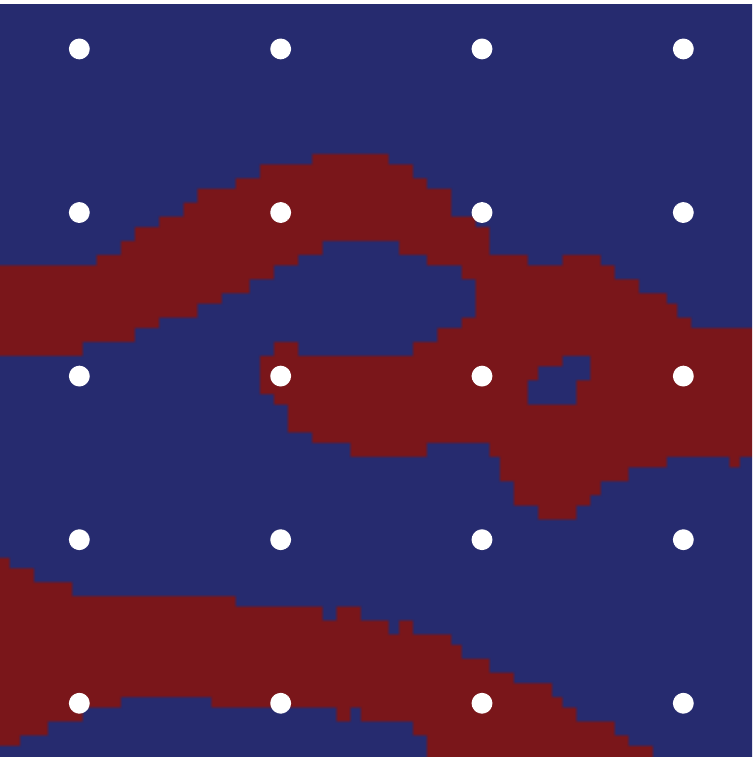}
  \hspace{0.1mm}\includegraphics[width=0.155\linewidth]{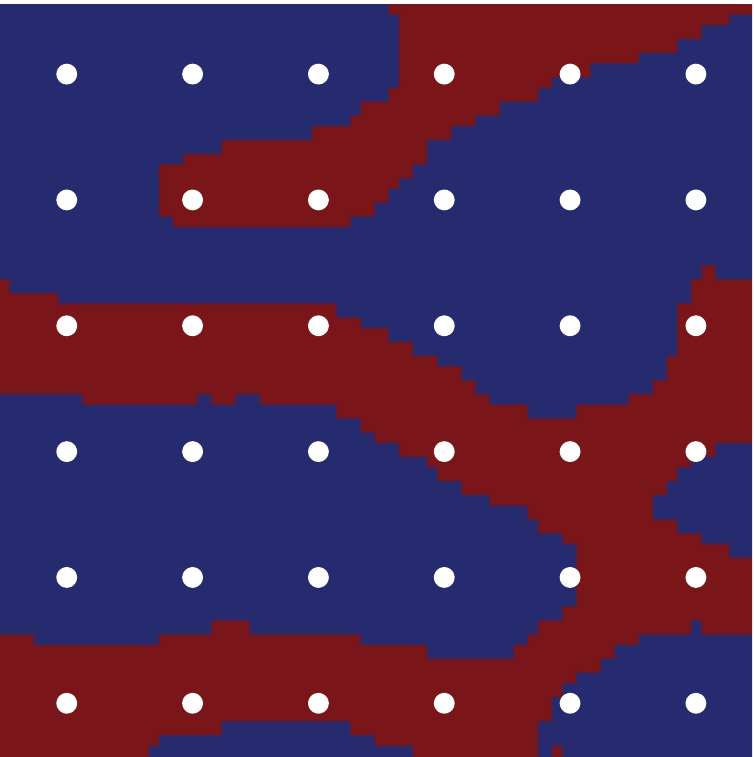}}\\
  \subfloat[PCA-Style]{\includegraphics[width=0.155\linewidth]{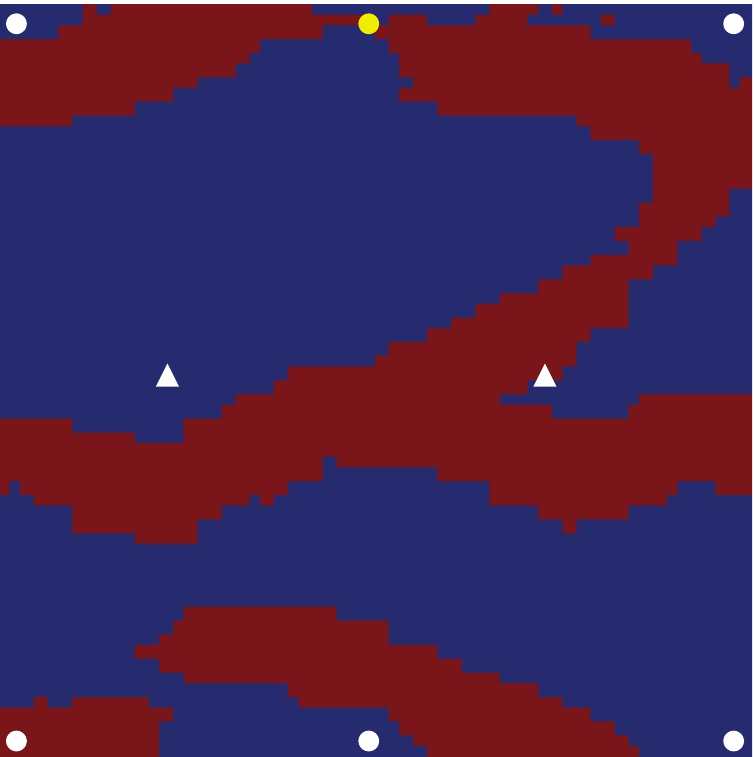}
  \hspace{0.1mm}\includegraphics[width=0.155\linewidth]{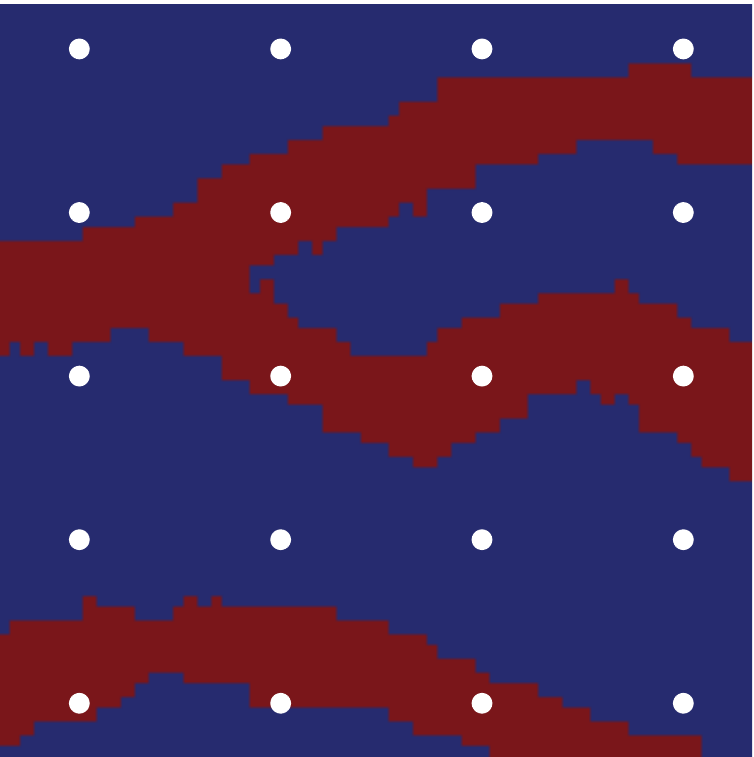}
  \hspace{0.1mm}\includegraphics[width=0.155\linewidth]{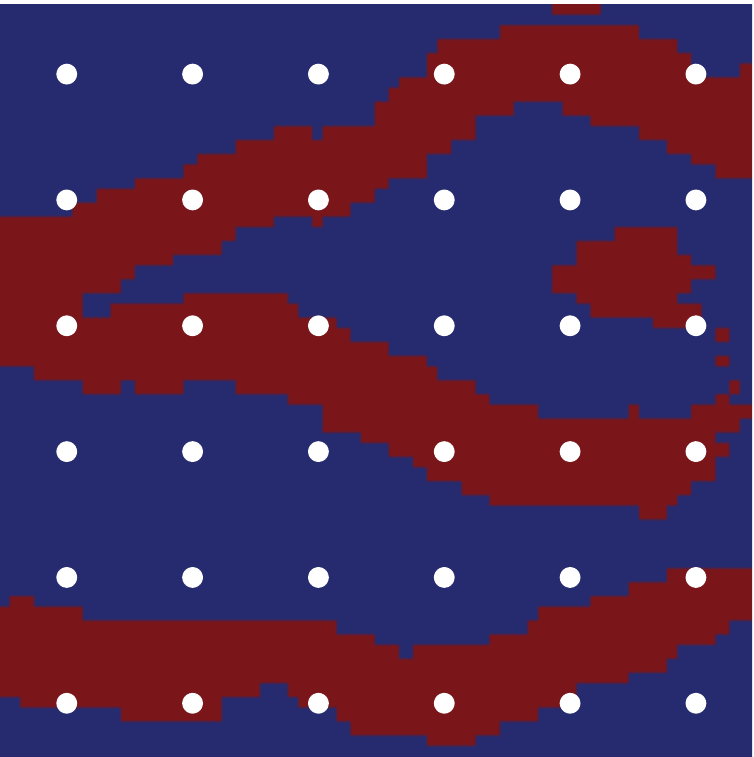}}
  \hspace{3mm}\subfloat[VAE-Style]{\includegraphics[width=0.155\linewidth]{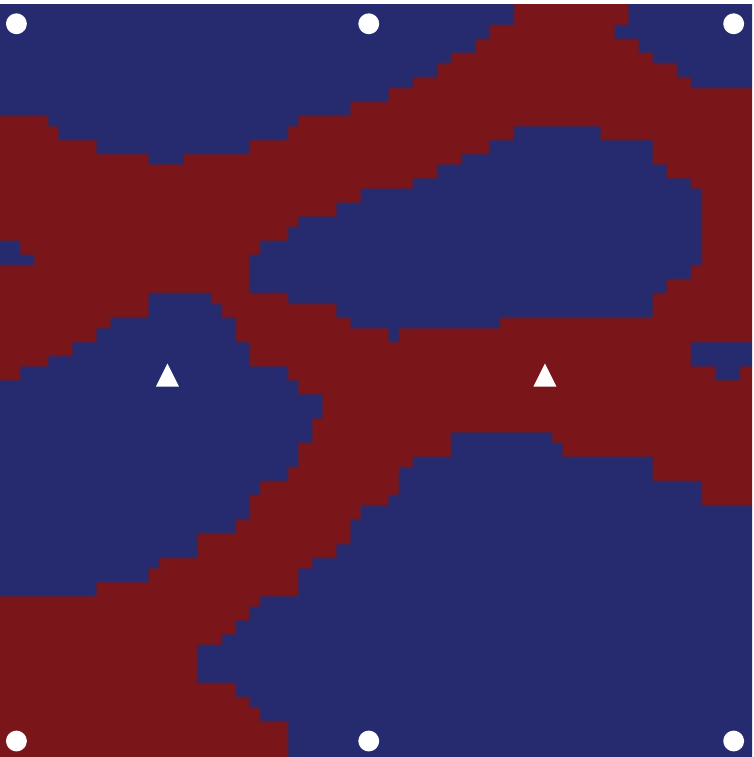}
  \hspace{0.1mm}\includegraphics[width=0.155\linewidth]{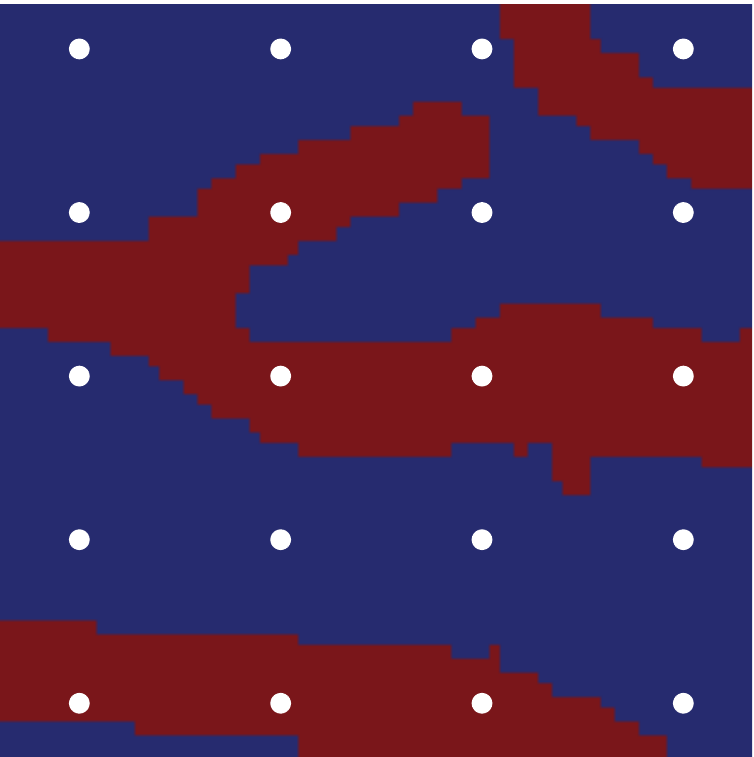}
  \hspace{0.1mm}\includegraphics[width=0.155\linewidth]{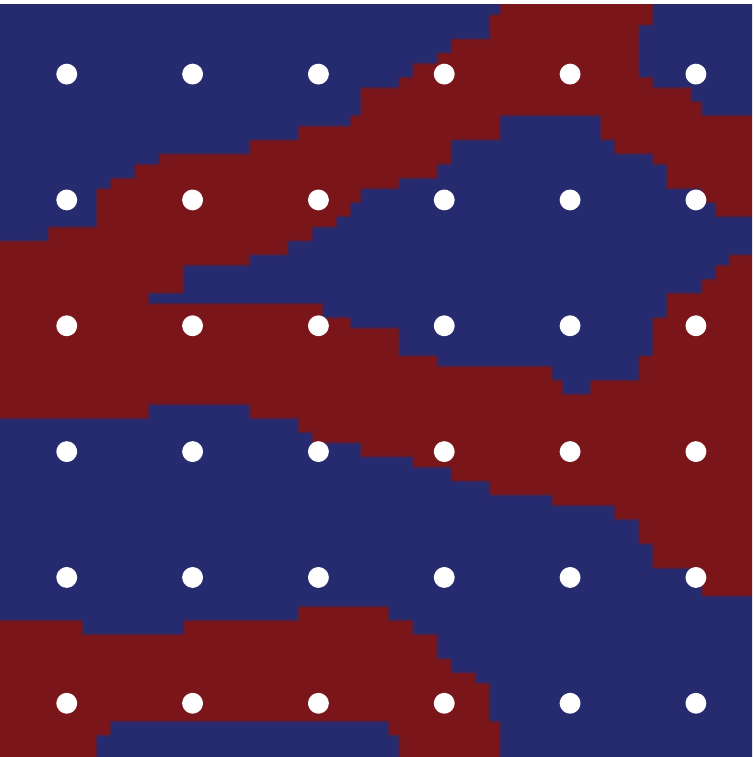}}
\caption{Random realization of facies obtained with the different networks after assimilation of facies data. Case 1. The columns correspond to 8, 20 and 36 data points. The yellow circles indicate position of wells with failure to obtain the correct facies type.}
\label{Fig:Case1-FaciesData}
\end{figure}

The training set was built without conditioning to facies type data at well locations. In this section, we test the ability of the method to assimilate this type of information. We applied ES-MDA with $N_a = 10$ data assimilations and $N_e = 200$ considering cases with eight, 20 and 36 data points. In all cases, the facies data were assumed equal to one for channel and zero for background. The standard deviation of the data-error was assumed equal to 0.05. Table~\ref{Tab:Case1-HardData} summarizes the results by showing the average number of failures to condition to the facies type in percentage. The results in this table show that WGAN an PCA-Style resulted in slightly larger numbers of failures, but overall all cases were able to successfully honor the hard data. We tried to increase the number of ES-MDA iterations, to check whether we could obtain a case with all data matched for all realizations. Only the VAE network was able to get zero error in the facies type. Fig.~\ref{Fig:Case1-FaciesData} shows one realization of the ensemble after assimilation of facies data for each case. We highlighted in yellow the well positions were the data assimilation failed to get the correct facies type. For this particular realization, only the prior and the WGAN missed a few data points.

\begin{table}[ht!]
\caption{Average percentage of failures conditioning to facies data. Case 1}
\label{Tab:Case1-HardData}
\begin{small}
\begin{center}
\begin{tabular}{lccc}
\toprule
& \multicolumn{3}{c}{Number of data points}\\
\cline{2-4} \\
Network & 8  & 20  & 36 \\
\midrule
Prior & 39.88\% & 46.48\% & 48.11\%\\
VAE & 0.00\% & 0.00\% & 0.03\% \\
GAN & 0.25\% & 0.20\% & 0.28\% \\
WGAN & 2.06\% & 1.30\% & 0.81\% \\
$\alpha$-GAN & 0.50\% & 0.33\% & 0.21\% \\
PCA-Cycle-GAN & 0.06\% & 0.18\% & 0.26\% \\
PCA-Style & 1.50\% & 0.18\% & 0.76\% \\
VAE-Style & 0.19\% & 0.08\% & 0.10\% \\
\bottomrule
\end{tabular}
\end{center}
\end{small}
\end{table}

\subsection{Assimilation of Production Data}
\label{Sec:AssimilationProdDataCase1}

In this section, we compare the performance of the networks combined to ES-MDA to history match production data from the wells. In the reservoir all oil producing wells are controlled by a constant bottom-hole pressure (BHP) of 150~bars, while the injectors are controlled by a constant BHP of 350~bars. The observed data used in the history matching correspond the monthly measurements of water cut and water injection rate for a period of 10 years. The synthetic measurements were generated adding random noise to the data predicted by the reference model with standard deviation corresponding to 5\% of the data values. The history matching started with $N_e = 200$ prior realizations of facies. The same realizations were used in all cases and they are not part of the training set. All results were obtained with $N_a = 8$ ES-MDA iterations with constant inflation.

\begin{figure}
  \centering
  \subfloat[Prior]{\includegraphics[width=0.20\linewidth]{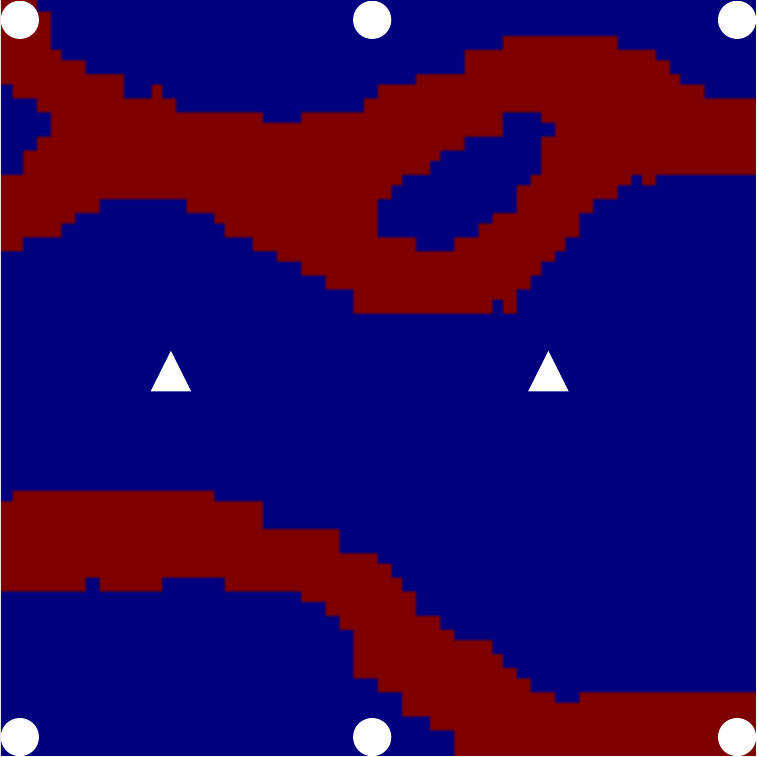}}
  \hspace{3mm}\subfloat[VAE]{\includegraphics[width=0.20\linewidth]{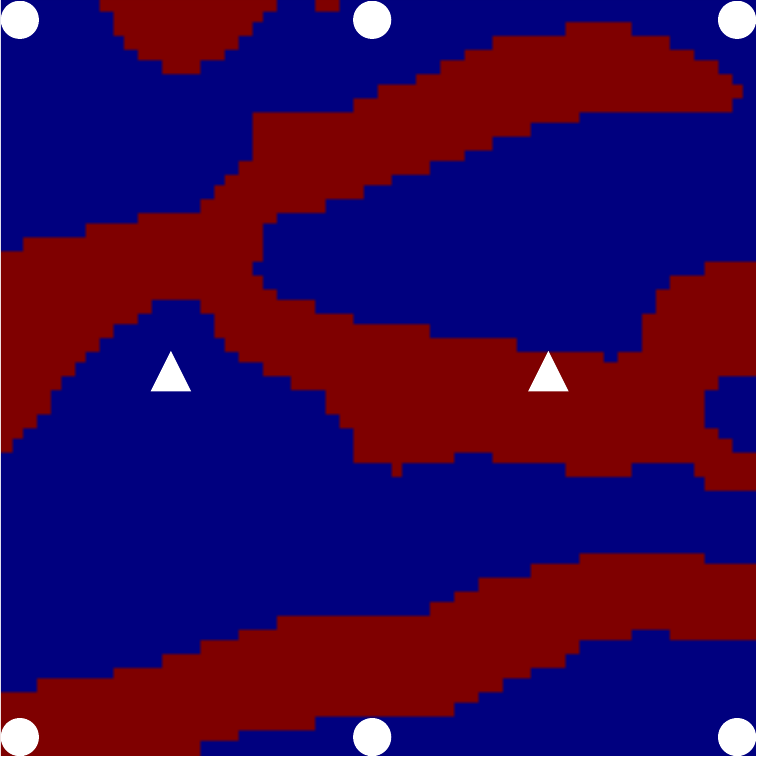}}
  \hspace{3mm}\subfloat[GAN]{\includegraphics[width=0.20\linewidth]{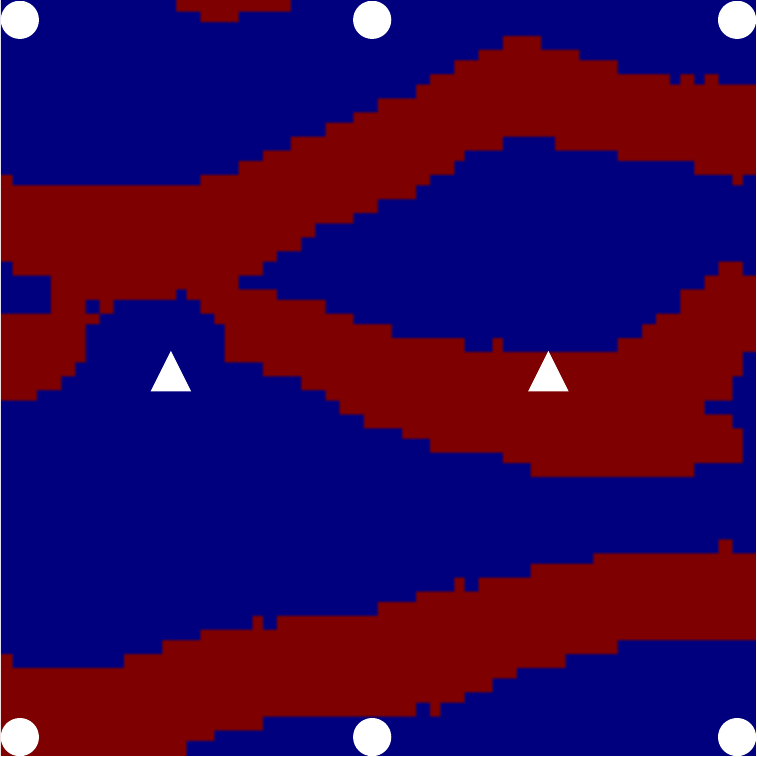}}
  \hspace{3mm}\subfloat[WGAN]{\includegraphics[width=0.20\linewidth]{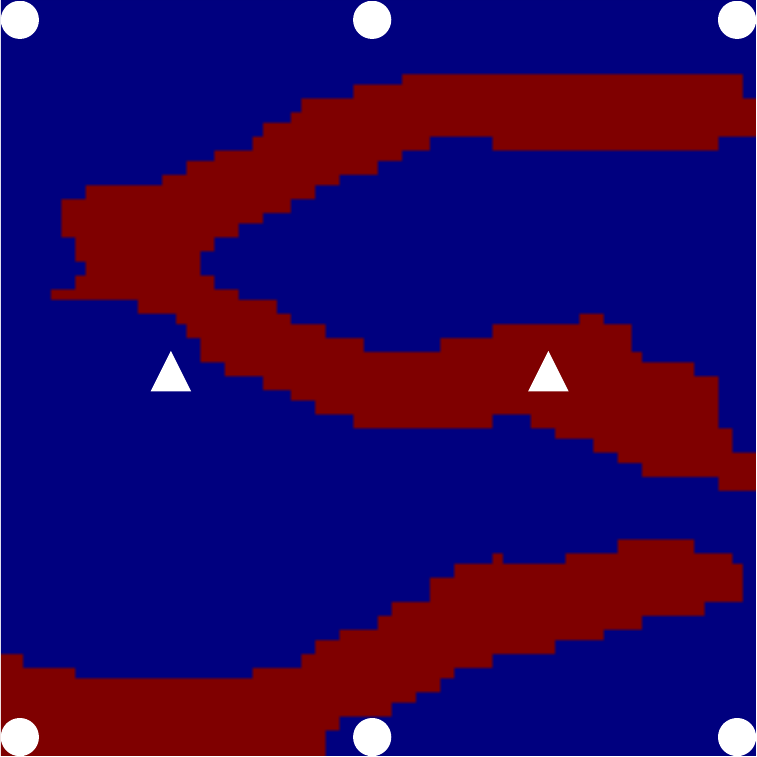}}\\
  \subfloat[$\alpha$-GAN]{\includegraphics[width=0.20\linewidth]{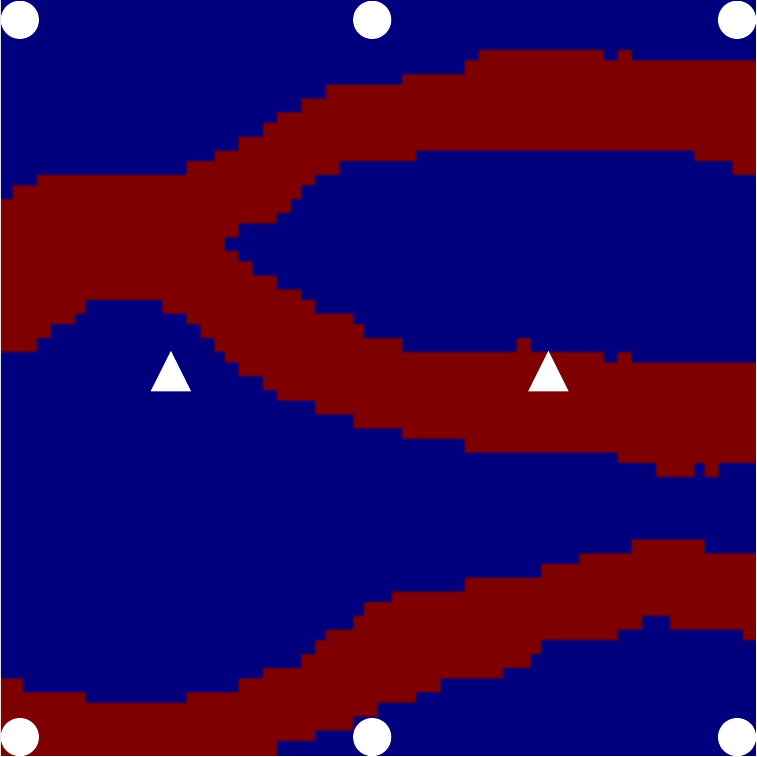}}
  \hspace{3mm}\subfloat[PCA-Cycle-GAN]{\includegraphics[width=0.20\linewidth]{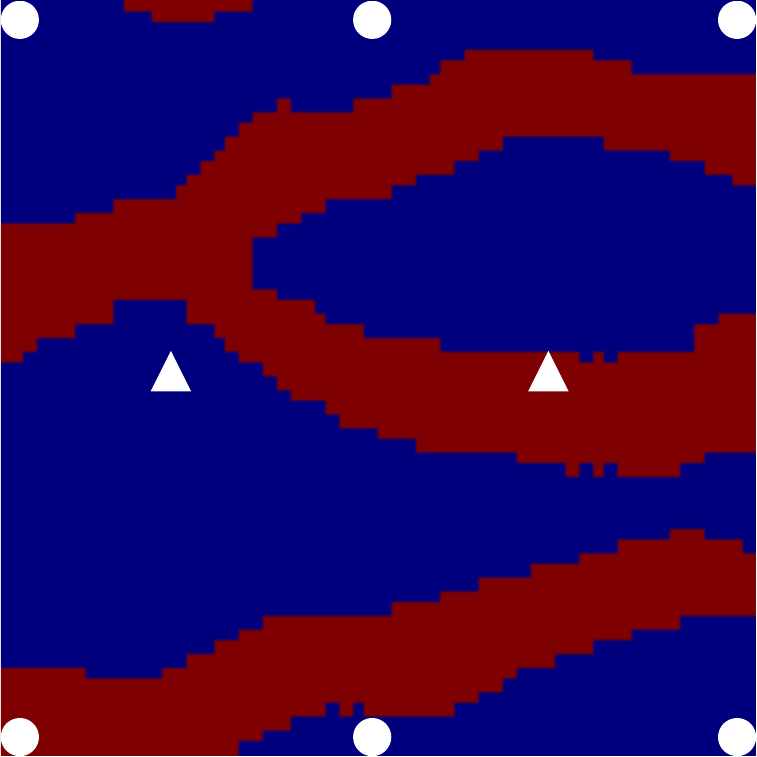}}
  \hspace{3mm}\subfloat[PCA-Style]{\includegraphics[width=0.20\linewidth]{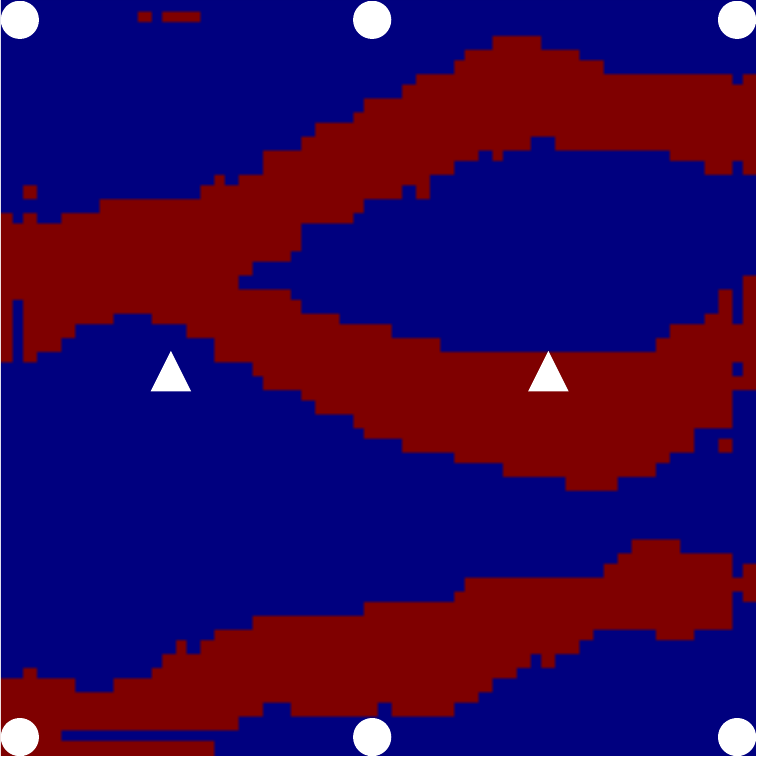}}
  \hspace{3mm}\subfloat[VAE-Style]{\includegraphics[width=0.20\linewidth]{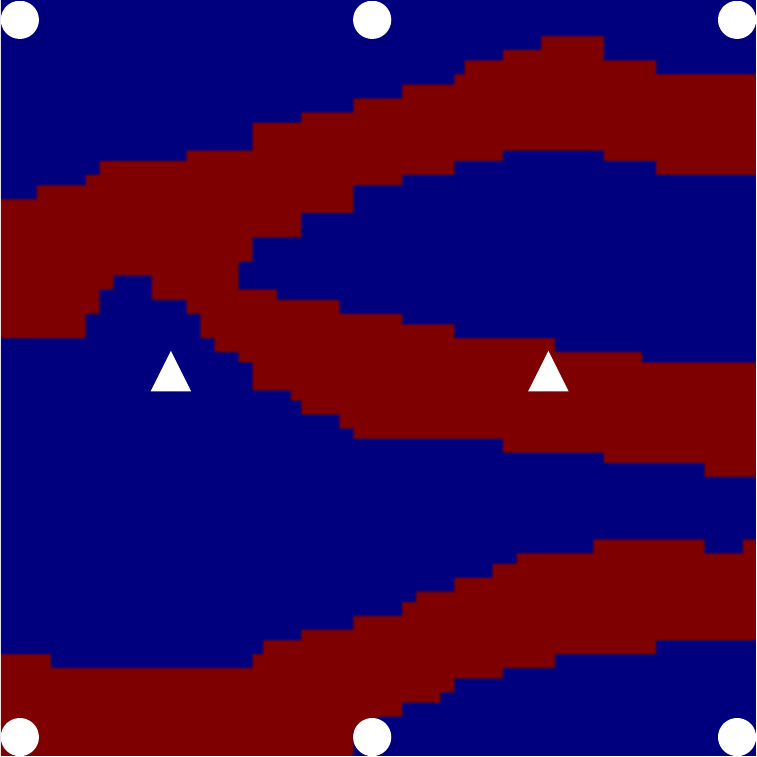}}
\caption{First realization of facies obtained with the different networks after assimilation of production data. Case 1.}
\label{Fig:Case1-001}
\end{figure}

\begin{figure}
  \centering
  \subfloat[Prior]{\includegraphics[width=0.20\linewidth]{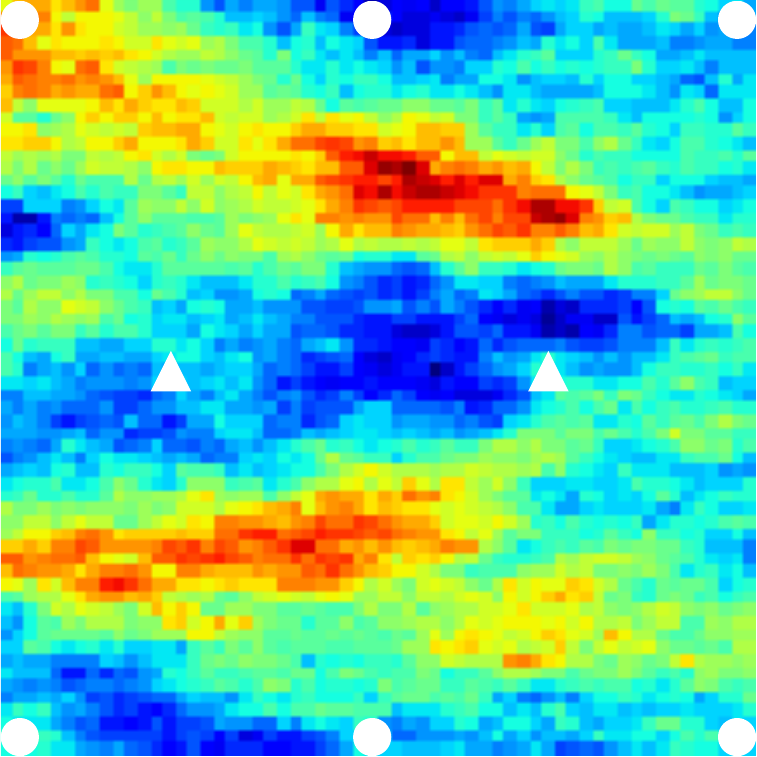}}
  \hspace{3mm}\subfloat[VAE]{\includegraphics[width=0.20\linewidth]{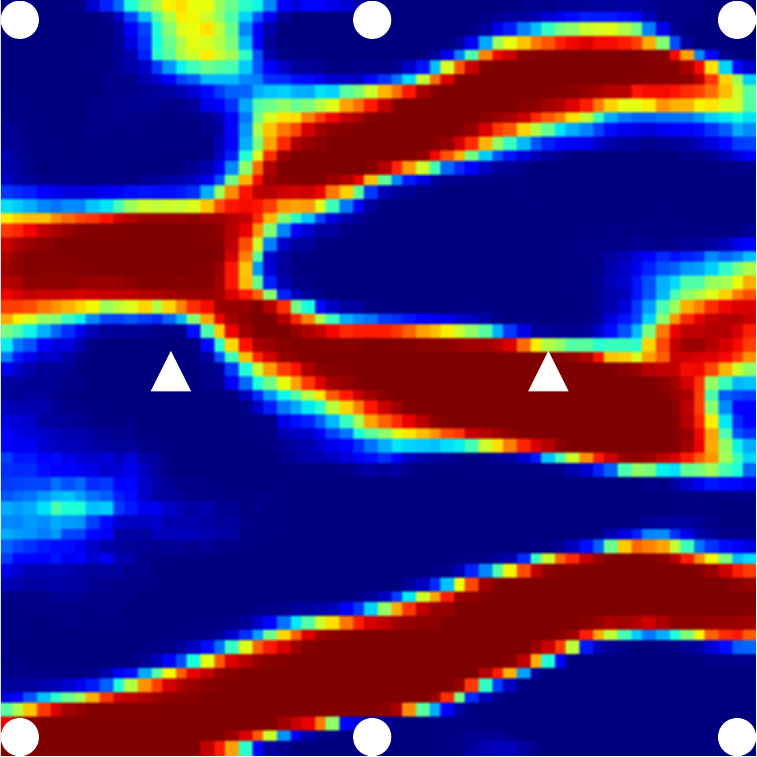}}
  \hspace{3mm}\subfloat[GAN]{\includegraphics[width=0.20\linewidth]{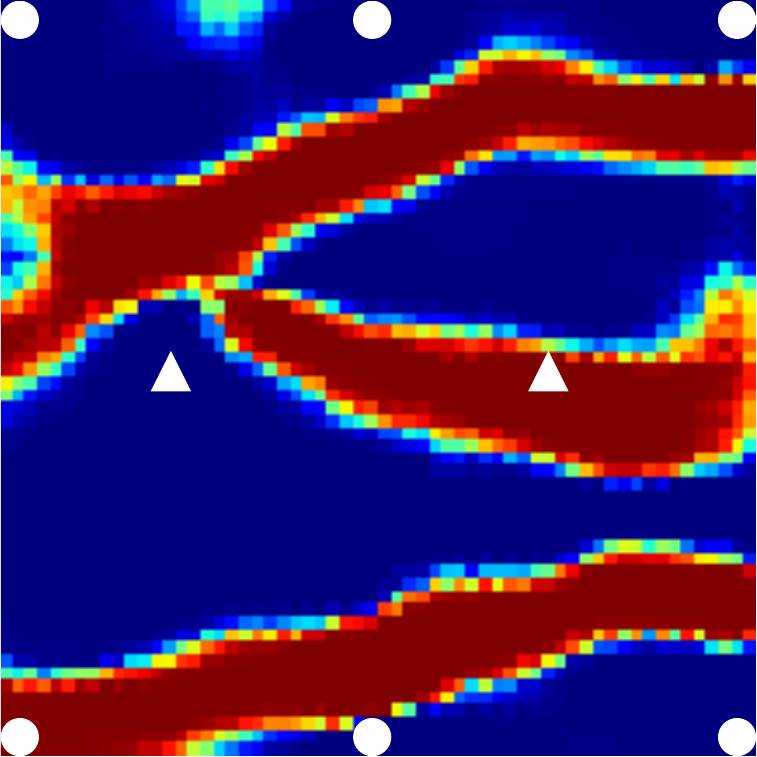}}
  \hspace{3mm}\subfloat[WGAN]{\includegraphics[width=0.20\linewidth]{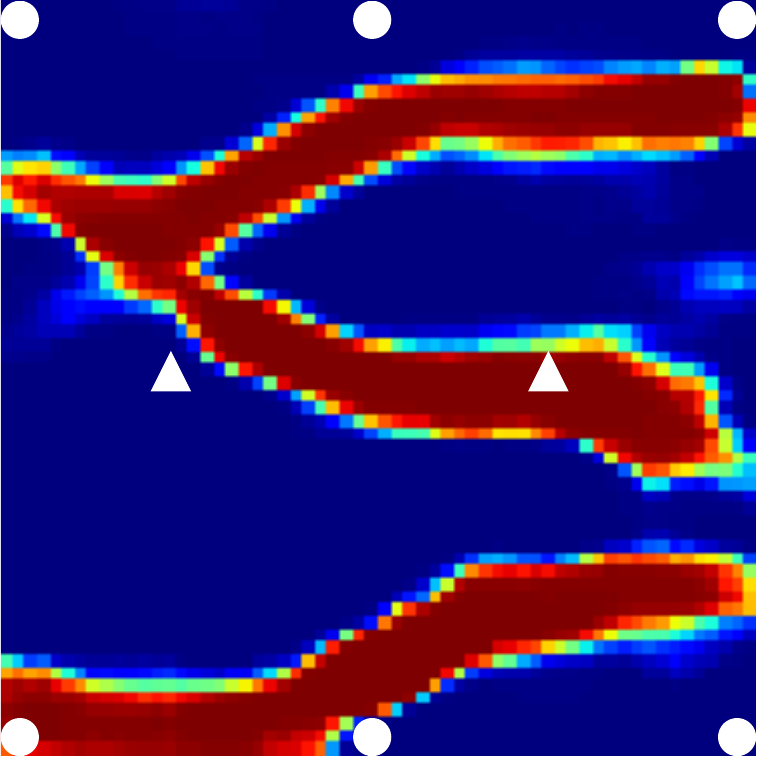}}\\
  \subfloat[$\alpha$-GAN]{\includegraphics[width=0.20\linewidth]{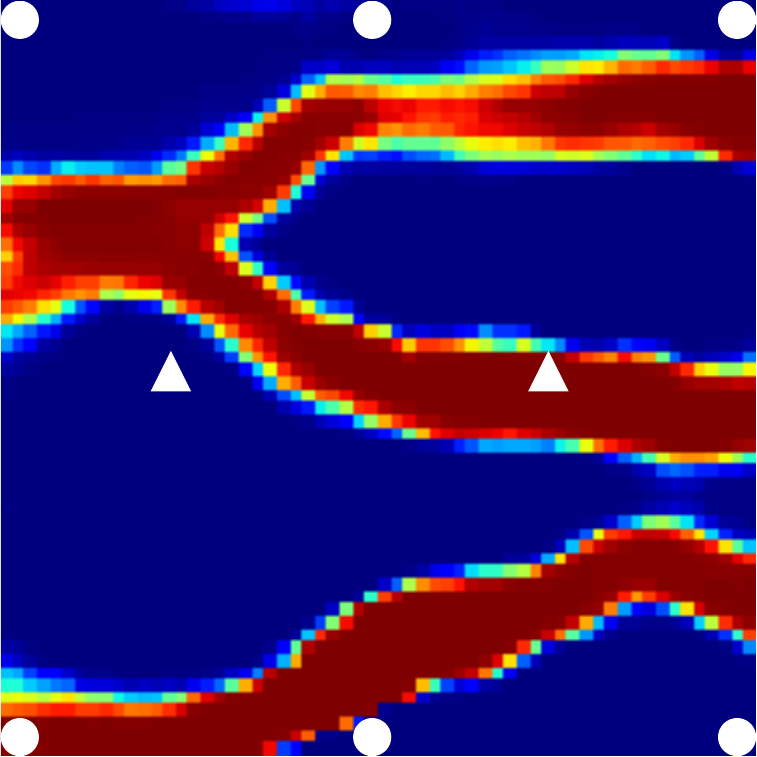}}
  \hspace{3mm}\subfloat[PCA-Cycle-GAN]{\includegraphics[width=0.20\linewidth]{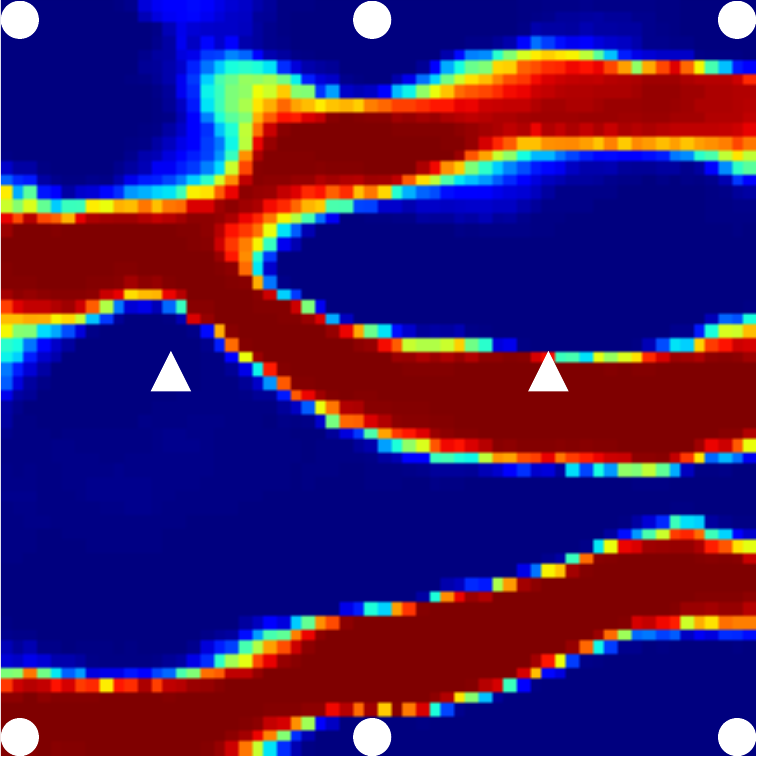}}
  \hspace{3mm}\subfloat[PCA-Style]{\includegraphics[width=0.20\linewidth]{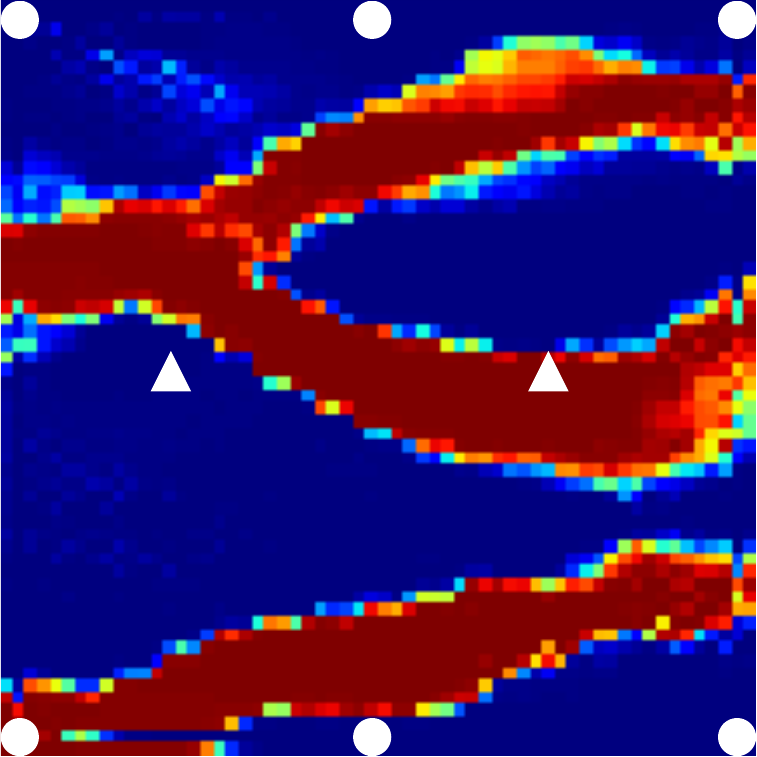}}
  \hspace{3mm}\subfloat[VAE-Style]{\includegraphics[width=0.20\linewidth]{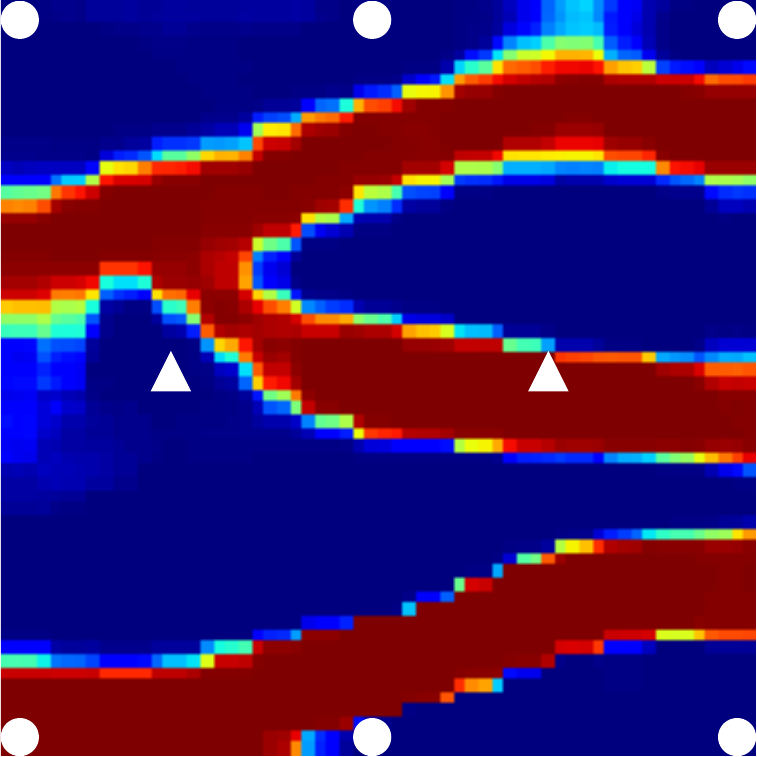}}
\caption{Ensemble mean of facies obtained with the different networks after assimilation of production data. Case 1.}
\label{Fig:Case1-mean}
\end{figure}

Fig.~\ref{Fig:Case1-001} shows the first posterior realization obtained with the different networks combined to ES-MDA. For visual comparison, the corresponding prior realization is also presented in this figure. Clearly, the data assimilation resulted in significant changes in the spatial distribution of the channels observed in the prior realization. Moreover, all cases were able to generate well-defined channels and recover the main features of the reference facies model (Fig.~\ref{Fig:Case1-true}). Fig.~\ref{Fig:Case1-mean} shows the ensemble mean (average values among the $N_e$ realizations using the values one for channels and zero for background). The prior mean is clearly smooth reflecting the high level of uncertainty in the position of the channels. All posterior means show that the main channels of the reference model were correctly captured in all networks. It is interesting to note that even for the GAN with a possible mode collapse, the history matching process was able to generate facies realizations very close to the reference case.

\begin{figure}
\centering
\includegraphics[width=1\linewidth]{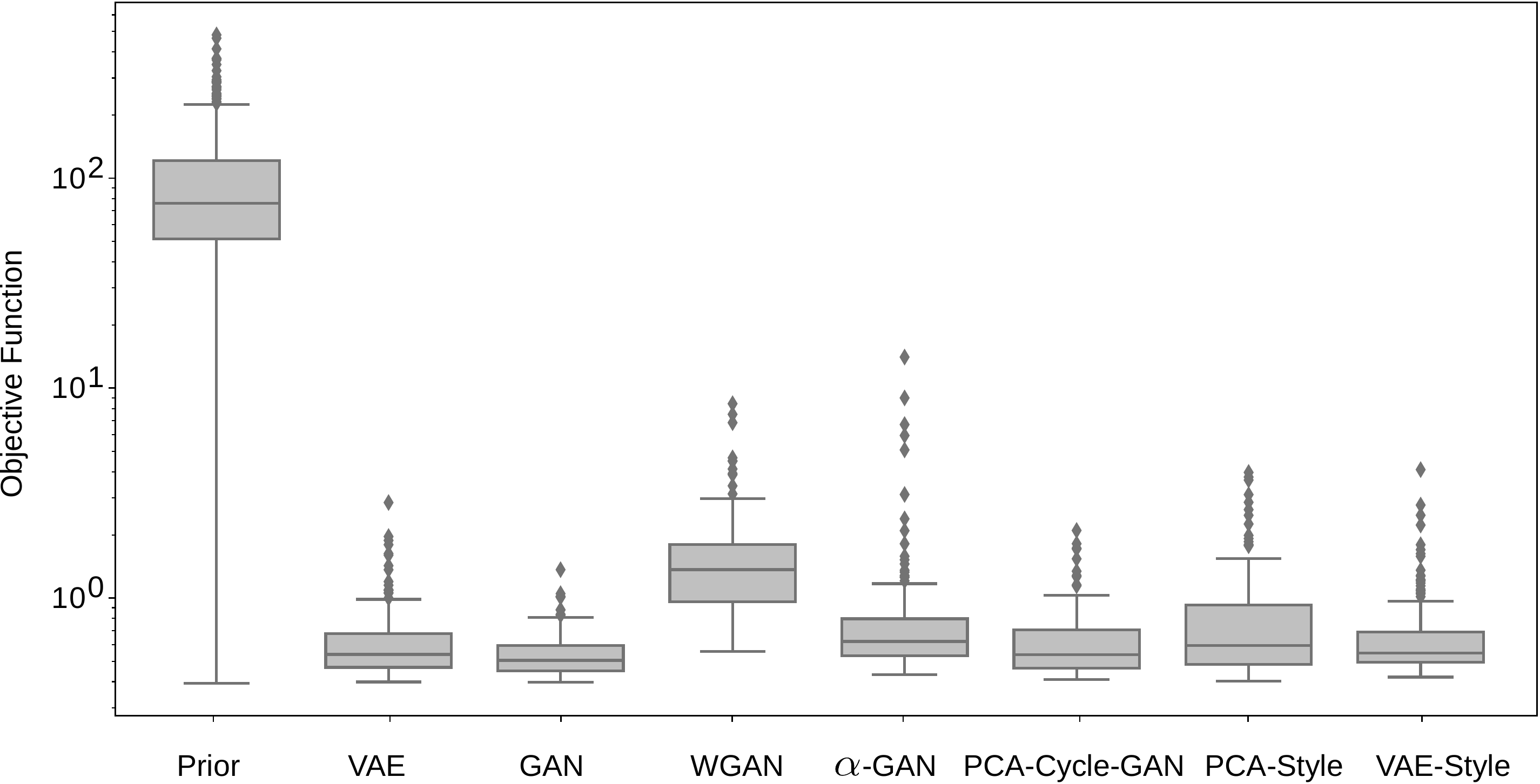}
\caption{Box-plots of normalized data mismatch objective function. Case 1.}
\label{Fig:Case1-BoxPlot}
\end{figure}

\begin{figure}
\centering
  \subfloat[VAE]{\includegraphics[width=0.42\linewidth]{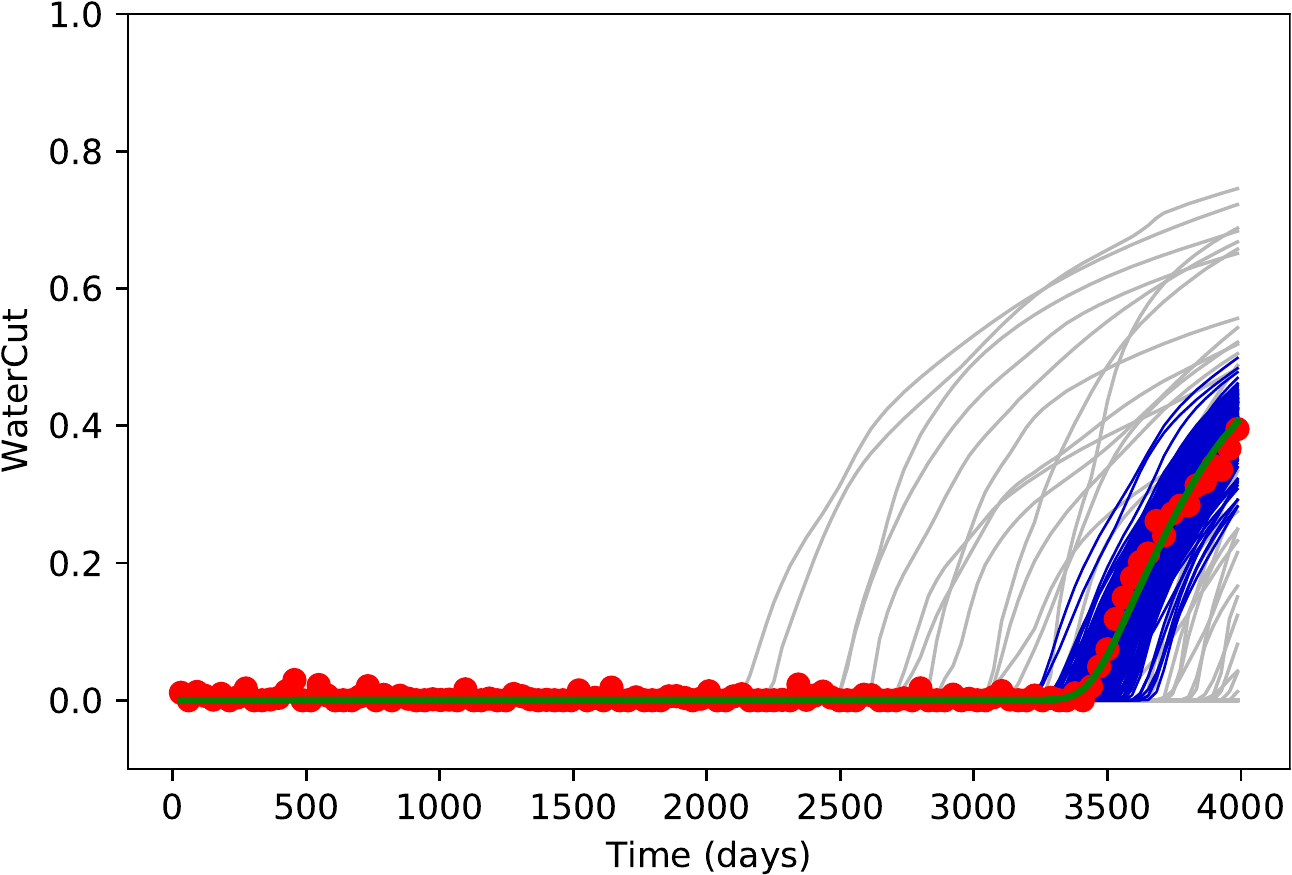}}
  \hspace{3mm}\subfloat[GAN]{\includegraphics[width=0.42\linewidth]{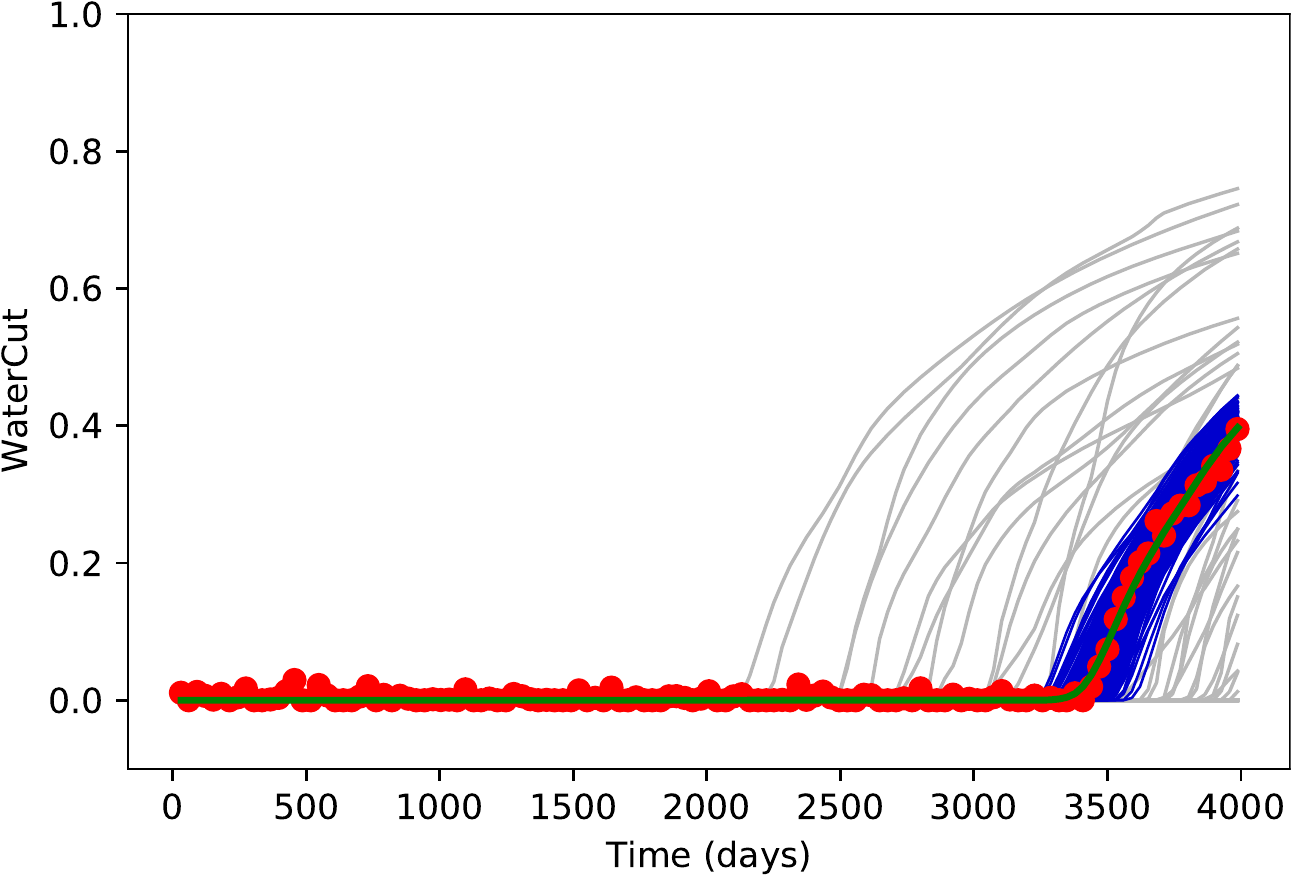}}\\
  \subfloat[WGAN]{\includegraphics[width=0.42\linewidth]{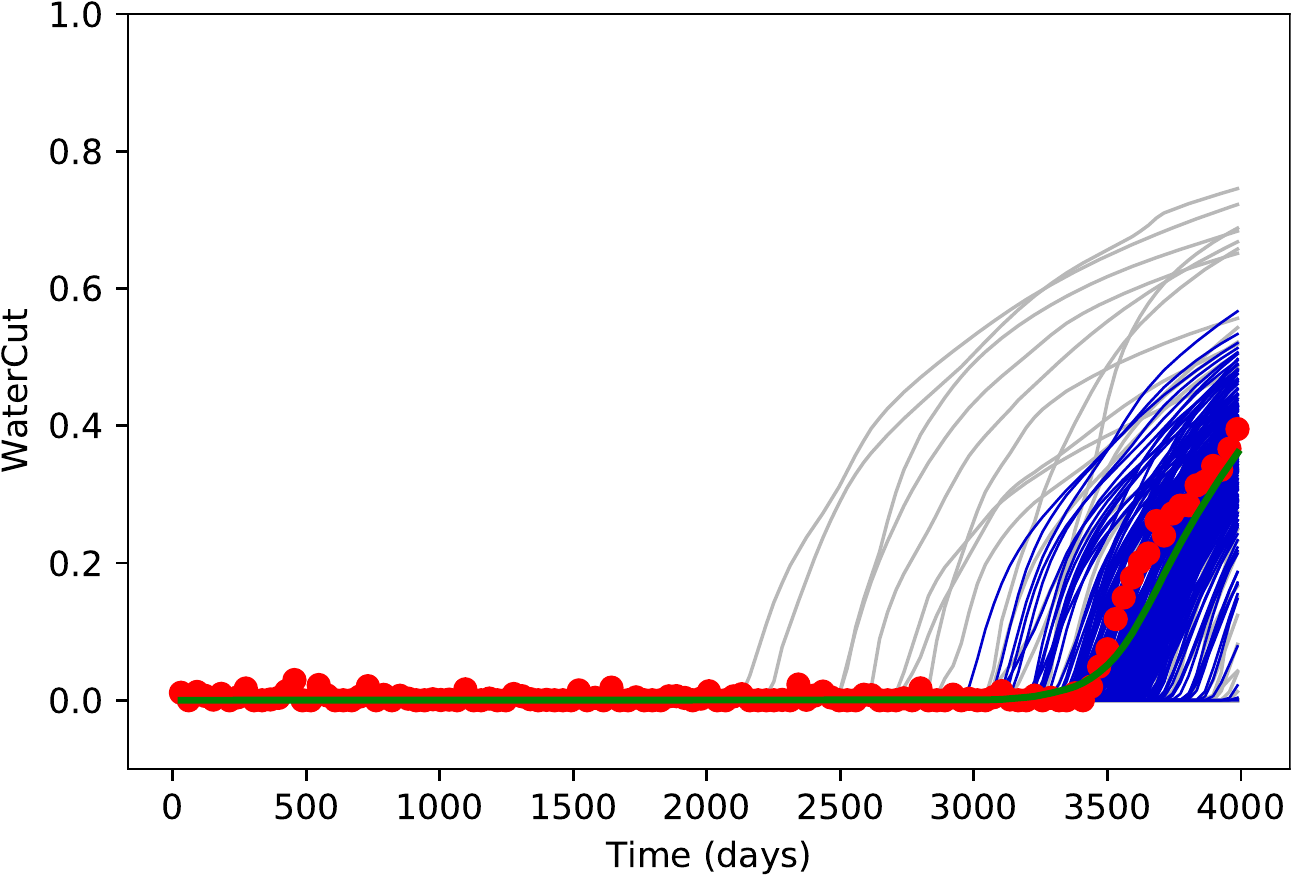}}
  \hspace{3mm}\subfloat[$\alpha$-GAN]{\includegraphics[width=0.42\linewidth]{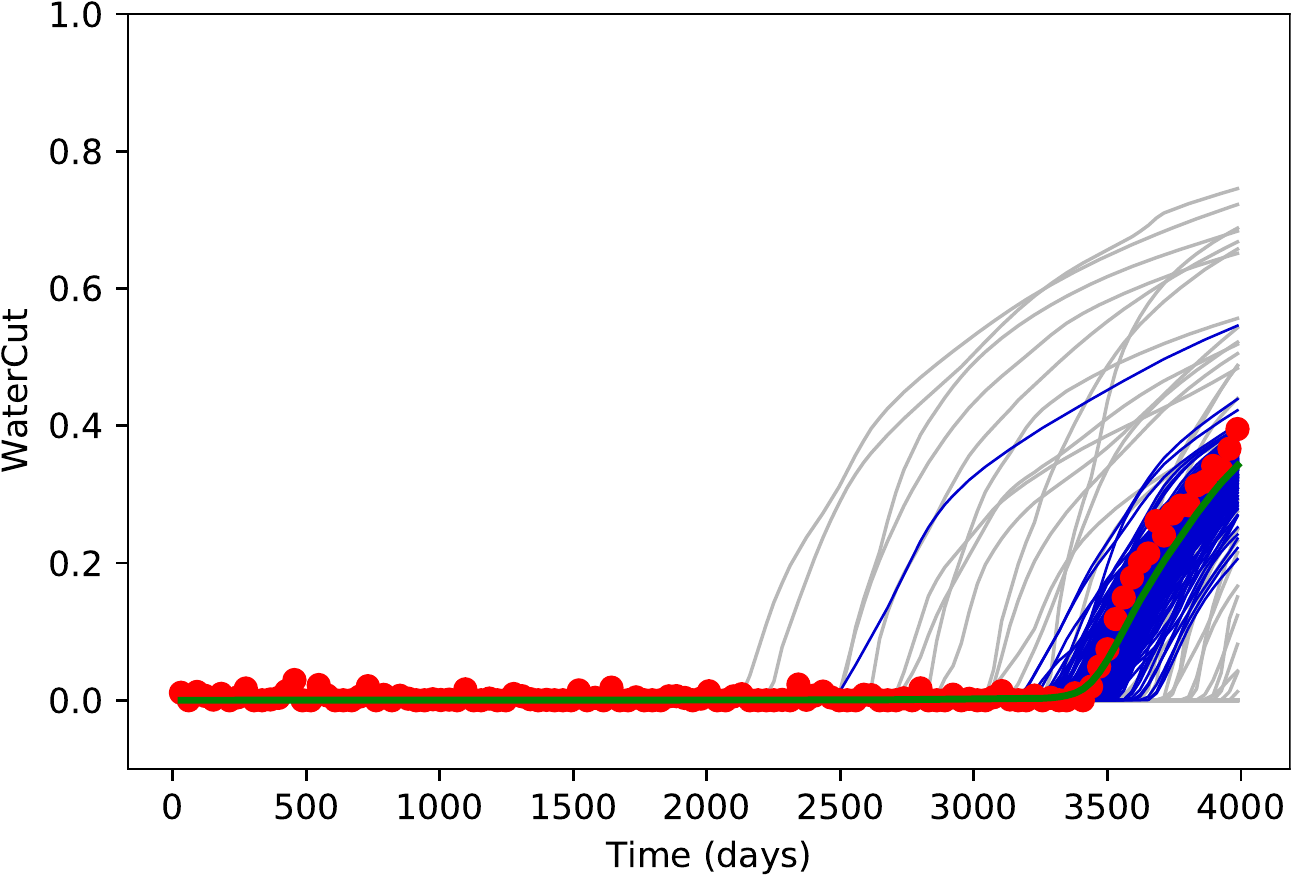}}\\
  \subfloat[PCA-Cycle-GAN]{\includegraphics[width=0.42\linewidth]{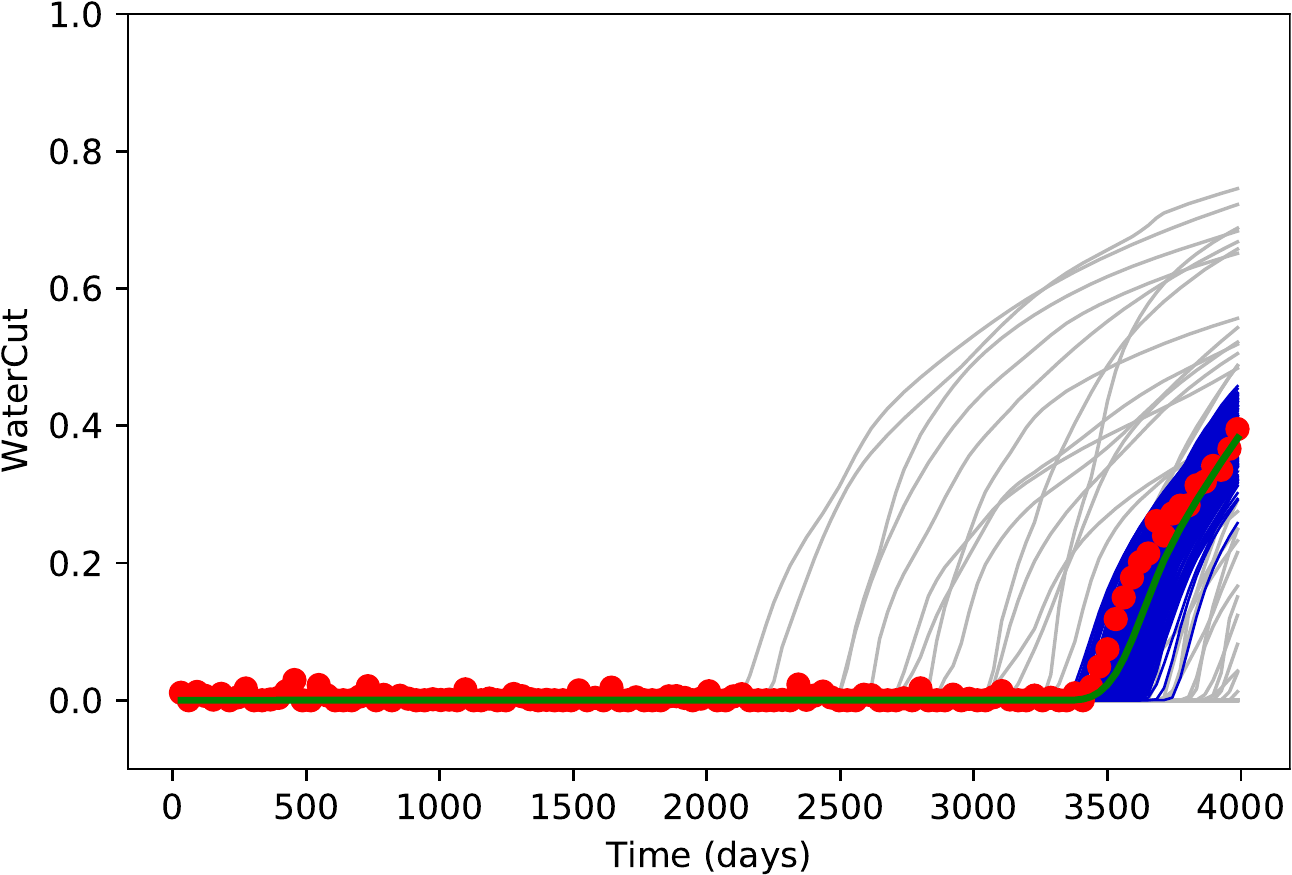}}
  \hspace{3mm}\subfloat[PCA-Style]{\includegraphics[width=0.42\linewidth]{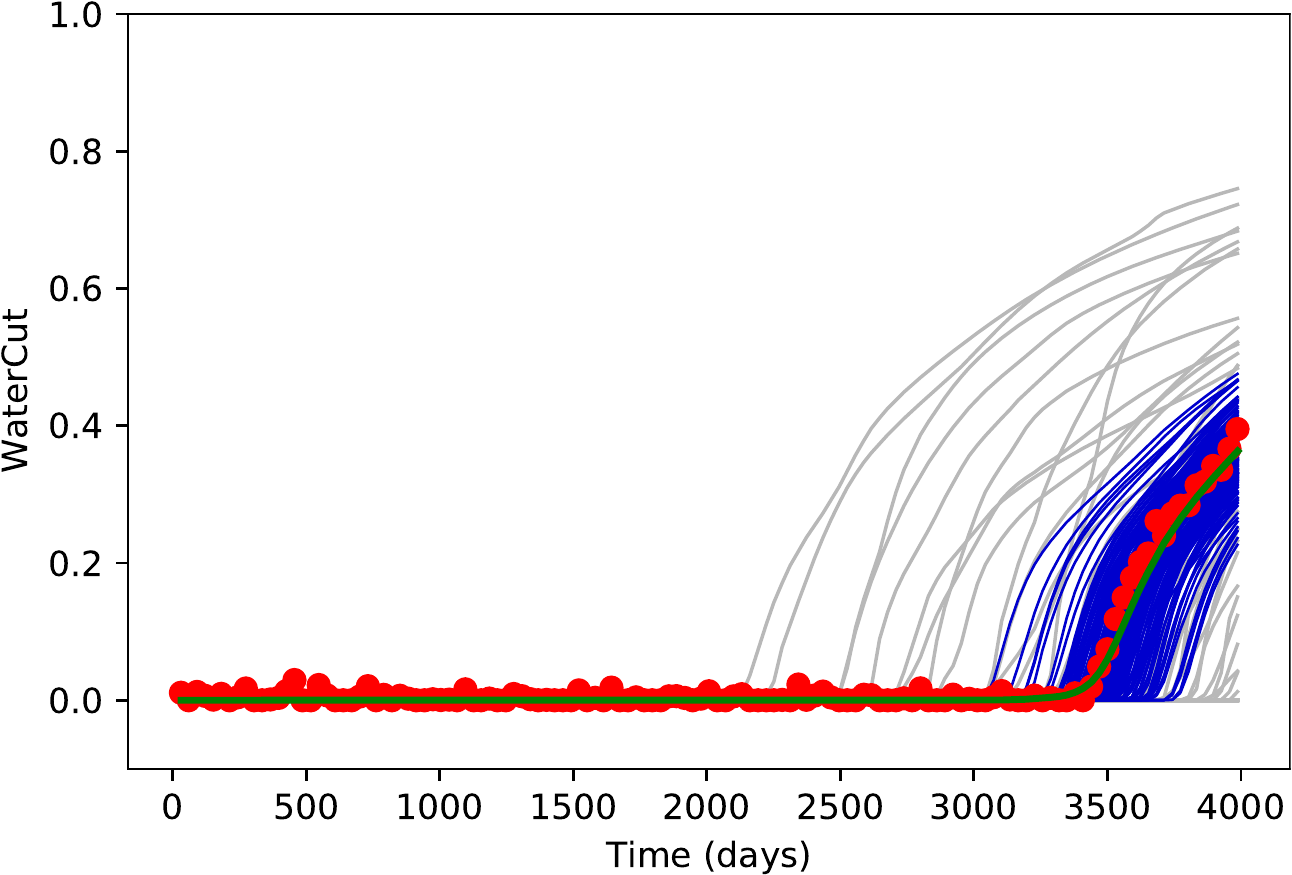}}\\
  \subfloat[VAE-Style]{\includegraphics[width=0.42\linewidth]{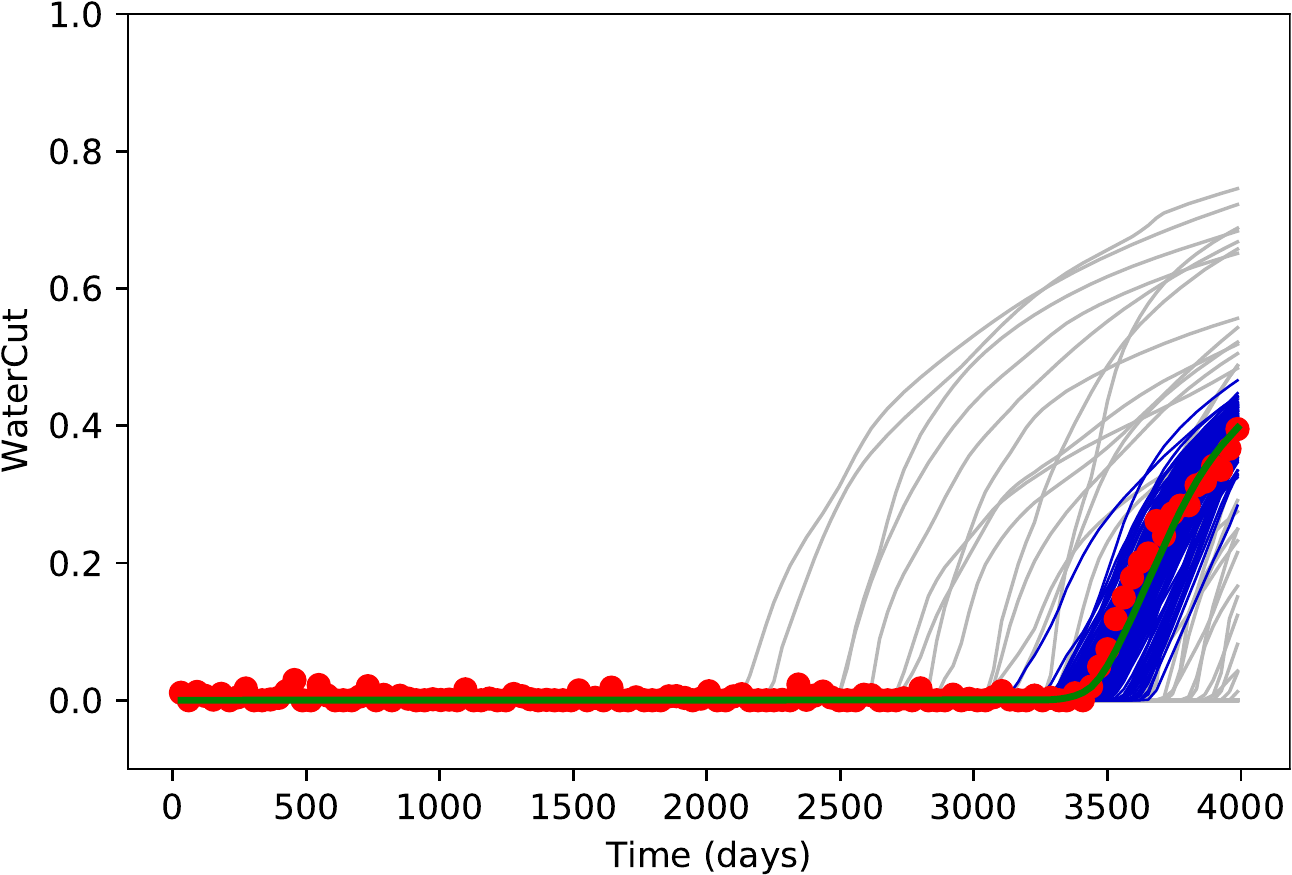}}
\caption{Water cut for well P6. Case 1. The red circles correspond to the observed data points. The gray and blue curves correspond to the predictions from the prior and posterior ensembles, respectively. The green curve is the posterior mean.}
\label{Fig:Case1-P6-WCT}
\end{figure}

In order to evaluate the quality of the data matches, we computed the objective function normalized by the number of data points for each realization using

\begin{equation}\label{Fig:ON}
  O_{N,j}  = \frac{1}{2N_d} \left( \dobs - \dsim_j \right)\trp \Ce\inv \left( \dobs - \dsim_j \right).
\end{equation}
Fig.~\ref{Fig:Case1-BoxPlot} shows the box-plot of $O_{N,j}$ obtained by each method. All cases resulted in significant improvements in the data matches compared to the prior ensemble. WGAN is the case with largest objective function values, which indicates worse data matches. Fig.~\ref{Fig:Case1-P6-WCT} illustrates this fact by showing the predicted water cut at well P6 before and after data assimilations for all cases. This is the well with worse data matches in the model. This figures shows a large spread of predicted water cut for the posterior ensembles obtained with WGAN and $\alpha$-GAN. It is important to note that all cases used eight ES-MDA iterations. The results of Figs.~\ref{Fig:Case1-BoxPlot} and \ref{Fig:Case1-P6-WCT} indicate a slower convergence of the data assimilation using the WGAN parameterization. In fact, we tested to increase the number of ES-MDA iterations with the WGAN and the median objective function reduced from 1.37 with eight iterations to 0.79 with 20 iterations. However, this value is still larger than the other cases. For example, $\alpha$-GAN with eight iterations obtained a median objective function of 0.62. All other cases obtained lower values of objective function with eight iterations. Although the slower convergence of WGAN is evident for this problem, it is not possible, however, to generalize this conclusion based on a single experiment.

\clearpage

\section{Localization Strategies}
\label{Sec:Local}

Localization is an \emph{ad-hoc} method designed to remove long-distance spurious correlations that arise in the computation of the matrices $\C_{\m\dsim}^{k}$ and $\C_{\dsim\dsim}^{k}$ in the ES-MDA update equation (Eq.~\ref{Eq:ES-MDA}) because of the limited size of the ensemble. Localization also increases the degrees of freedom to assimilate data \citep{aanonsen:09}. Localization is typically done using a correlation function based on the spatial distance between model parameters and data points \citep{houtekamer:01}. In practice, the use of localization is mandatory in realistic problems with large number of data points; see, \citep{emerick:16a} for an example of field application where localization was required to obtain reasonable estimates of petrophysical properties.

The parameterizations strategies discussed in the previous sections of this paper are based on the use of generative networks that are able to create facies realizations, $\x$, based on a latent representation, $\z$. The latent representation is updated with ES-MDA to incorporate the production data, as illustrated in the workflow shown in Fig.~\ref{Fig:DL-ESMDA}. However, the latent vector $\z$ losses the spatial relation in the model such that it no longer possible to compute the spatial distance between a well, which represents the spatial location of the data, and a model parameter (entry of the vector $\z$). For this reason, the direct application of distance-based localization is not possible. It is worth mentioning that there are more general localization methods which do not rely on the spatial distances; see, e.g., \citep{lacerda:19a} and references therein. However, our experience with practical applications of ES-MDA for history matching indicate that these methods are less effective than distance-based approaches.

In the following, we present two simple strategies to allow the use of distance-based localization in conjunct with the deep learning parameterizations. The first strategy is based on local analysis \citep{anderson:03,hunt:07} and can be applied to any parameterization discussed in this paper. The second strategy is based on using a symmetric square root of the covariance matrix for PCA and it is used with the Cycle-GAN network.

\subsection{VAE with Local Analysis}
\label{Sec:VAE-Local}

\begin{figure}
\centering
\includegraphics[width=0.4\linewidth]{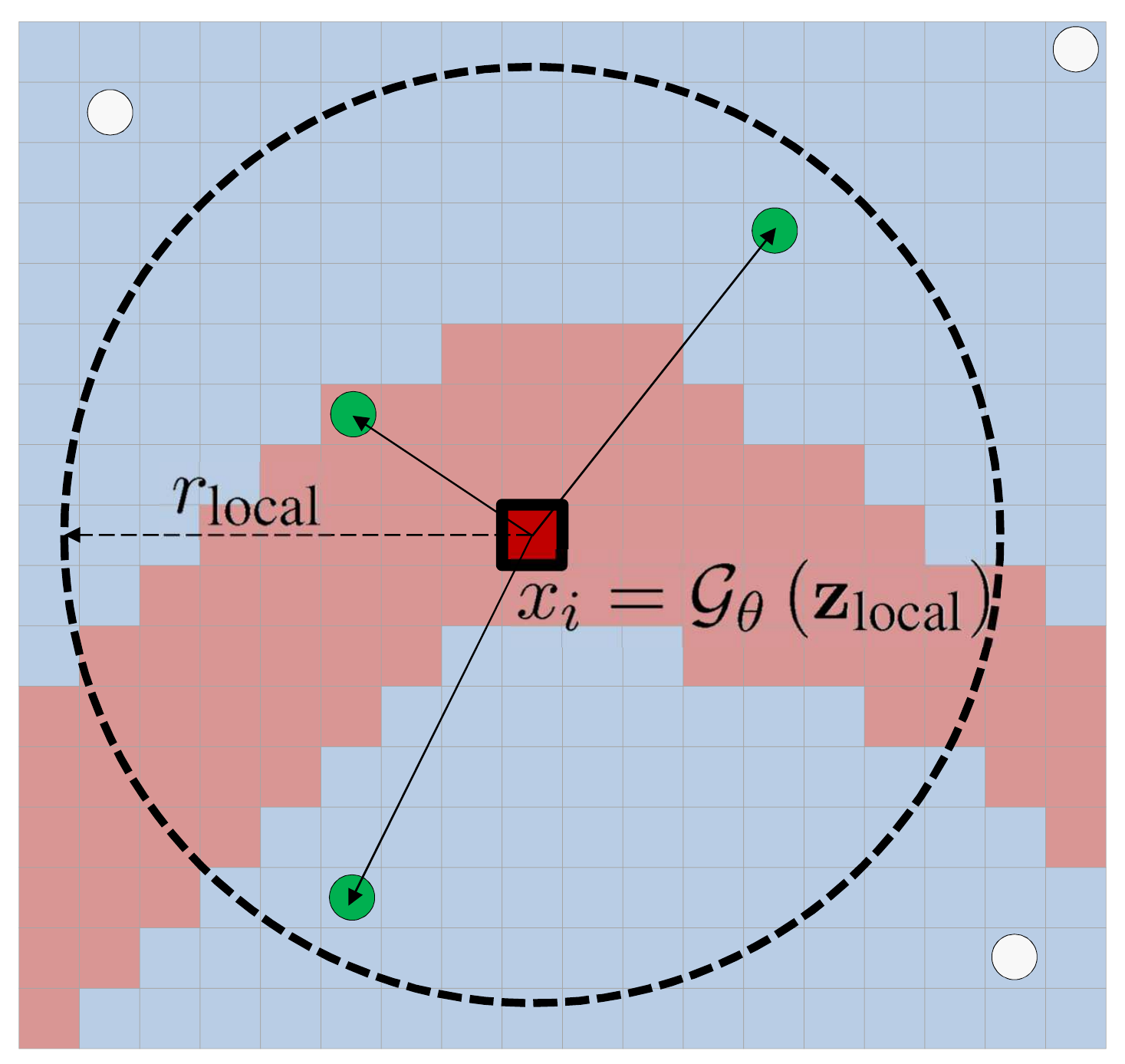}
\caption{Local analysis scheme. The facies at the $i$th gridblock, $x_i$, is obtained applying the generative netwrk with a latent vector $\z_{\textrm{local}}$ updated using only the observations within the localization region with radius $r_{\textrm{local}}$. The green circles indicate the position of the wells used in the analysis. Data from these wells are weighted based on the distance to the gridblock $i$ using Eq.~\ref{Eq:CeLocal}}
\label{Fig:LocalAnalysis}
\end{figure}

The first localization strategy consists of applying an independent ES-MDA analysis to the vector $\z$ for estimating the facies for each gridblock, $x_i$, of the model. Each analysis uses only the observations contained within the neighborhood region around the corresponding gridblock. Because the local regions overlap, we expect that the transition between analyses to be smooth. Moreover, for each analysis we use a local data-error covariance matrix $\C_{\mathbf{e},\textrm{local}}$ by tapering $\Ce\inv$ with a correlation function. This tapering is done using a Schur product with a localization matrix, i.e.,

\begin{equation}\label{Eq:CeLocal}
  \C_{\mathbf{e},\textrm{local}}\inv = \R \circ \Ce\inv,
\end{equation}
where $\R$ is a localization matrix whose entries are calculated using the Gaspari-Cohn correlation function \citep{gaspari:99} based on the distance between the gridblock where the facies is updated and the data location. This procedure effectively increase the data-error covariance of distant observations, limiting their influence in the facies estimation. After updating the latent vector $\z_{\textrm{local}}$ the facies is obtained using the trained generative network, i.e., $x_i = \mathcal{G}_\theta(\z_{\textrm{local}})$. This process is illustrated in Fig.~\ref{Fig:LocalAnalysis}. Note that this process require a large number of applications of the ES-MDA analysis equation, one for each gridblock of the model, which make the process computationally intensive. We minimized this problem by developing a GPU-version of the analysis using \verb"TensorFlow".

\subsection{PCA-Cycle-GAN with Localization}
\label{Sec:PCA-Cycle-GAN-Local}

\begin{figure}
\centering
  \subfloat[$\x_1$]{\includegraphics[width=0.18\linewidth]{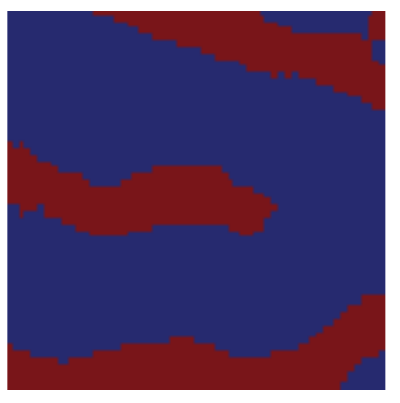}}
  \hspace{3mm}\subfloat[$\x_2$]{\includegraphics[width=0.18\linewidth]{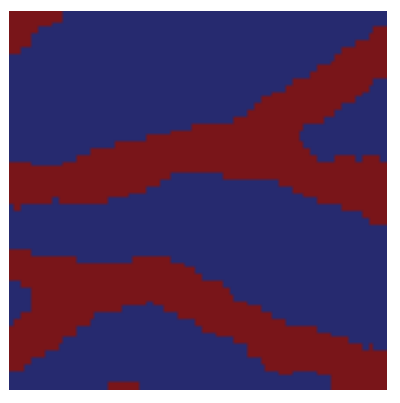}}
  \hspace{3mm}\subfloat[$\x_3$]{\includegraphics[width=0.18\linewidth]{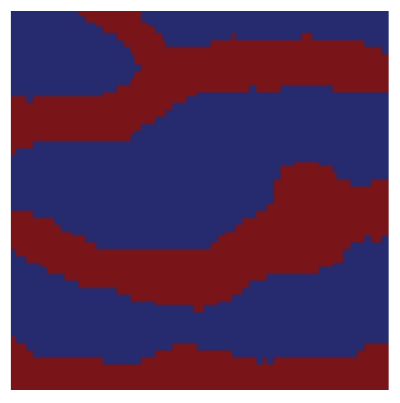}}\\
  \subfloat[$\z_1$]{\includegraphics[width=0.18\linewidth]{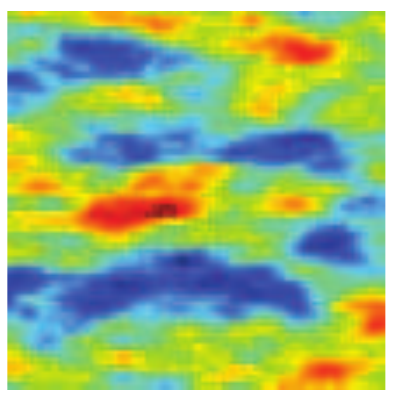}}
  \hspace{3mm}\subfloat[$\z_2$]{\includegraphics[width=0.18\linewidth]{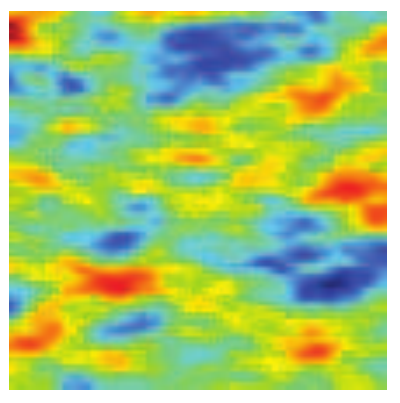}}
  \hspace{3mm}\subfloat[$\z_3$]{\includegraphics[width=0.18\linewidth]{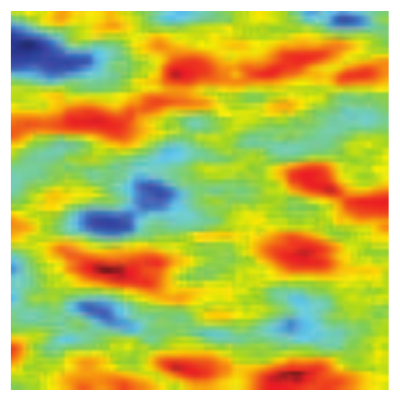}}\\
\caption{Three facies realizations $\x_i$ and the corresponding PCA coefficients $\z_i$.}
\label{Fig:FaciesToPCA}
\end{figure}

The second localization strategy is more straightforward but less general. It is based on our PCA-Cycle-GAN implementation with a simple trick to project the PCA coefficients in the same space of the facies model. Recall that Cycle-GANs learn mappings from discrete facies $\x$ to continuous representations $\z$ and \emph{vice versa}; see Fig.~\ref{Fig:cyclegans}. In our implementation, $\z$ are the PCA coefficients computed using a symmetric square root as described in \citep{emerick:17a}. This procedure leads to a vector of PCA coefficients with the same dimension (number of gridblock) of the facies. Fig.~\ref{Fig:FaciesToPCA} illustrates three facies realizations, $\x_1$, $\x_2$ and $\x_3$ and their corresponding PCA coefficients $\z_1$, $\z_2$ and $\z_3$. The PCA coefficients images exhibit hotter colors at the position of the channels and colder colors in the background. These are the vectors updated with ES-MDA, in which case we apply the standard Schur product-based localization \citep{aanonsen:09}.

\section{Test Case 2: Comparison of Localization Strategies}
\label{Sec:Case2}

\begin{figure}[!h]
    \centering
    \includegraphics[width=0.8\textwidth]{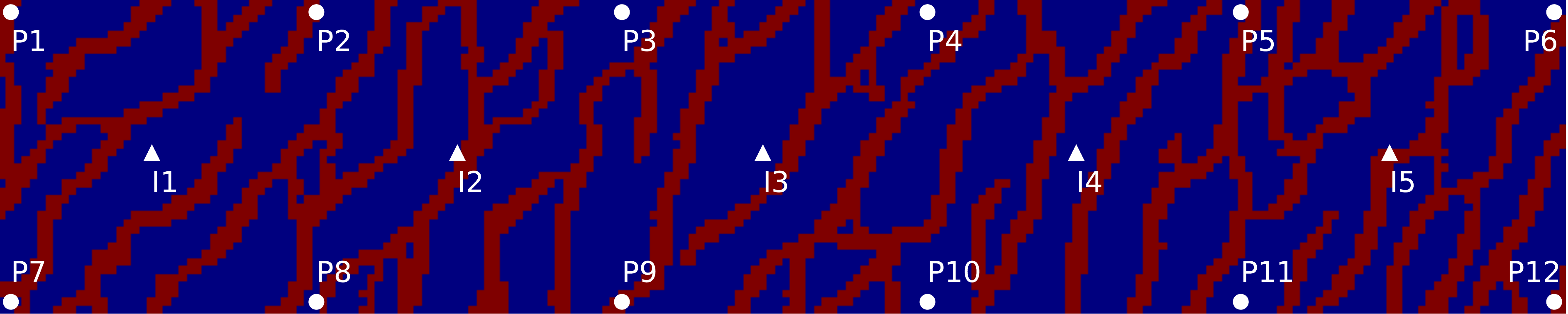}
    \caption{Reference facies model. Red colors correspond to high-permeability channels and blue the background sand. Case 2.}
    \label{Fig:Case2-true}
\end{figure}

The second test problem is also a synthetic case with two facies, channel with constant permeability of 1000~mD and background sand with permeability of a 100~mD. This is an extended version of the first problem with 200~$\times$~40 gridblocks, with a large number of channels and wells. In this reservoir there are 12 oil producing and five water injection wells placed in five-spots. This problem was designed to emphasise the need of a localization strategy to avoid a severe variance reduction due to the large number of measurements assimilated. Fig.~\ref{Fig:Case2-true} shows the reference facies which was generated using the MPG algorithm \verb"snesim".

\subsection{Training Process}
\label{Sec:TranningCase2}

\begin{figure}
  \centering
  \subfloat[Training]{\includegraphics[width=0.49\linewidth]{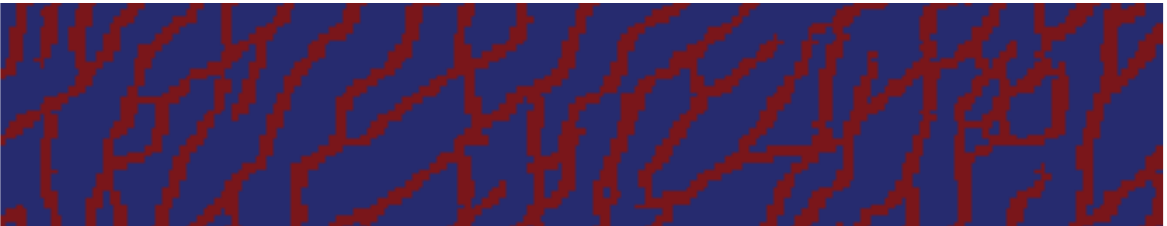}
  \hspace{2mm}\includegraphics[width=0.49\linewidth]{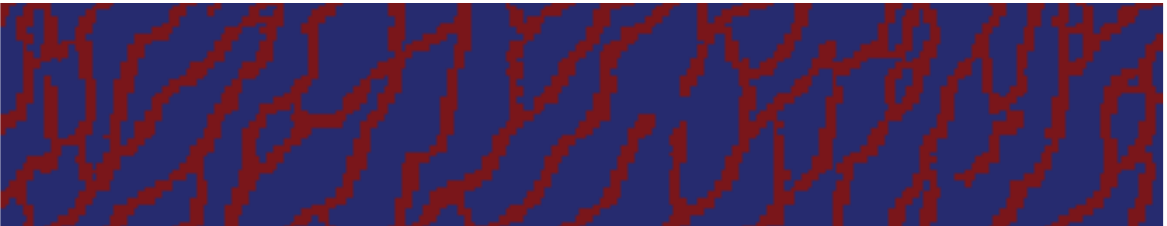}}\\
  \subfloat[VAE]{\includegraphics[width=0.49\linewidth]{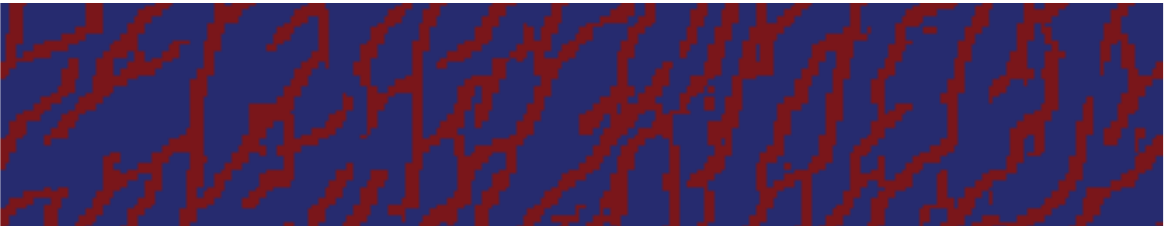}
  \hspace{2mm}\includegraphics[width=0.49\linewidth]{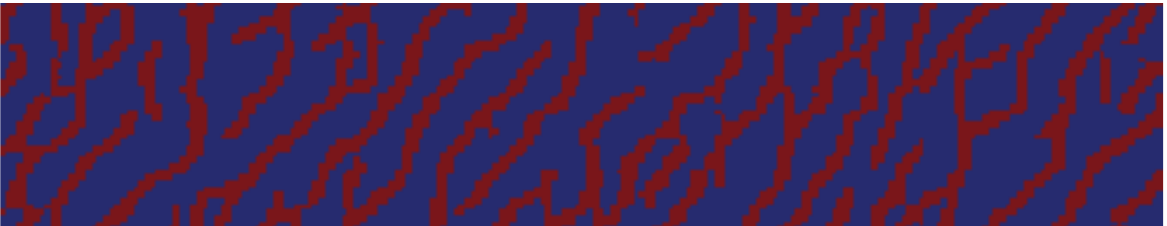}}\\
  \subfloat[PCA-Cycle-GAN]{\includegraphics[width=0.49\linewidth]{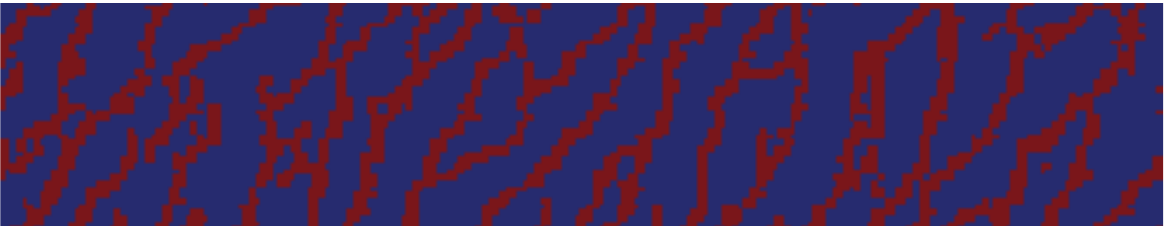}
  \hspace{2mm}\includegraphics[width=0.49\linewidth]{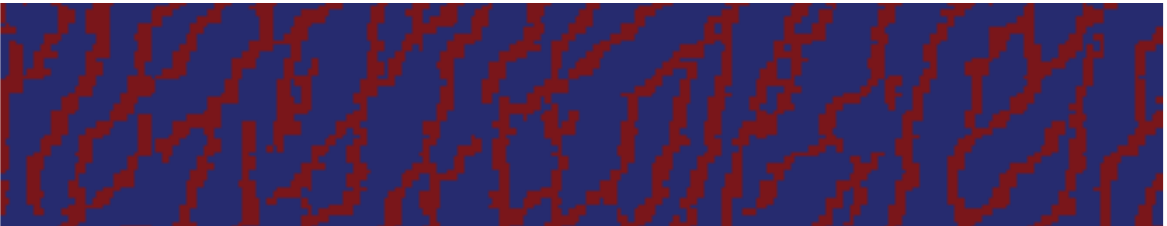}}
\caption{Random realizations of facies from the training set and generated with the VAE and PCA-Cycle-GAN. Case 2.}
\label{Fig:Case2-RandomSamples}
\end{figure}

The training process of the VAE network used $50000$ realizations of facies divided in $70\%$ for training and $30\%$ for validation and latent dimension of $N_z=1024$. The PCA for generating the prior latent realizations for PCA-Cycle-GAN used a covariance matrix estimated with 5000 samples and another 5000 random realizations for training the Cycle-GAN. Fig.~\ref{Fig:Case2-RandomSamples} shows two realizations of the prior training set and two random realizations generated with the trained VAE and PCA-Cycle-GAN networks. This figure shows that both networks were able to generate realistic facies realizations with the same characteristics of realizations from the training set.

\subsection{Assimilation of Production Data}
\label{Sec:AssimilationProdData}

In this case we use a prior ensemble with $N_e = 200$ realizations and $N_a = 16$ iterations in the ES-MDA without any facies data (hard data). We consider four data assimilation cases

\begin{itemize}
    \item VAE: the same VAE parameterization tested in the previous test case without localization.
    \item VAE-Local: VAE combined with the local analysis procedure.
    \item VAE-Local-VAE: VAE parameterization with local analysis followed by a pass in the VAE network to suppress noise in the facies realizations.
    \item PCA-Cycle-GAN-Local: PCA-Style-GAN with the Schur product localization.
\end{itemize}
The localization region corresponds to an ellipse centered at the updated gridblock (local analysis) or data location (Schur product localization) with major axis corresponding to 80 gridblocks and the minor axis of 20 gridblocks. The ellipse is rotated at 60$^\text{o}$ to follow the direction of larger continuity of the channels. In both cases, we use the Gaspari-Cohn correlation function. Similarly to the case 1, all oil producing wells are controlled by a constant BHP of 150~bars, while the injectors are controlled by a constant BHP of 350~bars. The observed data used in the history matching correspond the monthly measurements of water cut and water injection rate for a period of 10 years. The synthetic measurements were generated adding random noise to the data predicted by the reference model with standard deviation corresponding to 5\% of the data values.

\begin{figure}
  \centering
  \subfloat[Reference]{\includegraphics[width=0.49\linewidth]{case2_true.pdf}}
  \hspace{2mm}\subfloat[Prior]{\includegraphics[width=0.49\linewidth]{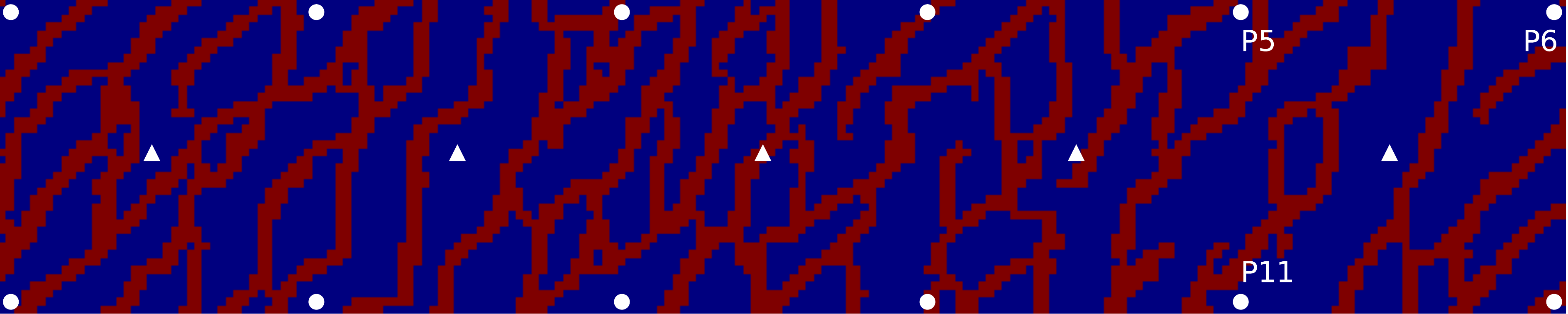}}\\
  \subfloat[VAE]{\includegraphics[width=0.49\linewidth]{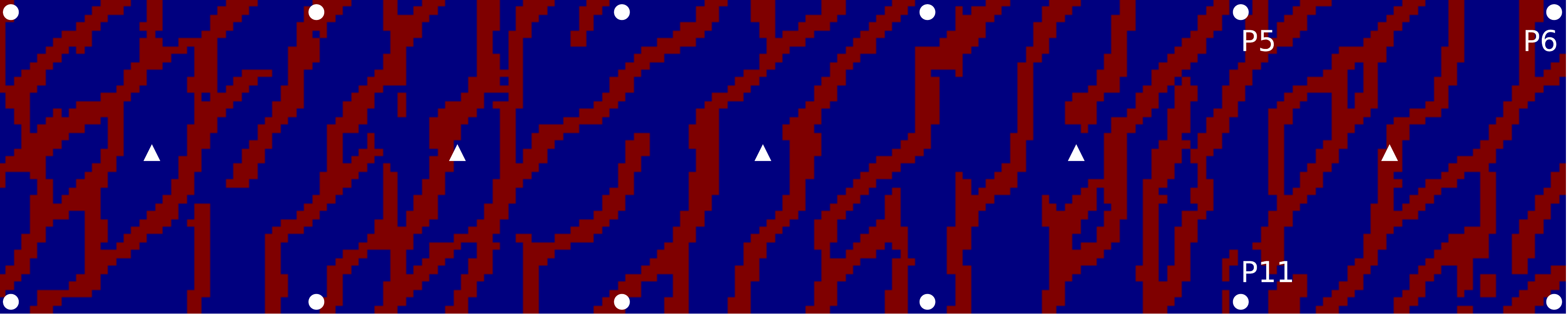}}
  \hspace{2mm}\subfloat[VAE-Local]{\includegraphics[width=0.49\linewidth]{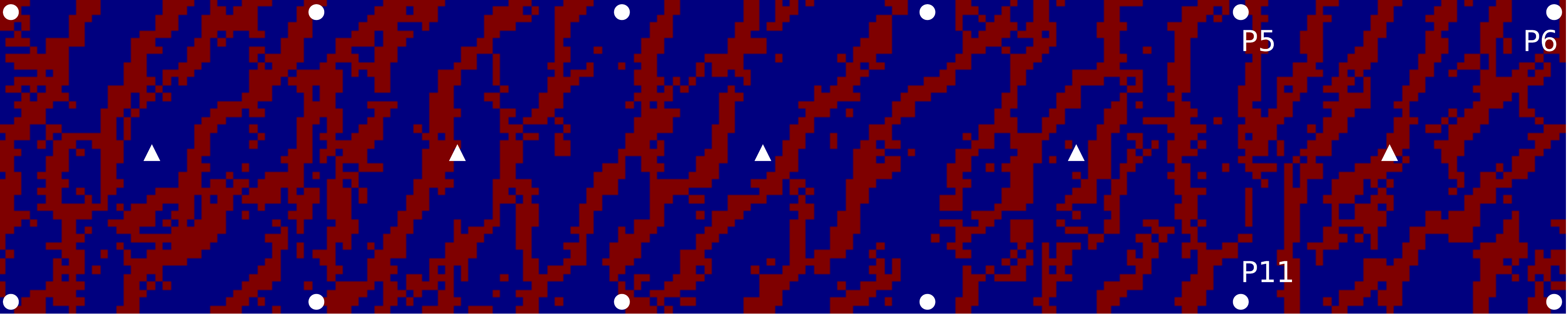}}\\
  \subfloat[VAE-Local-VAE]{\includegraphics[width=0.49\linewidth]{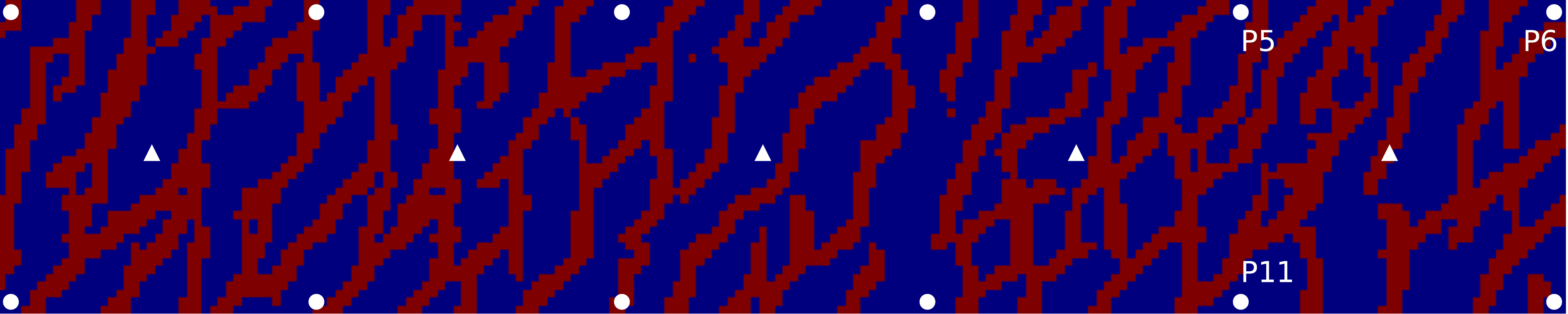}}
  \hspace{2mm}\subfloat[PCA-Cycle-GAN-Local]{\includegraphics[width=0.49\linewidth]{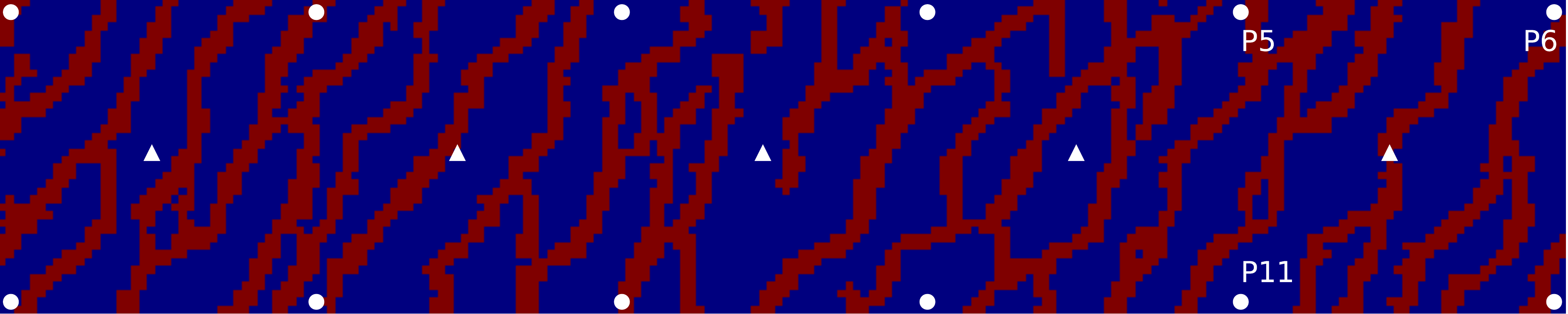}}
\caption{First realization of facies obtained with the different networks. Case 2.}
\label{Fig:Case2-001}
\end{figure}

\begin{figure}
  \centering
  \subfloat[Prior]{\includegraphics[width=0.49\linewidth]{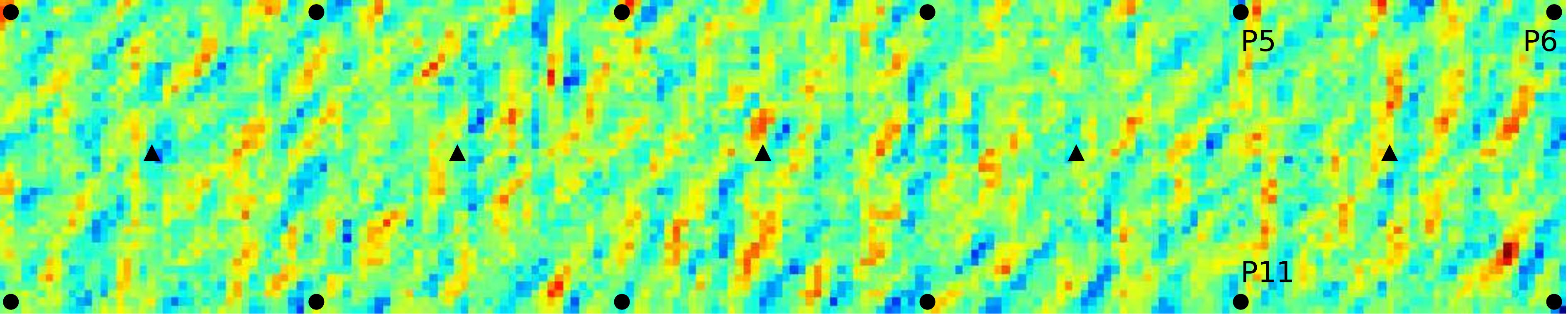}}\\
  \subfloat[VAE]{\includegraphics[width=0.49\linewidth]{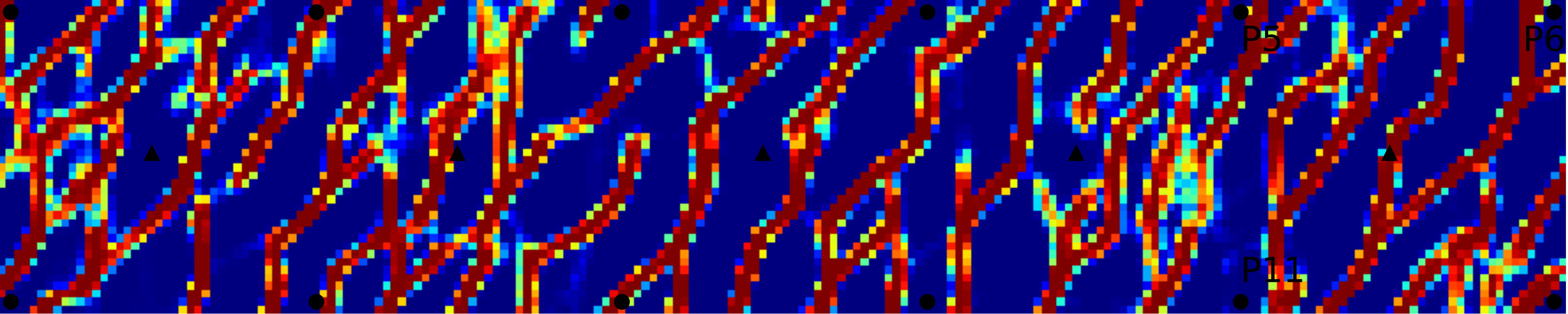}}
  \hspace{2mm}\subfloat[VAE-Local]{\includegraphics[width=0.49\linewidth]{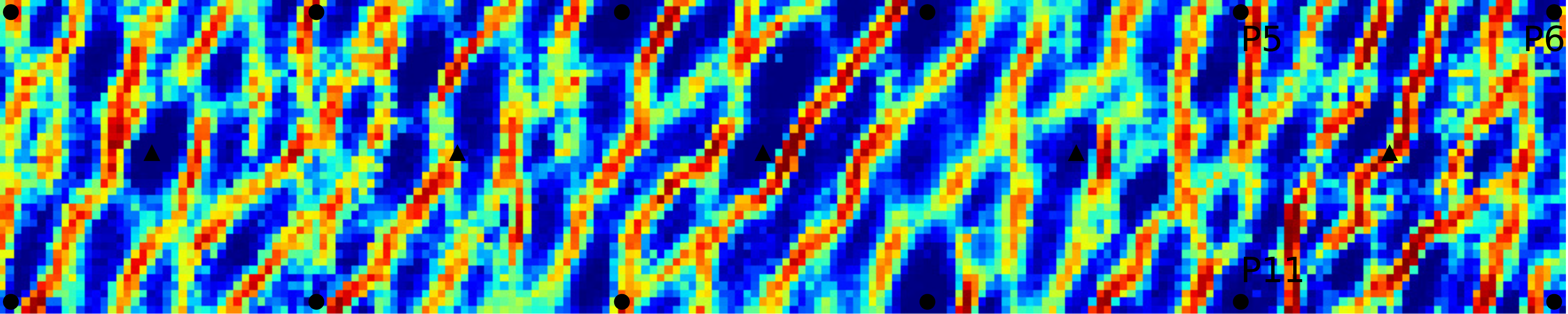}}\\
  \subfloat[VAE-Local-VAE]{\includegraphics[width=0.49\linewidth]{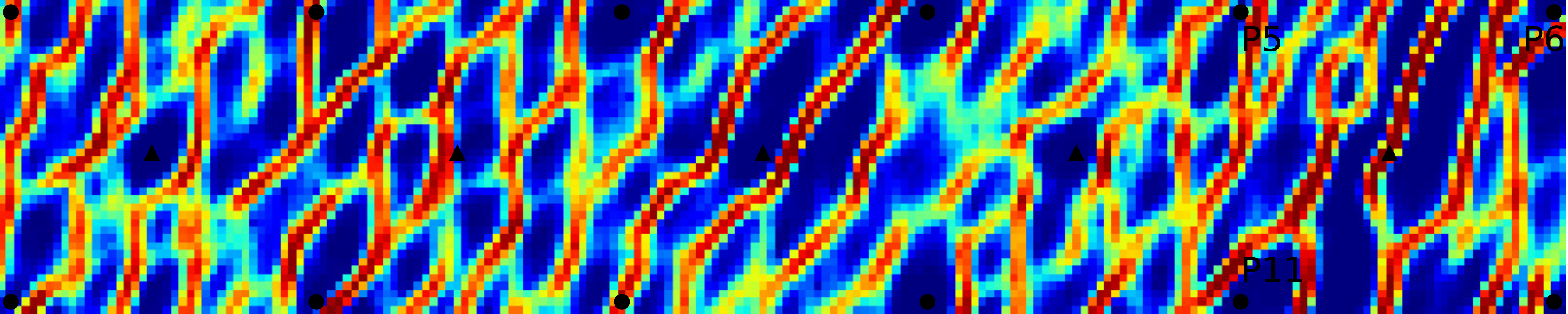}}
  \hspace{2mm}\subfloat[PCA-Cycle-GAN-Local]{\includegraphics[width=0.49\linewidth]{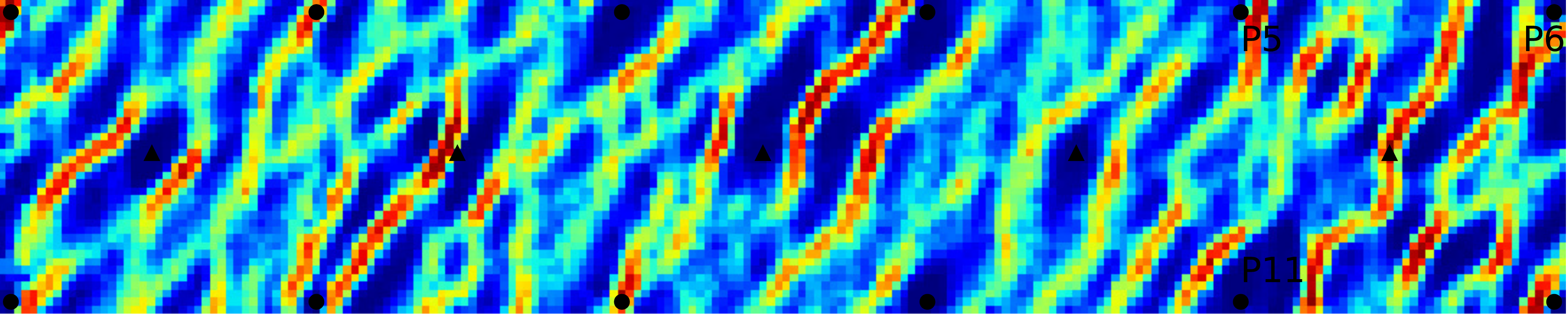}}
\caption{Ensemble mean of facies obtained with the different networks. Case 2.}
\label{Fig:Case2-mean}
\end{figure}

\begin{figure}
  \centering
  \subfloat[VAE]{\includegraphics[width=0.49\linewidth]{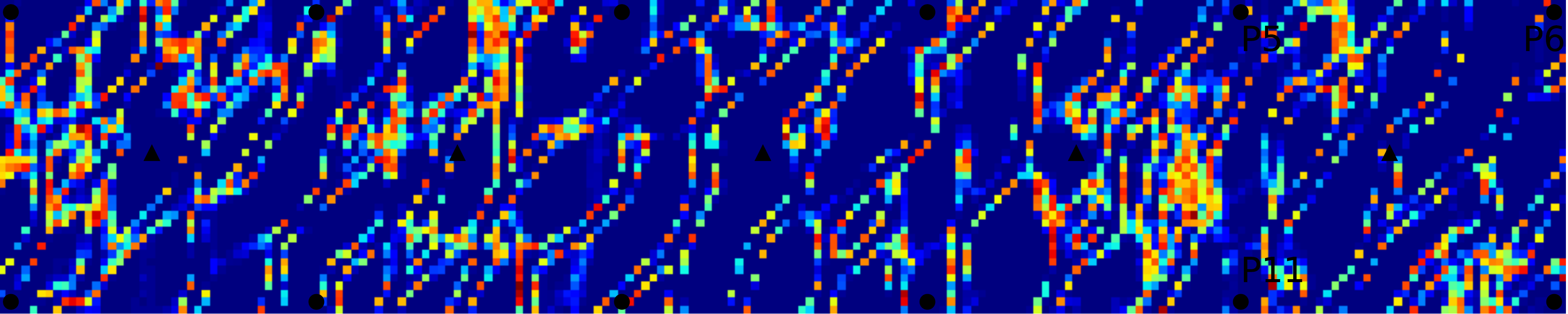}}
  \hspace{2mm}\subfloat[VAE-Local]{\includegraphics[width=0.49\linewidth]{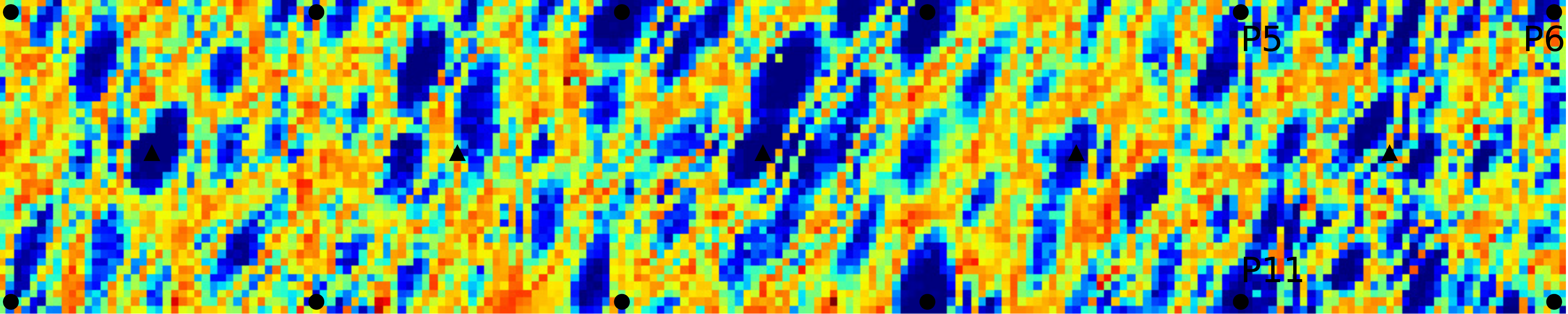}}\\
  \subfloat[VAE-Local-AE]{\includegraphics[width=0.49\linewidth]{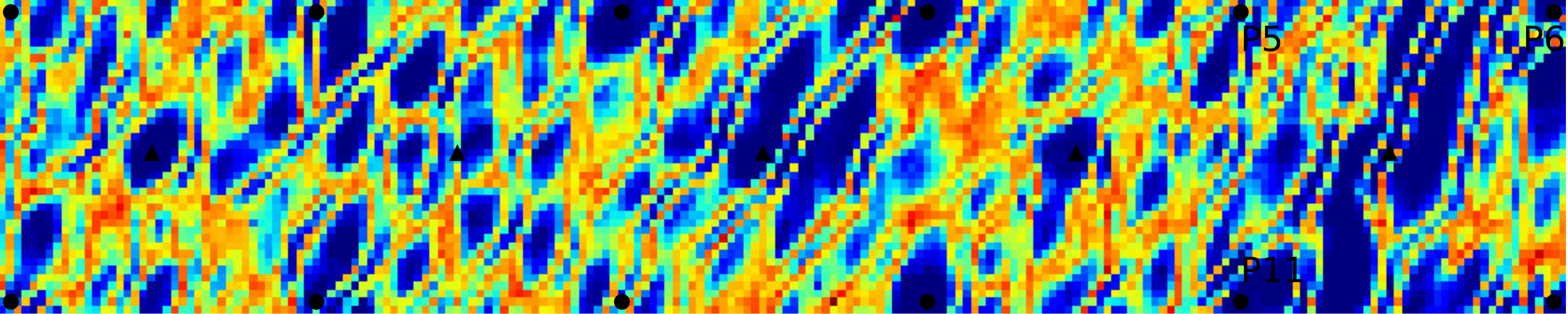}}
  \hspace{2mm}\subfloat[PCA-Cycle-GAN-Local]{\includegraphics[width=0.49\linewidth]{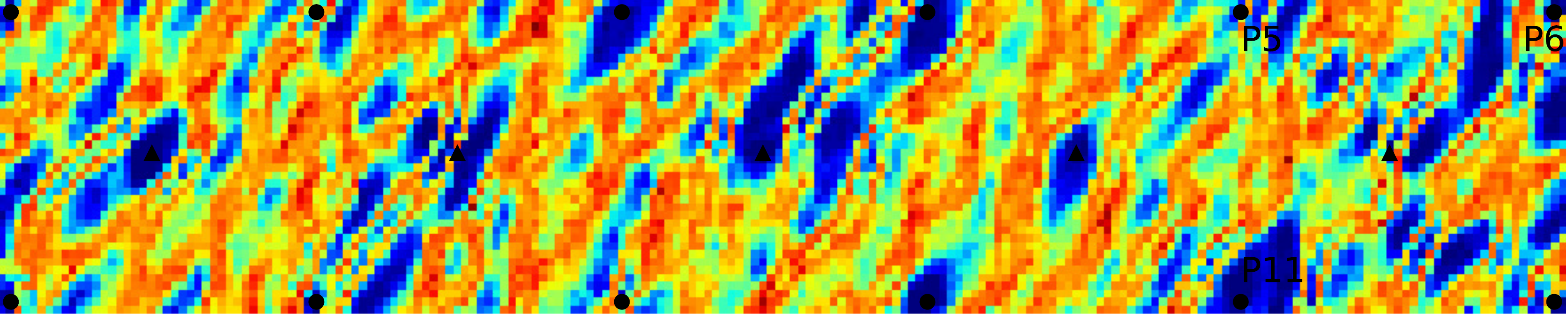}}
\caption{Normalized ensemble variance of log-permeability obtained with the different networks. Case 2.}
\label{Fig:Case2-nvar}
\end{figure}

Fig.~\ref{Fig:Case2-001} shows the first posterior realization obtained by each data assimilation strategy. For visual comparisons, we also included the reference facies and the corresponding prior realization. All cases were generated binary facies realizations with the well-defined channels. However, the VAE-Local case (Fig.~\ref{Fig:Case2-001}d) resulted into a ``noisy'' facies distribution as a consequence of the local analysis updating strategy. This result motivated the introduction of the case labeled as VAE-Local-VAE. In this case, after each ES-MDA iteration we feed the resulting facies realization to the same trained VAE for noise removal, resulting in more plausible facies distribution (Fig.~\ref{Fig:Case2-001}e). Fig.~\ref{Fig:Case2-mean} show the prior and the posterior ensemble mean of facies. The prior mean is relatively smooth making difficult to identify the channels (Fig.~\ref{Fig:Case2-mean}a). This occurs because the large uncertainty level in the position of the channels. The positions of the channels are easier to identify in the posterior realizations. In fact, the VAE case (Fig.~\ref{Fig:Case2-mean}b) resulted in well-defined channels boundaries, which indicates that all posterior realization for this case have very similar channel distribution. The cases with localization show a smeared transition in the channels, indicating some variability in the posterior ensembles. Fig.~\ref{Fig:Case2-nvar} shows images of normalized variance (ratio between posterior and prior variances). The normalized variance can be used as an approximate measure of reduction in uncertainty \citep{oliver:08bk}. Fig.~\ref{Fig:Case2-nvar}a shows that the VAE (without localization) resulted in almost a collapse of the ensemble variance, which demonstrates the importance of using some localization strategy. The cases VAE-Local, VAE-Local-VAE and PCA-Cycle-GAN-Local resulted in larger values of normalized variance, which means that there are more distinct facies realization in these posterior ensembles. The PCA-Cycle-GAN-Local is the case with large values of normalized variances. Unfortunately, the correct level of variance is unknown. However, it is well-known that ensemble-based methods have the tendency to underestimate the posterior variance, even when localization is used; see, e.g., \citep{emerick:16b} where it is shown that ES-MDA with localization and an ensemble of 100 realizations resulted in underestimation of posterior variance for the PUNQ-S3 case \citep{floris:01}. Hence, a larger ensemble variability after data assimilation tends to be desirable, especially if this ensemble will be used for production forecast or even for assimilation of further data.

\begin{figure}
\centering
\includegraphics[width=1\linewidth]{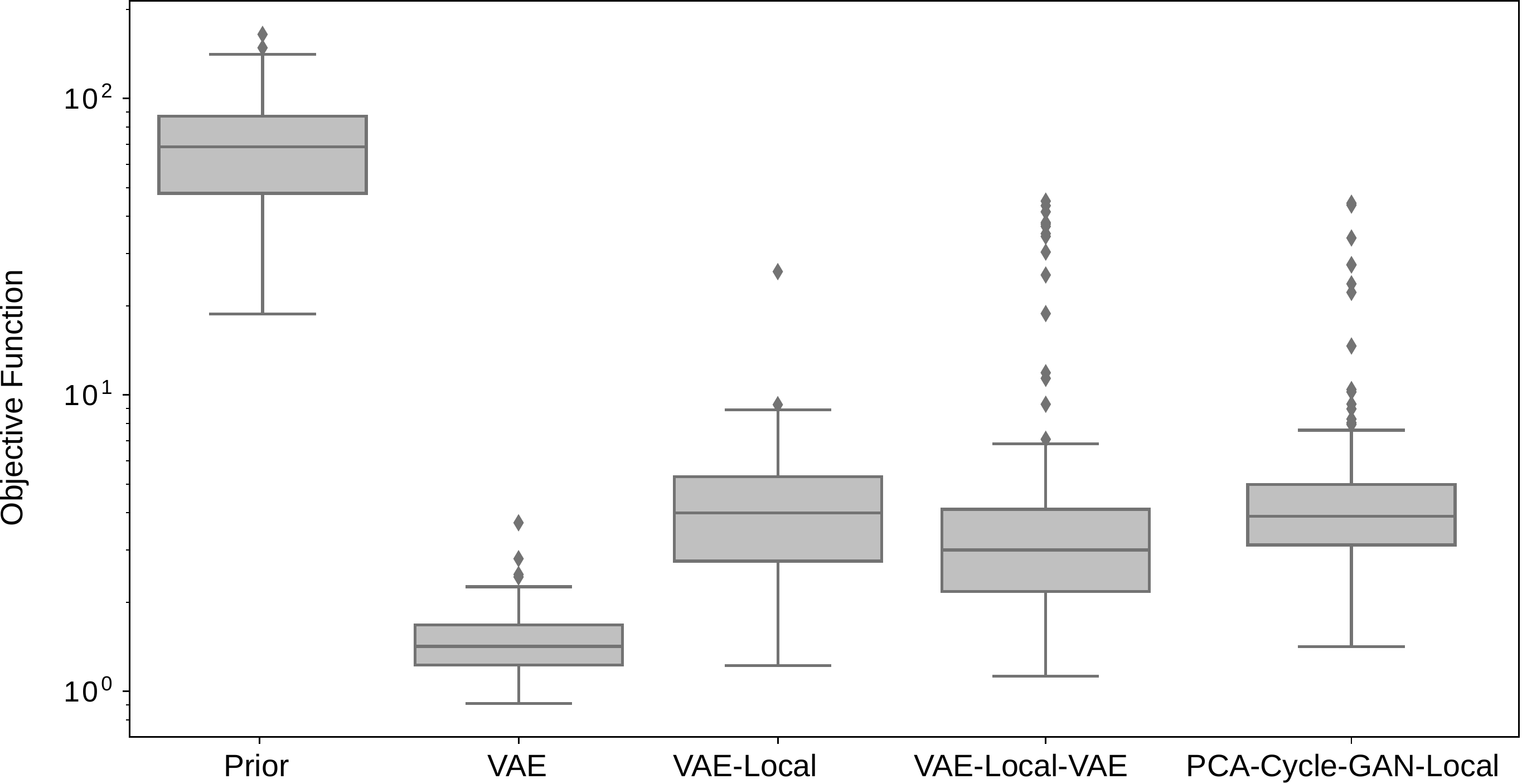}
\caption{Box-plots of normalized data mismatch objective function. Case 2.}
\label{Fig:Case2-BoxPlot}
\end{figure}

\begin{figure}
\centering
\subfloat[VAE]{\includegraphics[width=0.42\linewidth]{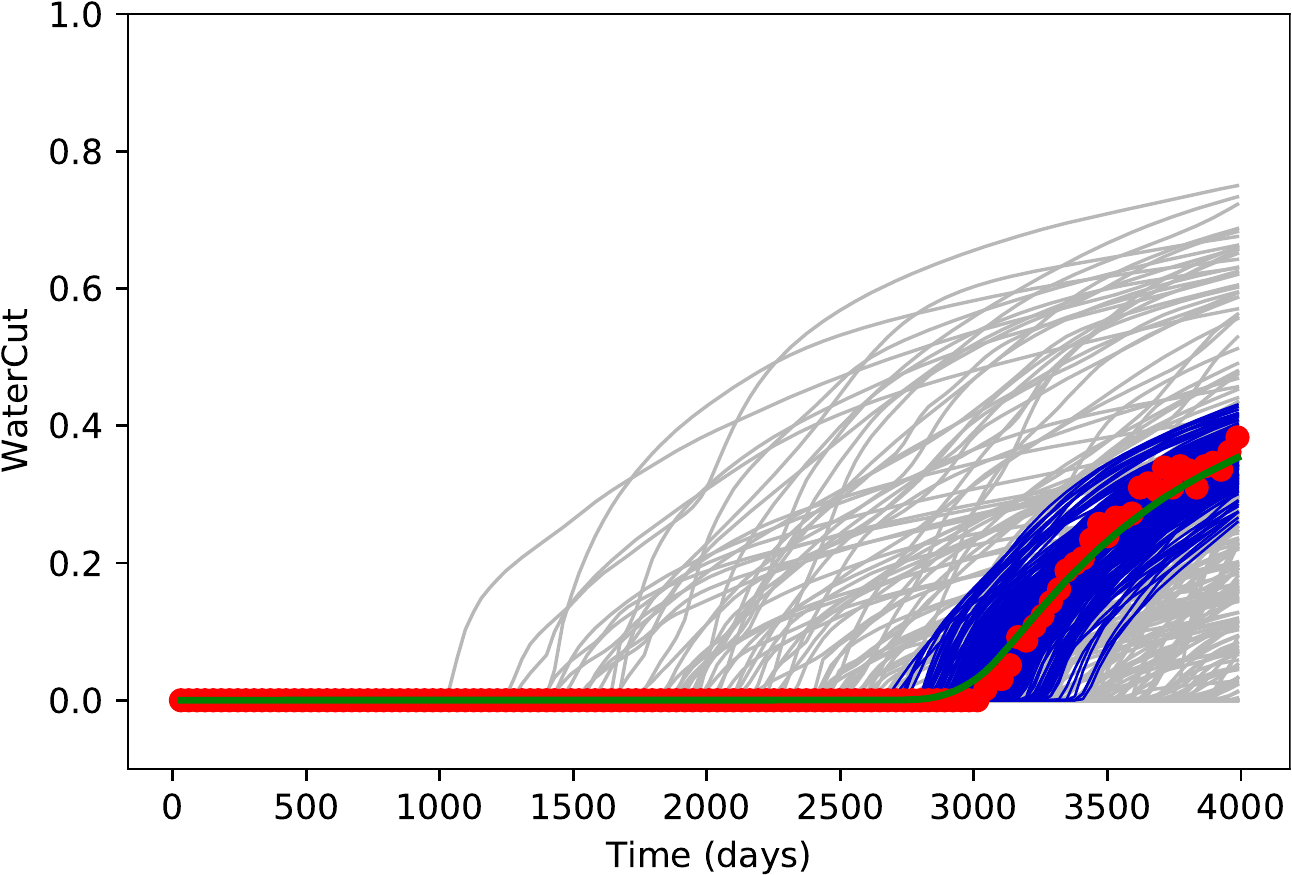}~\includegraphics[width=0.42\linewidth]{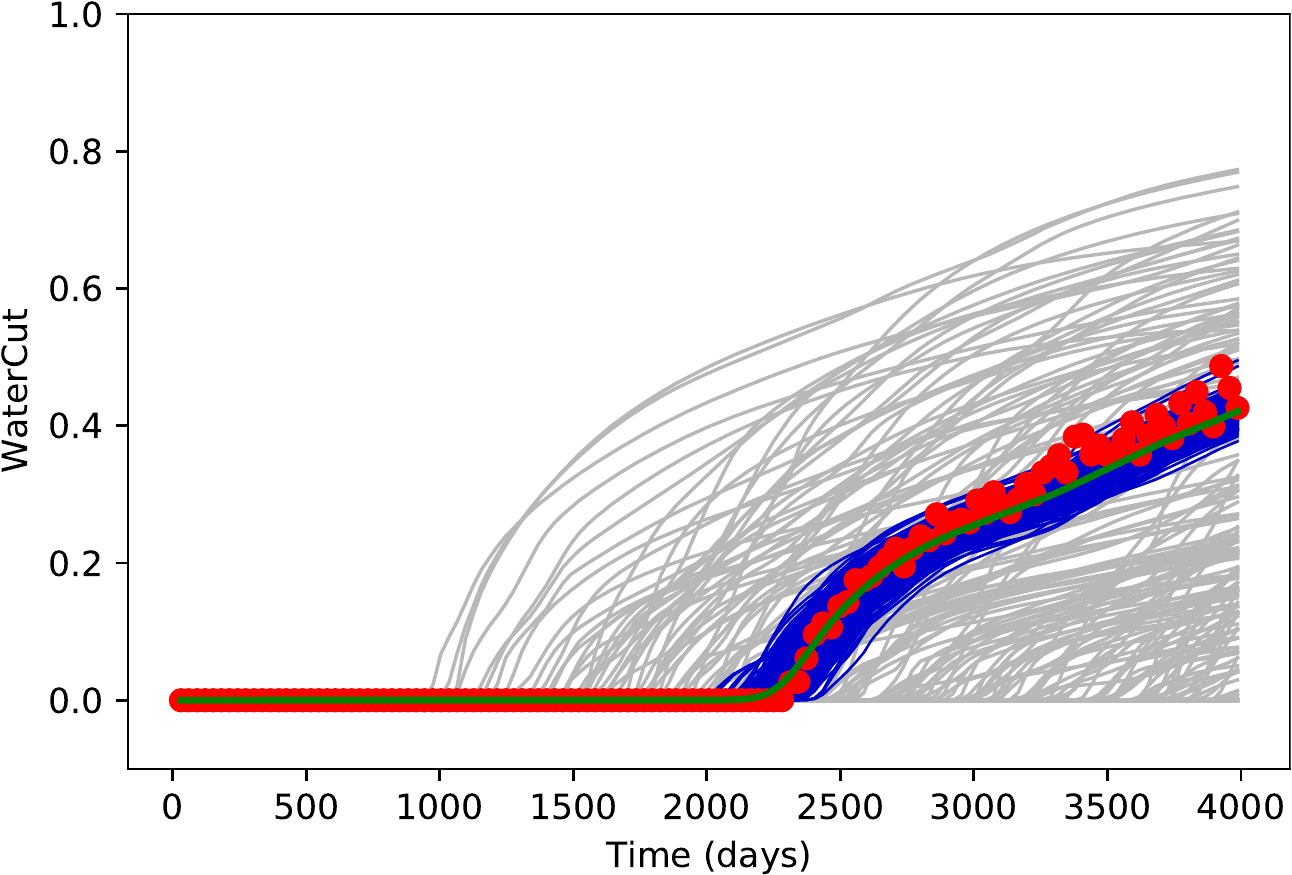}}\\
\subfloat[VAE-Local]{\includegraphics[width=0.42\linewidth]{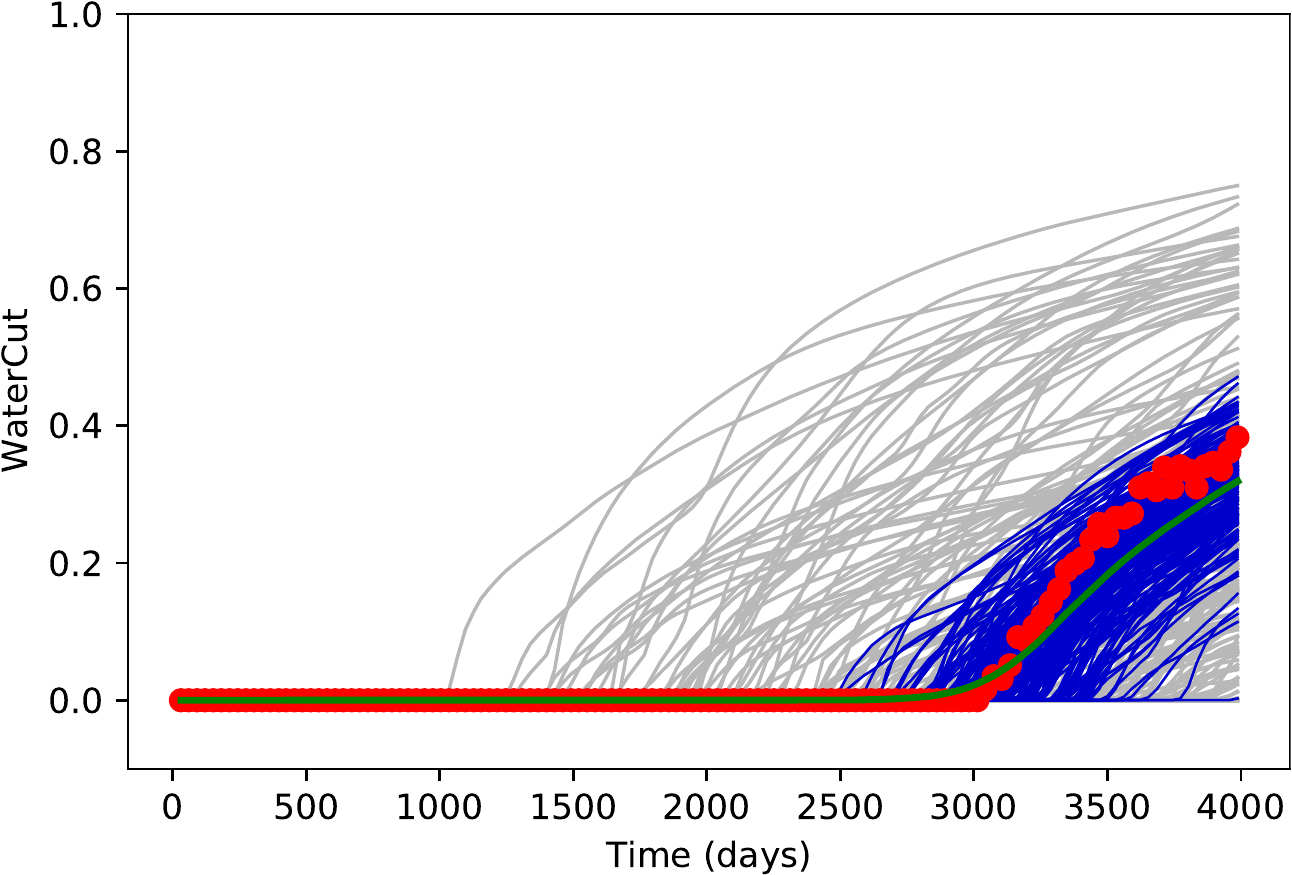}~\includegraphics[width=0.42\linewidth]{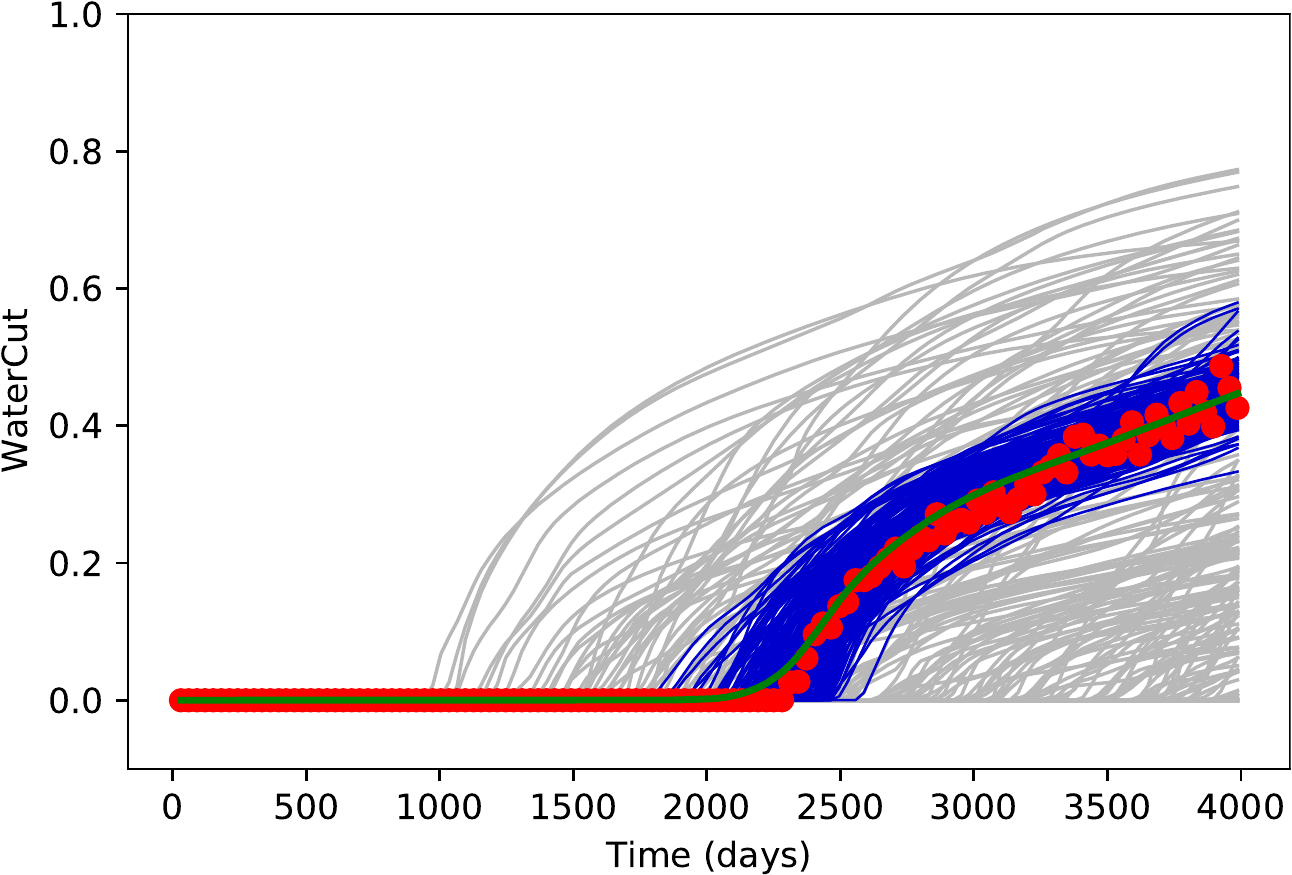}}\\
\subfloat[VAE-Local-VAE]{\includegraphics[width=0.42\linewidth]{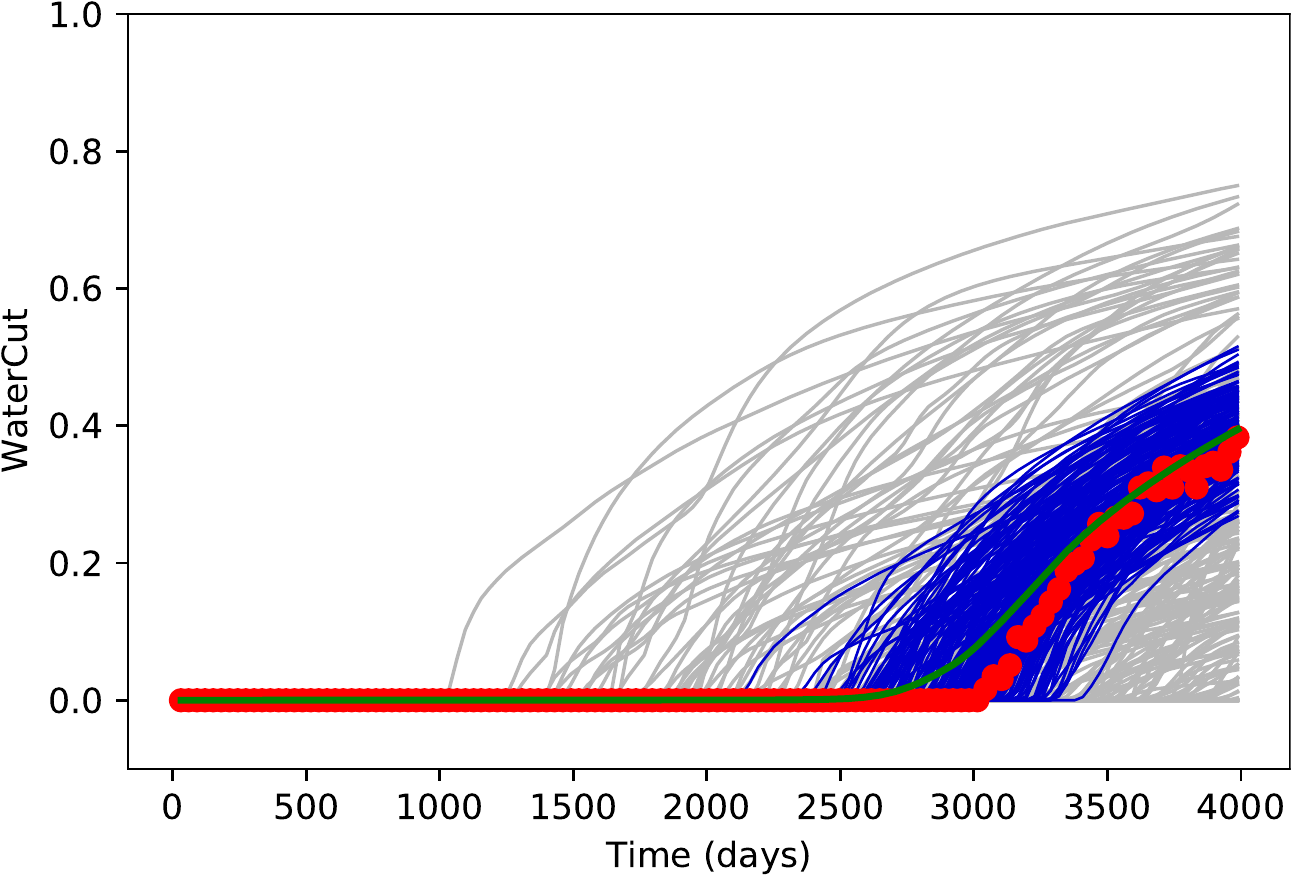}~\includegraphics[width=0.42\linewidth]{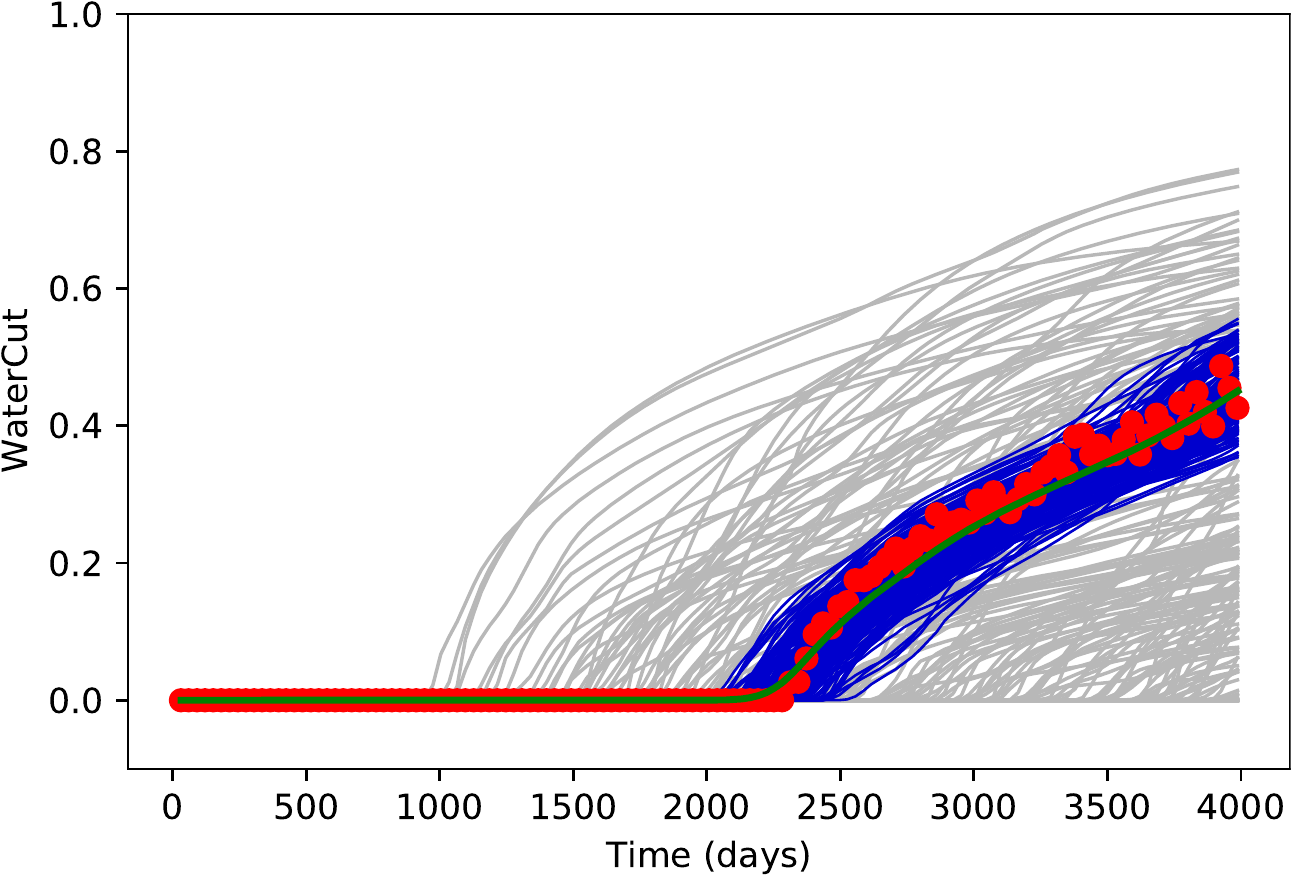}}\\
\subfloat[PCA-Cycle-GAN-Local]{\includegraphics[width=0.42\linewidth]{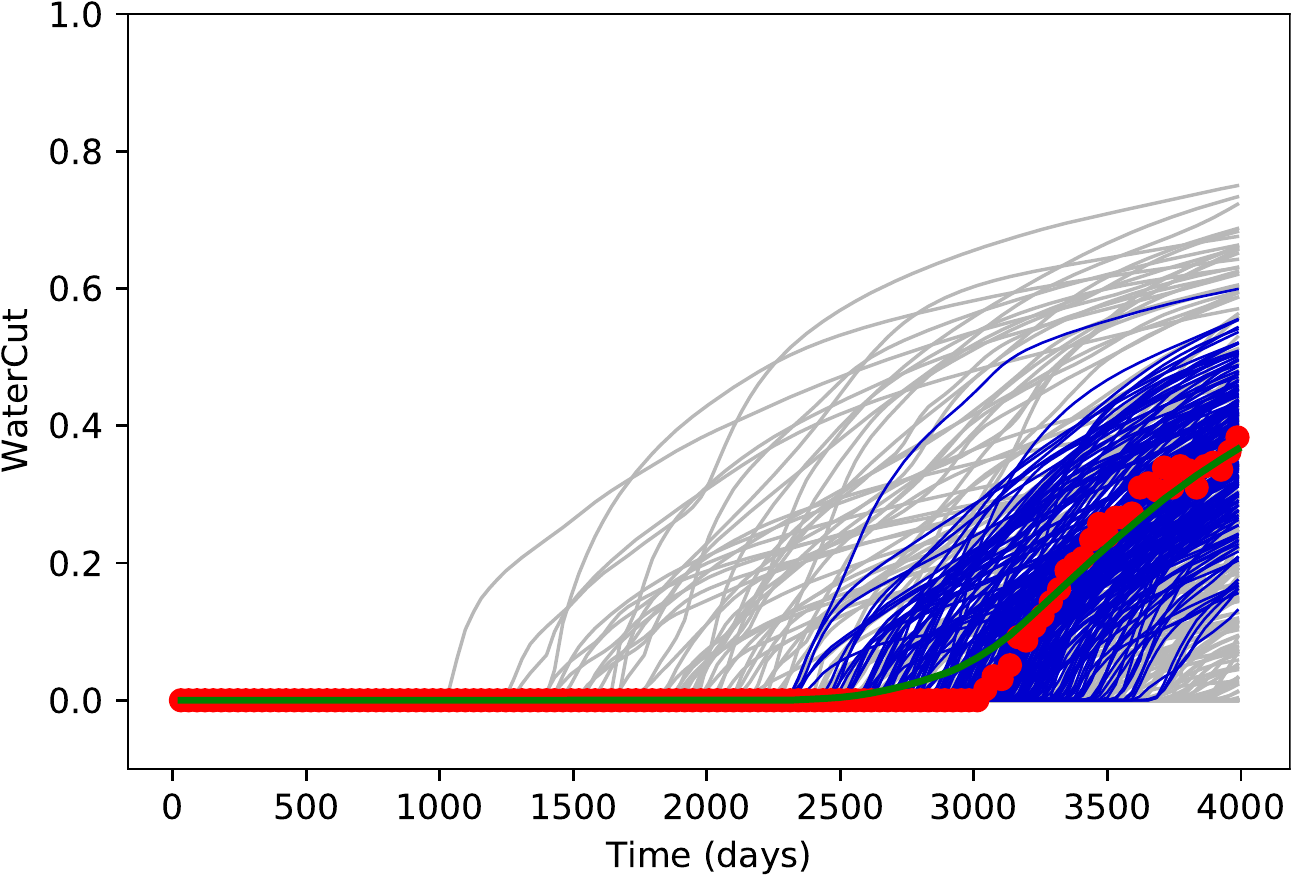}~\includegraphics[width=0.42\linewidth]{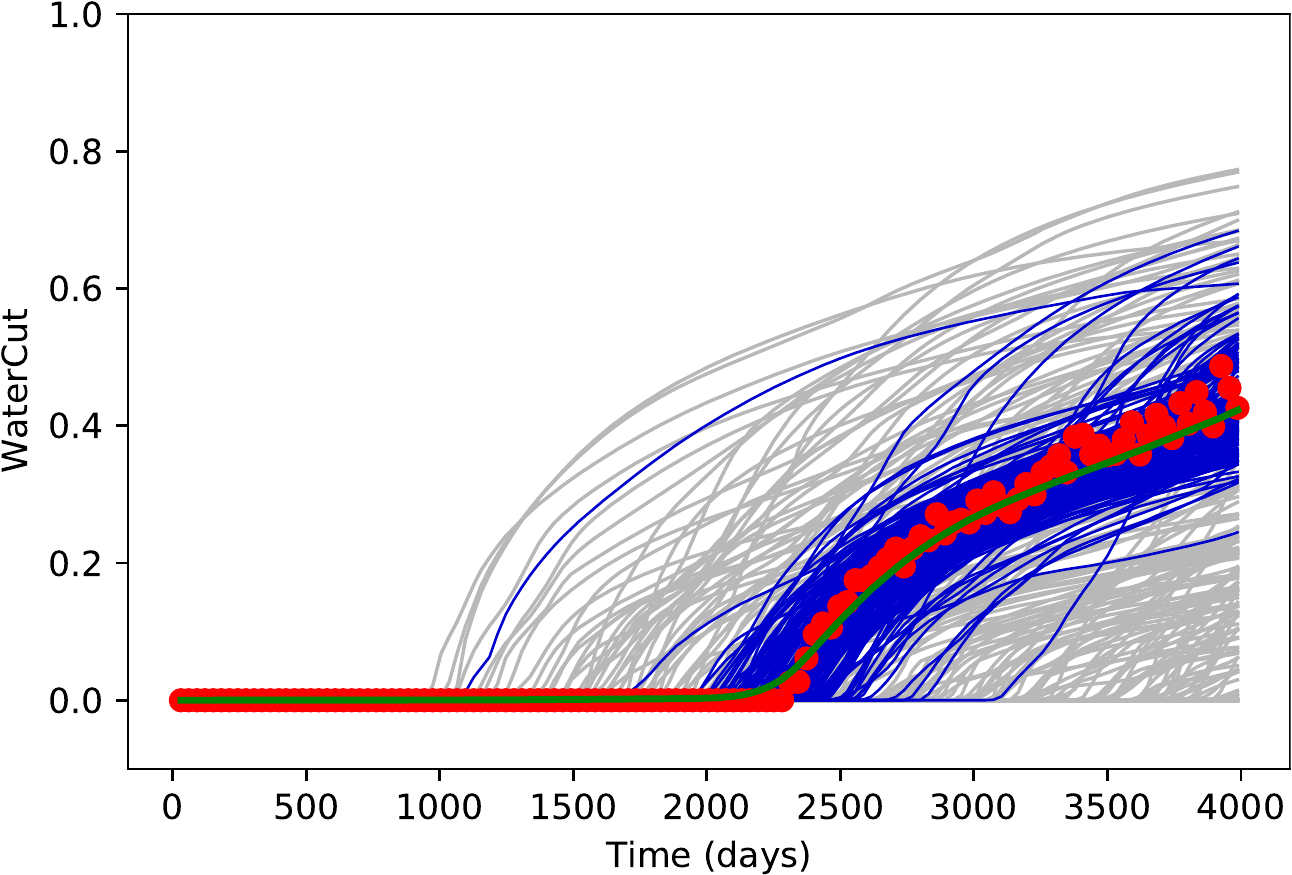}}
\caption{Water cut for wells P5 (left) and P11 (right). Case 2. The red circles correspond to the observed data points. The gray and blue curves correspond to the predictions from the prior and posterior ensembles, respectively. The green curve is the posterior mean.}
\label{Fig:Case2-WCT}
\end{figure}

Fig.~\ref{Fig:Case2-BoxPlot} shows box-plots of normalized objective function. This figure shows that VAE-Local, VAE-Local-VAE and PCA-Cycle-GAN-Local resulted in similar data-match quality. The VAE case (without localization) resulted in lower objective function values, but at a cost of a severe variance reduction in the ensemble variance. Fig.~\ref{Fig:Case2-WCT} illustrates the data matches by showing the predicted water cut data for two wells. It is interesting to note that even for the VAE case with indications of ensemble collapse, we observe an spread of predicted water cut. This happens because the time of water breakthrough is very sensitive to the position of the channels, and even small variations cause significant changes in the predicted water cut.

\clearpage

\section{Final Remarks}
\label{Sec:conclusions}

In this paper, we investigated the use of deep generative models as re-parameterization techniques combined with ES-MDA for facies history matching. In the first part of the paper, we tested the performance of seven network formulations including VAE, GAN, Wasserstein GAN, variational auto-encoding GAN, Cycle-GAN, transfer style networks and VAE with style loss. These networks were tested in a small reservoir model with channelized facies. The main findings can be summarized as follows:

\begin{itemize}
  \item All networks were able to generate realistic facies realizations with well-defined channels.
  \item We tested the effect of small perturbations in the latent representation implied by each network. The results shows that the generator of the trained GAN was relatively insensitive to perturbations in $\z$, which may be an indication of mode collapse during the training of the GAN.
  \item The training time of the networks is a limiting factor of the method. The fastest training required 23 minutes and the slowest required 4.5 hours using a single GPU, which is expensive considering that the test problem has only 3600 gridblocks.
  \item The combination of the networks with ES-MDA resulted in successful assimilation of facies and production data.
  \item The combination of WGAN and ES-MDA showed slower convergence during the history matching compared to the other networks.
\end{itemize}

The second part of the paper proposed two strategies to allow the use of distance-based localization with the deep learning parameterizations. The first strategy is more general and can be used with virtually any parameterization. It is based on local analysis, which makes the process computationally more demanding. The second localization strategy assumes the existence of a spatial relationship between the latent representation and the facies. We used PCA to generate the latent representation in the same grid of the reservoir model, which makes this strategy applicable only for PCA-Cycle-GAN and PCA-Style. We tested both localization strategies in a 2D reservoir model with a larger number of channels and wells. For the first localization strategy we used a VAE network. For the second one, we used the PCA-Cycle-GAN. The following conclusions were observed:

\begin{itemize}
  \item The data assimilation using VAE combined with ES-MDA without localization resulted in a near collapse of the ensemble variance, showing the importance of the use of some localization strategy for this problem.
  \item Combining the local analysis procedure with the trained VAE decoder resulted in ``noisy'' facies reconstructions. This problem was resolved by passing the generated facies realizations in the same VAE after the ES-MDA update.
  \item Both localization strategies were able to resolve the ensemble collapse, resulting in plausible facies distributions conditioned to production data.
\end{itemize}

This paper is part of a research project aiming to develop a robust parameterization for facies history matching. The next steps include the investigation of the performance of these networks in 3D problems. These problems require the use of 3D convolutional layers, which makes the training process very challenging. Moreover, the parameterizations with transfer style networks, PCA-Style and VAE-Style, cannot be directly applied because these networks rely on a VGG network trained only for 2D images. One alternative is to use a layer-by-layer strategy, but it may not preserve the vertical continuity of channels. Another remaining challenge is the computational cost of the training process. So far we have been testing the network in small test problems. In practice, however, geological models can have a few millions gridblocks, which cannot be addressed with our current implementations.

\section{Computer Code Availability}

\label{Sec:Code}
All implementation of this paper is available at \href{https://github.com/smith31t/GeoFacies_DL/}{\texttt{github.com/smith31t/GeoFacies\_DL}}. The network architectures were designed used \verb"Keras"  framework with \verb"TensorFlow" as the backend engine. The data set and the \verb"Python" codes of the neural networks and the ensemble smoother are freely available in this repository.

\section{Acknowledgments}

The authors thank Petrobras for the financial support.


\clearpage


\section*{Appendix: Architecture of the networks}
\label{Sec:appendix}

\begin{table}[ht!]
\caption{VAE network architecture}
\label{Tab:VAE}
\begin{scriptsize}
\begin{center}
\begin{tabular}{lll}
\toprule
\textbf{Layer} & \textbf{Configuration}  & \textbf{Comment} \\
\midrule
  \multicolumn{3}{c}{\textbf{Encoder}}\\\cmidrule{2-2}
  Input & Shape = (60, 60, 2) & Two facies \\
  2D convolution 1 & Kernels = 128, size = (3, 3), stride = (2, 2), activ. = ReLU  & -- \\
  2D convolution {2} & Kernels = 64, size = (3, 3), stride = (2, 2), activ. = ReLU & -- \\
  2D convolution {3} & Kernels = 32, size = (3, 3), stride = (1, 1), activ. = ReLU & -- \\
  Flatten & -- & Setup for the fully-connected \\
  Fully-connected {1} & Neurons = 2048, activ. = ReLU & -- \\
  Dropout &  10\%  & Strategy to avoid overfitting \\
  Fully-connected {2} & Neurons = 500, activ. = linear & Mean of the VAE ($\mu$) \\
  Fully-connected {3} & Neurons = 500, activ. = linear & Log-variance of the VAE ($\log \sigma$) \\
\midrule
  \multicolumn{3}{c}{\textbf{Code}}\\\cmidrule{2-2}
  Lambda &  $\z = \mu + \sigma \widehat{\z}$, where $\widehat{\z} \sim \mathcal{N}(\mathbf{0},\mathbf{I})$ & Sampling $\z$\\
\midrule
  \multicolumn{3}{c}{\textbf{Decoder}}\\\cmidrule{2-2}
  Fully-connected {1} & Neurons = 2048, activ. = ReLU & -- \\
  Dropout &  10\%  & Strategy to avoid overfitting \\
  Fully-connected {2} & Neurons = 7200, activ. = ReLU & -- \\
  Reshape & Output size = (15, 15, 32) & Setup for the transp. convol. \\
  2D transp. conv. {1} & Kernels = 32, size = (3, 3), stride = (1, 1), activ. = ReLU & -- \\
  2D transp. conv. {2} & Kernels = 64, size = (3, 3), stride = (2, 2), activ. = ReLU & -- \\
  2D transp. conv. {3} & Kernels = 128, size = (3, 3), stride = (2, 2), activ. = ReLU & -- \\
  Bilinear up-sampling & Output size = (60, 60, 128) & Resize output dimension \\
  2D convolution {1} & Kernels = 2, size = (3, 3), stride = (1, 1), activ. = sigmoid & Output image \\
\bottomrule
\end{tabular}
\end{center}
\end{scriptsize}
\end{table}

\clearpage

\begin{table}
\caption{GAN network architecture, including the encoder network}
\label{Tab:GAN}
\begin{scriptsize}
\begin{center}
\begin{tabular}{lll}
\toprule
\textbf{Layer} & \textbf{Configuration}  & \textbf{Comment} \\
\midrule
\multicolumn{3}{c}{\textbf{Encoder}}\\\cmidrule{2-2}
  Input & Shape = (60, 60, 2) & Two facies \\
  2D convolution {1} & Kernels = 32, size = (5, 5), stride = (2, 2), activ. = ReLU  & -- \\
  2D convolution {2} & Kernels = 64, size = (5, 5), stride = (2, 2), activ. = ReLU & -- \\
  2D convolution {3} & Kernels = 128, size = (5, 5), stride = (2, 2), activ. = ReLU & -- \\
  2D convolution {4} & Kernels = 256, size = (5, 5), stride = (1, 1), activ. = ReLU & -- \\
  Flatten  & -- & Setup for the fully-connected \\
  Dropout  &  50\%  & Strategy to avoid overfitting \\
  Fully-connected & Neurons = 500, activ. = linear & -- \\
  BatchNormalization & $\mu$=0, $\sigma$=1 & -- \\
  \midrule
  \multicolumn{3}{c}{\textbf{Generator}}\\\cmidrule{2-2}
  Input & Shape = (500) & Latent vector \\
  Fully-connected & Neurons = 8$\times$8$\times$256, activ. = ReLU & -- \\
  Reshape & Output size = (8, 8, 256) & Setup for the transp. convol. \\
  2D transp. conv. {1} & Kernels = 128, size = (5, 5), stride = (2, 2), activ. = ReLU & -- \\
  2D transp. conv. {2} & Kernels = 64, size = (5, 5), stride = (2, 2), activ. = ReLU & -- \\
  Bilinear up-sampling & Output size = (30, 30, 64) & Resize output dimension \\
  2D transp. conv. {3} & Kernels = 32, size = (5, 5), stride = (2, 2), activ. = ReLU & -- \\
  2D convolution & Kernels = 2, size = (5, 5), stride = (1, 1), activ. = tanh & Output image \\
  \midrule
  \multicolumn{3}{c}{\textbf{Discriminator}}\\\cmidrule{2-2}
  Input & Shape = (60, 60, 2) & Two facies \\
  2D convolution {1} & Kernels = 32, size = (4, 4), stride = (2, 2), activ. = ReLU  & -- \\
  2D convolution {2} & Kernels = 64, size = (4, 4), stride = (2, 2), activ. = ReLU & -- \\
  2D convolution {3} & Kernels = 128, size = (4, 4), stride = (2, 2), activ. = ReLU & -- \\
  2D convolution {4} & Kernels = 256, size = (4, 4), stride = (1, 1), activ. = ReLU & -- \\
  Flatten & -- & Setup for the fully-connected \\
  Fully-connected & Neurons = 1, activ. = sigmoid & -- \\
\bottomrule
\end{tabular}
\end{center}
\end{scriptsize}
\end{table}

\clearpage

\begin{table}
\caption{$\alpha$-GAN network architecture}
\label{Tab:alpha-GAN}
\begin{scriptsize}
\begin{center}
\begin{tabular}{lll}
\toprule
\textbf{Layer} & \textbf{Configuration}  & \textbf{Comment} \\
\midrule
\multicolumn{3}{c}{\textbf{Encoder}}\\\cmidrule{2-2}
  Input & Shape = (60, 60, 2) & Two facies \\
  2D convolution {1} & Kernels = 32, size = (4, 4), stride = (2, 2), activ. = ReLU  & -- \\
  2D convolution {2} & Kernels = 64, size = (4, 4), stride = (2, 2), activ. = ReLU & -- \\
  2D convolution {3} & Kernels = 128, size = (4, 4), stride = (2, 2), activ. = ReLU & -- \\
  2D convolution {4} & Kernels = 256, size = (4, 4), stride = (1, 1), activ. = ReLU & -- \\
  Flatten & -- & Setup for the fully-connected \\
  Dropout &  50\% & Strategy to avoid overfitting \\
  Fully-connected {1} & Neurons = 500, activ. = linear & -- \\
\midrule
\multicolumn{3}{c}{\textbf{Code Discriminator}}\\\cmidrule{2-2}
  Input & Shape = (500) & Latent vector \\
  Fully-connected {1} & Neurons = 750, activ. = ReLU & -- \\
  Fully-connected {2} & Neurons = 750, activ. = ReLU & -- \\
  Fully-connected {3} & Neurons = 1, activ. = sigmoid & -- \\
\midrule
\multicolumn{3}{c}{\textbf{Discriminator}}\\\cmidrule{2-2}
  Input & Shape = (60, 60, 2) & Two facies \\
  2D convolution {1} & Kernels = 32, size = (4, 4), stride = (2, 2), activ. = ReLU  & -- \\
  2D convolution {2} & Kernels = 64, size = (4, 4), stride = (2, 2), activ. = ReLU & -- \\
  2D convolution {3} & Kernels = 128, size = (4, 4), stride = (2, 2), activ. = ReLU & -- \\
  2D convolution {4} & Kernels = 256, size = (4, 4), stride = (1, 1), activ. = ReLU & -- \\
  Flatten & -- & Setup for the fully-connected \\
  Fully-connected {5} & Neurons = 1, activ. = sigmoid & -- \\
\midrule
\multicolumn{3}{c}{\textbf{Generator}}\\\cmidrule{2-2}
  Input & Shape = (500) & Latent vector \\
  Fully-connected & Neurons = 8$\times$8$\times$256, activ. = ReLU & -- \\
  Reshape & Output size = (8, 8, 256) & Setup for the transp. convol. \\
  2D transp. conv. {1} & Kernels = 128, size = (5, 5), stride = (2, 2), activ. = ReLU & -- \\
  2D transp. conv. {2} & Kernels = 64, size = (5, 5), stride = (2, 2), activ. = ReLU & -- \\
  Bilinear up-sampling & Output size = (30, 30, 64) & Resize output dimension \\
  2D transp. conv. {3} & Kernels = 32, size = (5, 5), stride = (2, 2), activ. = ReLU & -- \\
  2D convolution & Kernels = 2, size = (5, 5), stride = (1, 1), activ. = tanh & Output image \\
\bottomrule
\end{tabular}
\end{center}
\end{scriptsize}
\end{table}

\clearpage

\begin{table}
\caption{Cycle-GAN network architecture}
\label{Tab:PCA-Cycle-GAN}
\begin{scriptsize}
\begin{center}
\begin{tabular}{lll}
\toprule
\textbf{Layer} & \textbf{Configuration}  & \textbf{Comment} \\
\midrule
\multicolumn{3}{c}{\textbf{Generator 1}}\\\cmidrule{2-2}
  Input & Shape = (60, 60, 1) & PCA realization \\
  2D convolution {1} & Kernels = 32, size = (9, 9), stride = (1, 1), activ. = ReLU  & -- \\
  2D convolution {2} & Kernels = 64, size = (5, 5), stride = (2, 2), activ. = ReLU & -- \\
  2D convolution {3} & Kernels = 128, size = (5, 5), stride = (2, 2), activ. = ReLU & -- \\
  Residual block {1} & Kernels = 128, size = (5, 5), stride = (1, 1), activ. = ReLU & -- \\
  Residual block {2} & Kernels = 128, size = (5, 5), stride = (1, 1), activ. = ReLU & -- \\
  Residual block {3} & Kernels = 128, size = (5, 5), stride = (1, 1), activ. = ReLU & -- \\
  Dropout & 50\% & -- \\
  Residual block {4} & Kernels = 128, size = (5, 5), stride = (1, 1), activ. = ReLU & -- \\
  Dropout & 50\% & -- \\
  Residual block {5}  & Kernels = 128, size = (5, 5), stride = (1, 1), activ. = linear & -- \\
  Dropout & 50\% & -- \\
  2D transp. conv. {1} & Kernels = 64, size = (5, 5), stride = (2, 2), activ. = linear & -- \\
  Concatenate {1} & Features maps = (Residual block 5, 2D convolution 2) & -- \\
  Activation & ReLU & -- \\
  2D transp. conv. {2} & Kernels = 32, size = (5, 5), stride = (2, 2), activ. = linear & -- \\
  Concatenate {2} & Features maps = (2D transp. conv. 2, 2D convolution 1) & -- \\
  Activation & ReLU & -- \\
  2D transp. conv. {3} & Kernels = 2, size = (9, 9), stride = (1, 1), activ. = tanh & Output image \\
\midrule
\multicolumn{3}{c}{\textbf{Generator 2}}\\\cmidrule{2-2}
  Input & Shape = (60, 60, 2) & Two facies \\
  2D convolution {1} & Kernels = 32, size = (9, 9), stride = (1, 1), activ. = ReLU  & -- \\
  2D convolution {2} & Kernels = 64, size = (5, 5), stride = (2, 2), activ. = ReLU & -- \\
  2D convolution {3} & Kernels = 128, size = (5, 5), stride = (2, 2), activ. = ReLU & -- \\
  Residual block {1} & Kernels = 128, size = (5, 5), stride = (1, 1), activ. = ReLU & -- \\
  Residual block {2} & Kernels = 128, size = (5, 5), stride = (1, 1), activ. = ReLU & -- \\
  Residual block {3} & Kernels = 128, size = (5, 5), stride = (1, 1), activ. = ReLU & -- \\
  Dropout & 50\% & -- \\
  Residual block {4} & Kernels = 128, size = (5, 5), stride = (1, 1), activ. = ReLU & -- \\
  Dropout & 50\% & -- \\
  Residual block {5} & Kernels = 128, size = (5, 5), stride = (1, 1), activ. = linear & -- \\
  Dropout & 50\% & -- \\
  2D transp. conv. {1} & Kernels = 64, size = (5, 5), stride = (2, 2), activ. = linear & -- \\
  Concatenate {1} & Features maps = (Residual block 5, 2D convolution 2) & -- \\
  Activation & ReLU & -- \\
  2D transp. conv. {2} & Kernels = 64, size = (5, 5), stride = (2, 2), activ. = linear & -- \\
  Concatenate {2} & Features maps = (2D transp. conv. 2, 2D convolution 1) & -- \\
  Activation & ReLU & -- \\
  2D transp. conv. {2} & Kernels = 1, size = (9, 9), stride = (1, 1), activ. = linear & Output image \\
\midrule
\multicolumn{3}{c}{\textbf{Discriminator 1}}\\\cmidrule{2-2}
  Input & Shape = (60, 60, 1) & PCA realization \\
  2D convolution {1} & Kernels = 32, size = (4, 4), stride = (2, 2), activ. = ReLU  & -- \\
  2D convolution {2} & Kernels = 64, size = (4, 4), stride = (2, 2), activ. = ReLU & -- \\
  2D convolution {3} & Kernels = 128, size = (4, 4), stride = (2, 2), activ. = ReLU & -- \\
  2D convolution {4} & Kernels = 256, size = (4, 4), stride = (1, 1), activ. = ReLU & -- \\
  Flatten & -- & Setup for the fully-connected \\
  Fully-connected  & Neurons = 1, activ. = sigmoid & -- \\
\midrule
\multicolumn{3}{c}{\textbf{Discriminator 2}}\\\cmidrule{2-2}
  Input & Shape = (60, 60, 2) & Two facies \\
  2D convolution {1} & Kernels = 32, size = (4, 4), stride = (2, 2), activ. = ReLU  & -- \\
  2D convolution {2} & Kernels = 64, size = (4, 4), stride = (2, 2), activ. = ReLU & -- \\
  2D convolution {3} & Kernels = 128, size = (4, 4), stride = (2, 2), activ. = ReLU & -- \\
  2D convolution {4} & Kernels = 256, size = (4, 4), stride = (1, 1), activ. = ReLU & -- \\
  Flatten & -- & Setup for the fully-connected \\
  Fully-connected & Neurons = 1, activ. = sigmoid & -- \\
\bottomrule
\end{tabular}
\end{center}
\end{scriptsize}
\end{table}

\clearpage

\begin{table}
\caption{Style network architecture}
\label{Tab:Style}
\begin{scriptsize}
\begin{center}
\begin{tabular}{lll}
\toprule
\textbf{Layer} & \textbf{Configuration}  & \textbf{Comment} \\
\midrule
\multicolumn{3}{c}{\textbf{Tranfert style net}}\\\cmidrule{2-2}
  Input & Shape = (60, 60, 1) & PCA realization \\
  2D convolution {1} & Kernels = 32, size = (9, 9), stride = (1, 1), activ. = ReLU  & -- \\
  2D convolution {2} & Kernels = 64, size = (3, 3), stride = (2, 2), activ. = ReLU & -- \\
  2D convolution {3} & Kernels = 128, size = (3, 3), stride = (2, 2), activ. = ReLU & -- \\
  Residual block {1} & Kernels = 128, size = (3, 3), stride = (1, 1), activ. = ReLU & -- \\
  Residual block {2} & Kernels = 128, size = (3, 3), stride = (1, 1), activ. = ReLU & -- \\
  Residual block {3} & Kernels = 128, size = (3, 3), stride = (1, 1), activ. = ReLU & -- \\
  Residual block {4} & Kernels = 128, size = (3, 3), stride = (1, 1), activ. = ReLU & -- \\
  Residual block {5} & Kernels = 128, size = (3, 3), stride = (1, 1), activ. = ReLU & -- \\
  2D transp. conv. {1} & Kernels = 64, size = (3, 3), stride = (2, 2), activ. = ReLU & -- \\
  2D transp. conv. {2} & Kernels = 32, size = (3, 3), stride = (2, 2), activ. = ReLU & -- \\
  2D transp. conv. {3} & Kernels = 1, size = (9, 9), stride = (1, 1), activ. = tanh & Output image \\
\bottomrule
\end{tabular}
\end{center}
\end{scriptsize}
\end{table}

\begin{table}
\caption{Residual Block}
\label{Tab:Residual-Block}
\begin{scriptsize}
\begin{center}
\begin{tabular}{lll}
\toprule
{\textbf{Layer}} & {\textbf{Configuration}}  & {\textbf{Comment}} \\
\midrule
  {Input} & {Shape = ($N_i/4$, $N_j/4$, 128)} & {input tensor} \\
  {2D convolution} {1} & {Kernels = 128, size = (3, 3)}, {stride = (1, 1)}, {activ. = ReLU}  & {-- }\\
  {2D convolution} {2} & {Kernels = 128}, {size = (3, 3)}, {stride = (1, 1)}, {activ. = Linear} & {--} \\
  {Add}  & {Features maps = (2D convolution 1, 2D convolution 2)} & {--} \\
\bottomrule
\end{tabular}
\end{center}
\end{scriptsize}
\end{table}


\end{document}